\documentclass[twoside,11pt]{article}

\usepackage{jmlr2e}
\usepackage{macros}

\ShortHeadings{Distribution-dependent and Time-uniform Bounds for Piecewise i.i.d Bandits}{Mukherjee and Maillard}
\firstpageno{1}

\begin{document}

\title{Distribution-dependent and Time-uniform Bounds for Piecewise i.i.d Bandits}

\author{\name Subhojyoti Mukherjee \email subho@cs.umass.edu \\
       \addr College of Information and Computer Sciences\\
       University of Massachusetts Amherst\\
       Amherst, MA 01003, USA
       \AND
       \name Odalric-Ambrym Maillard \email odalric.maillard@inria.fr \\
       \addr SequeL Team\\
       Inria Lille - Nord Europe\\
       59650 Villeneuve d’Ascq, France}

%\editor{Kevin Murphy and Bernhard Sch{\"o}lkopf}
%\editor{To be decided}

\maketitle

\begin{abstract}
We consider the setup of stochastic multi-armed bandits in the case when reward distributions are piecewise i.i.d. and bounded with unknown changepoints. We focus on the case when changes happen simultaneously on all arms, and in stark contrast with the existing literature, we target gap-dependent (as opposed to only gap-independent) regret bounds involving the magnitude of changes $(\Delta^{chg}_{i,g})$ and optimality-gaps ($\Delta^{opt}_{i,g}$). Diverging from previous works, we assume the more realistic scenario that there can be undetectable changepoint gaps and under a different set of assumptions, we show that as long as the compounded delayed detection for each changepoint is bounded there is no need for forced exploration to actively detect changepoints. We introduce two adaptations of UCB-strategies that employ scan-statistics in order to actively detect the changepoints, without knowing in advance the changepoints and also the mean before and after any change. Our first method \UCBLCPD does not know the number of changepoints $G$ or time horizon $T$ and achieves the first time-uniform concentration bound for this setting using the Laplace method of integration. The second strategy \ImpCPD makes use of the knowledge of $T$ to achieve the order optimal regret bound of $\min\big\lbrace O(\sum\limits_{i=1}^{K} \sum\limits_{g=1}^{G}\frac{\log(T/H_{1,g})}{\Delta^{opt}_{i,g}}), O(\sqrt{GT})\big\rbrace$, (where $H_{1,g}$ is the problem complexity) thereby closing an important gap with respect to the lower bound in a specific challenging setting. Our theoretical findings are supported by numerical experiments on synthetic and real-life datasets.
\end{abstract}


\section{Introduction}
\label{psbandit:intro}
We consider the piecewise i.i.d multi-armed bandit problem, an interesting variation of the stochastic multi-armed bandit (SMAB) setting. \blfootnote{An initial version was accepted at Reinforcement Learning for Real Life (RL4RealLife) Workshop in the 36 th International Conference on Machine Learning, Long Beach, California, USA, 2019. Copyright 2019 by the author(s).} The learning algorithm is provided with a set of decisions (or arms) which belong to the finite set $\A$ with individual arm indexed by $i$ such that $i={1,\!\ldots,\! K}$. The learning proceeds in an iterative fashion, wherein each time step $t$, the algorithm chooses an arm $i\in\A$ and receives a stochastic reward that is drawn from a distribution specific to the arm selected. There exist a finite number of changepoints $G$ such that the reward distribution of \textit{all} arms changes at those changepoints. $\G$ denotes the finite set of changepoints indexed by $g  = {1,\!\ldots,\! G}$, and $t_g$ denotes the time step when the changepoint $g$ occurs. At $\tau_g$ the learner detects $t_g$ with some delay. The reward distributions of all arms are unknown to the learner. The learner has the goal of maximizing the cumulative reward at the end of the horizon $T$. Our setting follows the restless bandit model \citep{whittle1988restless}, \citep{auer2002nonstochastic}  where the distribution of the arms evolve independently, irrespective of the arm being pulled. In contrast, the rested bandit setting \citep{warlop2018fighting} assumes the distribution evolves when an arm is pulled. A special case is observed for the rotting bandits \citep{heidari2016tight}, \citep{warlop2018fighting} when the reward of an arm decreases when it is pulled. In this paper, we focus on the \textcolor{blue}{global changepoint} setup, with abrupt change of mean, in the \textcolor{blue}{restless model}.

The global changepoint piecewise i.i.d setting is extremely relevant to a lot of practical areas such as recommender systems, industrial manufacturing, and medical applications. In the health-care domain, the non-stationary assumption is more realistic than the i.i.d. assumption, and thus progress in this direction is important. An interesting use-case of this setting arises in drug-testing for a cure against a resistant bacteria, or virus such as AIDS. Here, the arms can be considered as various treatments while the feedback can be considered as to how the bacteria/virus reacts to the treatment administered. It is common in this setting that the behavior of the bacteria/virus changes after some time and thus its response to \textit{all} the arms also change simultaneously. Moreover, in the health-care domain, it is extremely risky to conduct additional forced exploration of arms to detect simultaneous abrupt changes and relying only on the past history of interactions might lead to reliable detection of changepoints and safer policies. 

\begin{figure}[!th]
\centering
\vspace*{-0.8em}
\begin{tabular}{cc}
\setlength{\tabcolsep}{0.1pt}
\subfigure[0.25\textwidth][\centering A $2$-arm and $2$ changepoint scenario.]
{
    \includegraphics[width=0.5\linewidth]{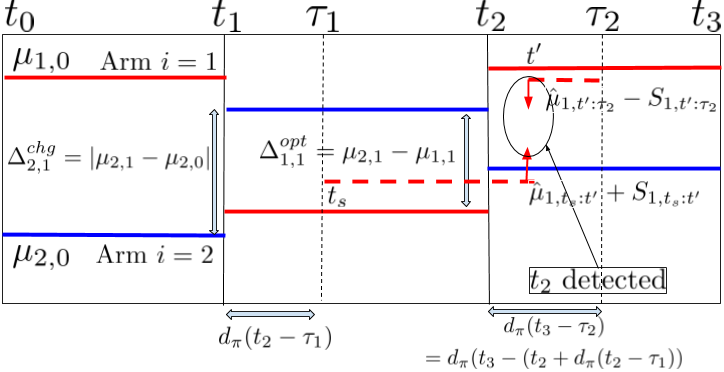}
\label{psbandit:fig:explanation}
    }
\subfigure[0.25\textwidth][\centering Expt-$1$: The effect of $\eta$ on \ImpCPD and \UCBLCPD]
    {
    \includegraphics[width=0.4\linewidth]{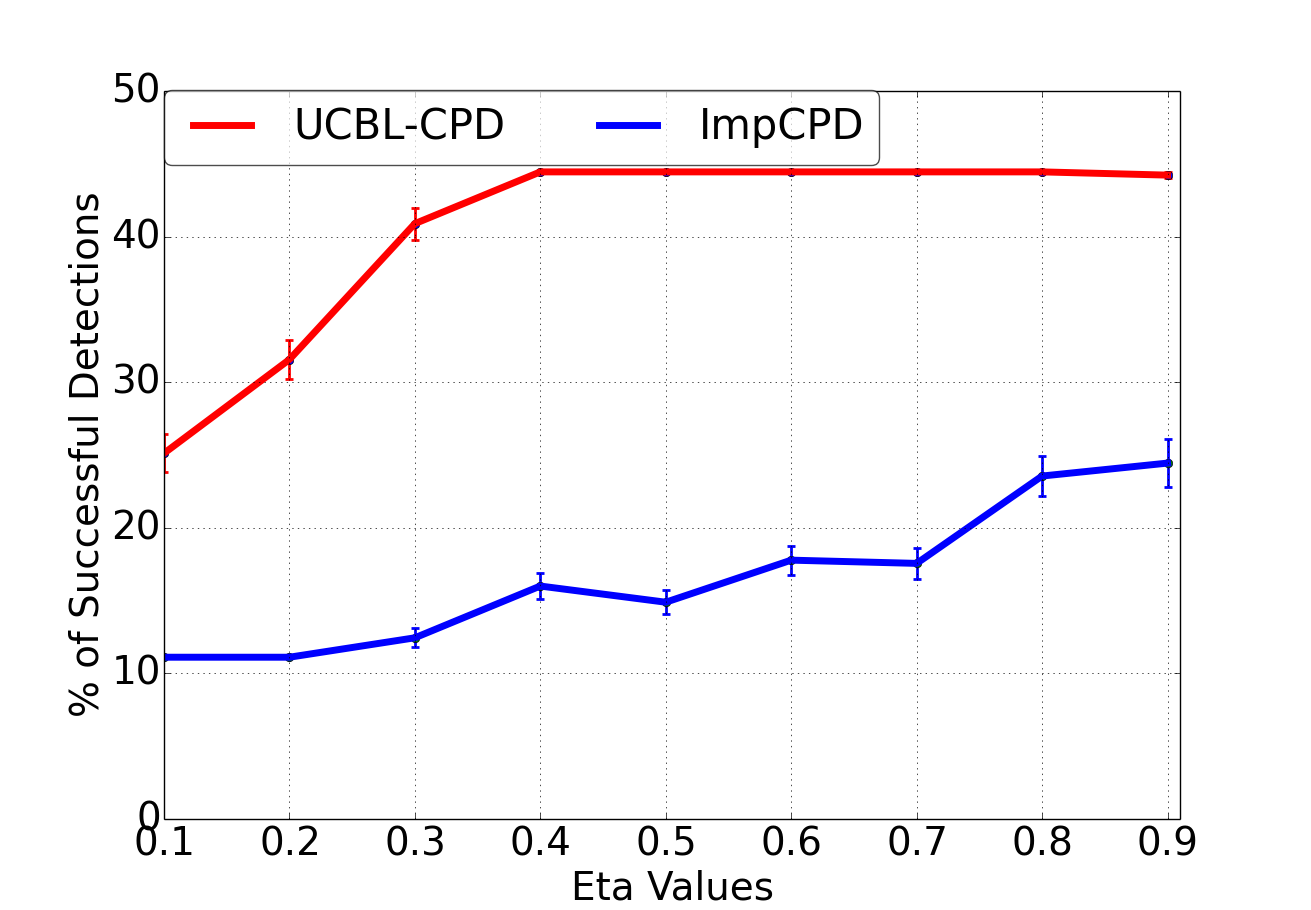}
  		\label{psbandit:fig:eta}
    }
 \end{tabular}
 \vspace*{-1em}
    \caption{Illustration of Changepoint Detection}
    \label{fig:karmed0}
    \vspace*{-1em}
\end{figure}

A reasonable detection policy $\pi$ has to contend with two things, 1) to find the optimal arm between $t_{g-1}$ to $t_g$ with high probability, 2) to detect abrupt changes and restart. Previous works \citep{liu2017change}, \citep{cao2018nearly}, \citep{besson2019generalized} have concentrated mainly on the assumption that at $t_g$ mean of any arm can change. Moreover, \citet{cao2018nearly}, \citet{besson2019generalized} also consider the per-arm local changepoint setting forcing them to conduct additional exploration of arms to detect abrupt changes, leading to gap-independent results. We consider a slightly stricter global changepoint assumption \ref{assm:global} such that at every $t_g, \forall g\in\G$ the mean of all the arm changes allowing us to derive stronger gap-dependent results. Additionally, the vast majority of the previous settings assumed that all changepoints are well-separated, and all changepoint gaps are significant so that they can be detected well in advance (see \textcolor{blue}{Table \ref{tab:comp-bds}}). Diverging from this line of thinking we propose a more realistic setting where there can be undetectable changepoint gaps and $\pi$ will not be able to detect all changepoints with high probability. This leads to assumption \ref{assm:space-gap} such that for three consecutive changepoints $t_{g-1}$, $t_g$, and $t_{g+1}$, even if $t_g$ has low probability of detection due to large detection delay of $t_{g-1}$, sufficient number of observations are available to detect $t_{g+1}$. Finally, we introduce assumption \ref{assm:chg-gap} to restrict the scenario where a series of undetectable changepoint gaps for the same arms are encountered. An illustrative explanation of a $2$-arm $2$ changepoint scenario is shown in Table  \ref{psbandit:fig:explanation}. Our main contributions are as follows:

\textbf{1. Problem setup:} We consider the more realistic setting where there can be \textcolor{blue}{undetectable changepoint gaps}. In assumption \ref{assm:global} we assume that at each changepoint the mean of all the arms change which leads to the removal of forced exploration of arms allowing us to derive the first gap-dependent logarithmic regret for this setting. Additionally, assumption \ref{assm:space-gap} and assumption \ref{assm:chg-gap} allows us to tackle some worst-case scenarios even when there are undetectable changepoint gaps.

\textbf{2. Algorithmic:} We propose two actively adaptive upper confidence bound (UCB) algorithms, referred to as UCBLaplace-Changepoint Detector (\UCBLCPD) and Improved Changepoint Detector (\ImpCPD). Unlike CD-UCB \citep{liu2017change}, M-UCB \citep{cao2018nearly} and CUSUM \citep{liu2017change}, \UCBLCPD and \ImpCPD \textcolor{blue}{do not conduct forced exploration} to detect changepoints. They divide the time into slices, and for each time slice, for every arm, they check the UCB, LCB mismatch based on past observations only to detect changepoints. While \UCBLCPD checks for every such combination of time slices, \ImpCPD only checks at certain estimated points in time horizon and hence saves on computation time. 

\begin{table}[!th]
\begin{center}
\vspace*{-0.8em}
\caption{Comparison of Algorithms}
\label{tab:comp-bds}
\begin{tabular}{|p{7.5 em}|p{2.7 em}|p{0.7em}|p{0.7em}|p{4.6em}|p{5.5 em}|p{6.5em}|}
\toprule
Algorithm  &   \hspace*{1mm}Type & T & G & Assumptions & Gap-Dependent & Gap-Independent\\
\hline
\ImpCPD (ours)    & Active  & Y & N & \ref{assm:global}, \ref{assm:space-gap}, \ref{assm:chg-gap} & Theorem \ref{psbandit:Theorem:2} & $\! O\left(\sqrt{GT}\right)$\\
\UCBLCPD (ours)     & Active & N & N & \ref{assm:global}, \ref{assm:space-gap}, \ref{assm:chg-gap} & Theorem \ref{psbandit:Theorem:1} & $\! O\left(\sqrt{GT} \log T\right)$\\%\midrule
CUSUM             & Active & Y & Y & \ref{assm:global}\footnote{Also works for per-arm changepoint setting. \label{footnote:0}}, 4\footnote{Assumption 4: CUSUM and M-UCB requires a constant large separation between changepoints. M-UCB requires the constant to be known for theoretical guarantees. \label{footnote:1}}, 5\footnote{Assumption 5: Requires all changepoint gaps above a known minimum threshold to be detectable. \label{footnote:2}} & N/A & $\! O\left( \sqrt{GT\log \frac{T}{G}}\right)$\\%\midrule
M-UCB             & Active & Y & Y & 1$^{\ref{footnote:0}}$, 4$^{\text{\ref{footnote:1}}}$, 5$^{\ref{footnote:2}}$ &N/A & $\! O\left(\sqrt{GT\log T}\right)$\\
EXP3.R         & Active & Y & N & 5$^{\text{\ref{footnote:2}}}$ &N/A& $\! O\left( G\sqrt{T\log T}\right)$\\%\midrule
DUCB        & Passive & Y & Y & 1$^{\text{\ref{footnote:0}}}$, 5$^{\text{\ref{footnote:2}}}$ & N/A & $\! O\left(\sqrt{GT}\log T\right)$\\
SWUCB         & Passive & Y & Y & 1$^{\text{\ref{footnote:0}}}$, 5$^{\text{\ref{footnote:2}}}$  & N/A& $\! O\left(\sqrt{GT\log T}\right)$\\%\midrule
Lower Bound         & Oracle & Y & Y & \ref{assm:global}, \ref{assm:chg-gap} & Theorem \ref{psbandit:Theorem:3} & $\! \Omega\left( \sqrt{GT}\right)$\\\midrule
\end{tabular}
\end{center}
\vspace*{-2em}
\end{table}
\textbf{3. Regret bounds:} The CDUCB, CUSUM uses the Hoeffding inequality and M-UCB uses McDiarmid’s inequality with union bound to obtain the regret bound. DUCB \citep{garivier2011upper} and SWUCB \citep{garivier2011upper} uses the peeling argument which results in slightly tighter concentration bound but both these techniques results in less tight bounds than the Laplace method of integration used for \UCBLCPD which has the strongest bound proposed for this setting. Moreover, \UCBLCPD has \textcolor{blue}{time-uniform bound} as its confidence interval does not depend explicitly on $t$ as opposed to other methods. On the contrary, \ImpCPD which is not anytime and has access to $T$ uses the usual union bound with geometrically increasing phase length to bound the regret. Both these proofs are of independent interest which can be used in other settings as well. A detailed comparison of the union, peeling and Laplace bound can be found in discussion \ref{Disc:Laplace}. We prove the first \textcolor{blue}{gap-dependent logarithmic regret upper bound} that consist of both the changepoint gaps ($\Delta^{chg}_{i,g}$) and optimality gaps ($\Delta^{opt}_{i,g}$) for each changepoint $g\!\in\!\G$, $i\!\in\!\A$ in Theorem \ref{psbandit:Theorem:1}, and  \ref{psbandit:Theorem:2} (under assumptions \ref{assm:global}, \ref{assm:space-gap}, and  \ref{assm:chg-gap}).     
We introduce the hardness parameter $H^{}_{1,g}$ and $H_{2,g}$ which captures the problem complexity of the piecewise i.i.d setting. For the gap-independent result we show in the challenging case when all the gaps are small such that for all $i\!\in\!\A,g\!\in\!\G$, $\Delta^{opt}_{i,g} = \Delta^{chg}_{i,g} = \Delta(t_g,\delta)$, $H^{}_{1,g}\! =\! {K}{(\Delta(t_g,\delta))^{-2}}$ and $H_{2,g}\!=\! 1$, \UCBLCPD and \ImpCPD achieves $\sqrt{GT}\log T$ and $\sqrt{GT}$ respectively (\textcolor{blue}{Table \ref{tab:comp-bds}}, and Corollary \ref{psbandit:Corollary:1}). We show that \UCBLCPD and \ImpCPD perform very well across diverse piecewise i.i.d environments even when our modelling conditions do not hold (Section \ref{psbandit:expt}). 

    The rest of the paper is organized as follows. We first setup the problem in Section~\ref{psbandit:notations}. Then in Section~\ref{psbandit:algorithm} we present the changepoint detection algorithms. Section~\ref{psbandit:results} contains our main result, remarks and discussions. Section~\ref{psbandit:expt} contains numerical simulations, Section~\ref{psbandit:related1} contains related works, and we conclude in Section \ref{psbandit:conclusion}. The proofs are provided in Appendices in the supplementary material.

\section{Problem Setup}
\label{psbandit:notations}
\subsection{Preliminaries}
Let $t_0=1$ by convention, so that $t_0<\!t_1<\!\!\dots<\!\! t_G$. We introduce the time interval (piece) $\rho_g = [t_g, t_{g+1}-1]$, so the piece $\rho_g$ starts at $t_g$ and ends at $t_{g+1}-1$. Let, $\mu_{i,g}$ be the expectation of an arm $i$ for the piece $\rho_{g}$. The learner does not know $t_g,\forall g\in\G$ or even $|\G|$. The reward drawn for the $i$-th arm, when it is pulled, for the $t$-th time instant is denoted by $X_{i,t}$. 
%\begin{figure}[!th]
%\centering
%\vspace*{-1.7em}
%\begin{tabular}{cc}
%\setlength{\tabcolsep}{0.1pt}
%\subfigure[0.25\textwidth][A $2$-arm and $2$ changepoint scenario]
%    %with $r_{i_{{i}\neq {*}}}=0.07$ and $r^{*}=0.1$
%    {
%          \includegraphics[scale=0.28, trim={0 0 0 0}, clip]{img/PiecewiseIID.png}
%          \label{psbandit:fig:explanation}
%    }
%\setlength{\tabcolsep}{0.1pt}
%\subfigure[0.25\textwidth][The effect of $\eta$ on \ImpCPD and \UCBLCPD on $3$-arm Bernoulli setting.]
%    %with $r_{i_{{i}\neq {*}}}=0.07$ and $r^{*}=0.1$
%    {
%          \includegraphics[scale=0.16, trim={0 0 0 25}, clip]{img/figure_eta2.png}
%          \label{psbandit:fig:eta}
%    }
%\end{tabular}
%\vspace*{-1em}
%\caption{Illustration of changepoint detection}
%\end{figure}
%\vspace*{-0.8em}
%\begin{wrapfigure}{l}{0.36\textwidth}
%\includegraphics[scale=0.14, trim={0 0 0 25}, clip]{img/figure_eta2.png}
%          \label{psbandit:fig:eta}
%          \caption{The effect of $\eta$ on \ImpCPD and \UCBLCPD on $3$-arm Bernoulli setting.}
%\end{wrapfigure}
We assume all rewards are bounded in $[0,1]$. $N_{i,\tau_{g}:\tau_{g+1}-1}$ denotes the number of times arm $i$ has been pulled between $\tau_{g}$ to $\tau_{g+1}-1$ timesteps for any sequence of increasing $(\tau_g)_{g}$ of integers. Also, we define $\hat{\mu}_{i,\tau_{g}:\tau_{g+1}-1}$ as the empirical mean of the arm $i$ between $\tau_{g}$ to $\tau_{g+1}-1$ timesteps. 
%Let $\Delta_{\epsilon_0,g}$ be the minimum changepoint gap at $t_g$ that can be detected with $(1 - \epsilon_0)$ probability by collecting a sample of size $n_{\epsilon_0,g}$. Note, that $n^{}_{\epsilon_0, g}$, and $ \Delta^{}_{\epsilon_0, g}$ is not indexed by $i$ as it is common for any arm $i\in\A$  satisfying the condition.
We consider that on each arm i,  the process generating the reward is piecewise mean constant according to the sequence $(t_g)_{g}$. That is, if $\mu_{i,t}$ denotes the mean reward of arm i at time t, then $\mu_{i,t}$ has same value for all $t \in \rho_g$. 
\begin{definition}
\label{Def:tcj}
We define $t_g$ as, $t_g = \min \{ t > t_{g-1} : \exists i, \mu_{i,t-1} \neq \mu_{i,t}  \}$.
%\vspace*{-0.4em}
%\begin{align*}
%t_g = \min \{ t > t_{g-1} : \exists i, \mu_{i,t-1} \neq \mu_{i,t}  \}
%\end{align*}
\end{definition}

\begin{assumption}\textbf{(Global changepoint)}
\label{assm:global}
We assume the global changepoint setting, that is $t_{g}=t$ implies $\mu_{i_{t-1}} \neq \mu_{i_{t}}$, for all $i\in\A$. 
\end{assumption}
%and $i^\ast_{t-1} \neq i^\ast_t$
%such that $t_{g}$ is common across all the arms and the learner does not know the $t_{g},\forall g\in\G$. 
\begin{definition}
\label{Def:chg-gap}
The changepoint gap at $t_{g}$ for an arm $i\in\A$ between the segments $\rho_{g}$ and $\rho_{g+1}$ is denoted by $\Delta^{chg}_{i,g}=|\mu_{i,g}- \mu_{i,g+1}|.$
%\begin{align*}
%\Delta^{chg}_{i,g}=|\mu_{i,g}- \mu_{i,g+1}|.
%\end{align*}
\end{definition}
Thus, at each changepoint we assume that the mean of all the arm changes. Note, that this assumption is stricter than \citet{liu2017change}, \citet{cao2018nearly}, \citet{besson2019generalized} where at $t_g$, $\mu_{i_t}$ of any arm may or may not change requiring the forced exploration of all arms to detect changepoint. Next we incorporate a more general and realistic structure to our problem. We assume that at $t_g$ there could be undetectable changepoint gaps (Definition \ref{Def:und-gap}) and every changepoint $t_g$ need not be detectable with high probability  but three consecutive changepoints $t_{g-1}, t_g$ and $t_{g+1}$ must follow assumption \ref{assm:space-gap}.
%Other arms $i\in\A\setminus \{i^\ast_{t-1} \cup i^\ast_t \}$ may or may not change, or it could be two different optimal arms for consecutive sections or it could be the same optimal arm with changing mean. Therefore, we make a distinction between the finite set of all arms $\A$ and $\A_g^{chg}$, such that $\A_g^{chg}$ denotes only those arms $i\in\A$ whose $\Delta^{chg}_{i,g} > 0$ at the $t_g$-th changepoint. Note, that this assumption is stricter than  \citep{liu2017change}, \citep{cao2018nearly}, \citep{besson2019generalized} where at $t_g$, $\mu_{i^*_t}$ may or may not change while other arm(s) change requiring the forced exploration of all arms to detect changepoint.
% The learner does not know the $t_{g},\forall g\in\G$.
% and the consecutive section $\rho_{g-1}$ and $\rho_g$ have two different optimal arms.
Now, we carefully setup our problem. First, we distinguish between any best detection policy $\pi^*$ which has observations exactly from $t_{g-1}$ and any reasonable detection policy $\pi$ which has observations after $t_{g-1}$ with some delay. We denote the delay of a policy $\pi$ restarting from $\tau_{g-1}\! > \! t_{g-1}$ as $d_{\pi}(t_g - \tau_{g-1})$. We define the minimum number of samples $\textcolor{blue}{n(t_g, \Delta,\delta)}$ required for an arm $i\in\A$ so that a deviation of $\Delta > 0$ of $\hat{\mu}_{i,g-1}$ (or $\hat{\mu}_{i,g}$) from $\mu_{i,g-1}$ (or $\mu_{i,g}$)  before (or after) the $t_g$-th changepoint can be controlled with $(1 - \delta)$ probability. This is shown in Lemma \ref{psbandit:Lemma:0}. We then bound the delay of $\pi^\ast$ in detecting a changepoint at $t_g$ starting exactly from $t_{g-1}$ (Lemma \ref{psbandit:Lemma:01}). Finally, we define the detectable changepoint gap $\Delta(t_g,\delta)$ based on the maximal delay of $\pi^\ast$ and $n(t_g, \Delta,\delta)$.
% Next we define the notion of how large a gap $\Delta(t_g,\delta)$ is possible to detect at the $t_g$-th changepoint based on .

\begin{customlemma}{1}\textbf{(Control of large deviations)}
\label{psbandit:Lemma:0}
For our detection policy $\pi^\ast$ using estimated means starting exactly from $t_{g-1}$, it is sufficient to collect a minimum number of samples $n(t_g,\Delta,\delta) = \lceil\frac{1}{2}\log(\frac{2 (t - t_{g-1})^2}{\delta})/\Delta^2 \rceil$ for a single arm $i\in\A$, a changepoint $g\in\G$  before or after $t_g$ so that $|\hat{\mu}_{i,g-1} - \mu_{i,g-1}| \leq \Delta$ or $|\hat{\mu}_{i,g} - \mu_{i,g}| \leq \Delta$ with $(1-\delta)$ probability and $t_{g-1} < t_g < t$.
%the probability of large deviation from $\mu_{i,g-1}$ or $\mu_{i,g}$ is bounded by $(1-\delta)$ probability, where $t_{g-1} < t_g < t$.
%to detect a minimum changepoint of magnitude $\Delta^{}_{\epsilon_0, g}$ between timestep $t_0$ and $t$ with probability $(1-\epsilon_0)$, it is sufficient to collect a minimum number of samples $n^{}_{\epsilon_0, g} \! \geq \! \dfrac{\log(\frac{{2}(t - t_0)^2}{{\epsilon_0}})}{2\left(\Delta^{}_{\epsilon_0, g}\right)^2}$ for an arm $i$,  such that $t_0 \! < \!  t \! < \! t_g$, and $\epsilon_{0}\in (0,1)$. 
\end{customlemma}
%where $\tau_{g}$ is the time the algorithm detects the changepoint $g\in \G$
\begin{customproof}{1}
The proof of Lemma \ref{psbandit:Lemma:0} is given in Appendix \ref{sec:proof:Lemma:0}.
\end{customproof}
\begin{assumption} \textbf{(Separated Changepoints)}
\label{assm:space-gap}
We assume that for every two consecutive segments $\rho_{g-1}$ and $\rho_{g}, \forall g\in\G$ all the three changepoints $t_{g-1}, t_{g}$ and $t_{g+1}$ satisfy the following condition,
\begin{align*}
t_g \! + \! d_{\pi^\ast}\left(t_{g} \! - \! t_{g-1}\right) \! \leq \! t_g \! + \!  \eta (t_{g+1}-t_g)  \! = \!  \eta t_{g+1} + (1-\eta) t_g
\end{align*}
where $\eta \in(0,1)$, $d_{\pi^\ast}(t_{g} - t_{g-1})$ is the maximal delay of best detection policy $\pi^*$ starting at $t_{g-1}$.
\end{assumption}

%\begin{assumption}
%\label{assm:space-gap}
%We assume that for every two consecutive segments $\rho_{g}$ and $\rho_{g+1}, \forall g=1,2,\ldots,G$ all the three changepoints $t_{g-1}, t_{g}$ and $t_{g+1}$ satisfy the following condition,
%\begin{align*}
%\dfrac{d_{\pi}(t_{g} - t_{g-1})}{t_{g+1} - t_{g}} \leq \dfrac{\beta_0}{1-\beta_0}
%\end{align*}
%where $d_{\pi}(t_{g} - t_{g-1})$ denotes the maximal delay of policy $\pi$ in detecting a changepoint at $t_{g}$ and $ \beta_0 \in (0,1)$.
%\end{assumption}

Using Lemma \ref{psbandit:Lemma:0} and Assumption \ref{assm:space-gap} now we properly define \textcolor{blue}{detectable changepoint gap $\Delta(t_g, \delta)$} to be such $\Delta(t_g, \delta)  \geq   \sqrt{\log(2( x ^2 / \delta))  / 2x}$ where $x = t_g + d_{\pi^*}(t_g - t_{g-1}) - t_{g-1}$. 
\begin{customlemma}{2}\textbf{(Detection Delay)}
\label{psbandit:Lemma:01}
With the standard assumption that at $t_g$, $\Delta(t_g, \delta)$ scales atleast as $\Omega(\sqrt{\frac{\log t}{t}})$ for $K$ arms, then to detect a deviation of $\Delta\geq \Delta(t_g, \delta)$ at $t_g$ with $(1-\delta)$ probability, $\pi^*$ suffers a worst case maximum delay of 
%\begin{align*}
$d_{\pi^\ast}(t_g \!-\! t_{g-1}) \!\leq \!\! \left( \frac{C(t, \delta, \eta)K\log(\frac{t^2}{{\delta}})}{2(\Delta(t_g, \delta))^{2}}\right) \!+ \!K\delta$,
%\end{align*}
where $C(t, \delta, \eta) \leq \eta \log (t/\delta)$.
%, and $t > t_g$
\end{customlemma}
\begin{customproof}{2}
The proof of Lemma \ref{psbandit:Lemma:01} is given in Appendix \ref{sec:proof:Lemma:01}.
\end{customproof}
\begin{definition}
\label{Def:e-chg-gap}
The $\Delta^{chg}_{i,g}$ for an arm $i\in\A, g\in\G$ is a $\delta$-optimal changepoint gap if $\Delta^{chg}_{i,g} \geq \Delta (t_g, \delta)$.  Let $\A_g^{chg}$ denote only those arms $i\in\A$ whose $\Delta^{chg}_{i,g} \geq  \Delta (t_g, \delta)$ at the $t_g$-th changepoint.
\end{definition}
\begin{definition}
\label{Def:und-gap}
The $\Delta^{chg}_{i,g}$ for an arm $i\in\A, g\in\G$ is an  \textcolor{blue}{undetectable gap} if $\sqrt{\frac{e}{T}} \!\!\leq\!\!\Delta^{chg}_{i,g} \!\!<\!\! \Delta (t_g, \delta )$.  
\end{definition}
\begin{definition}
\label{Def:und-chp}
A changepoint $t_g$ is \textcolor{blue}{undetectable changepoint} if $\exists \Delta^{chg}_{i,g}$ at $t_g$ s.t. $\sqrt{\frac{e}{T}} \!\!\leq\!\!\Delta^{chg}_{i,g} \!\!<\!\! \Delta (t_g, \delta )$.  
\end{definition}
\begin{assumption}\textbf{(Isolated Changepoints)}
\label{assm:chg-gap}
We assume that at $t_g$ each gap $\!\Delta^{chg}_{i,g}$ is either $\delta$-optimal or is undetectable. If $t_g$ is undetectable, then $\!\Delta^{chg}_{i,g-1}$ and $\!\Delta^{chg}_{i,g+1}$ for all $i\in\A$ are $\delta$-optimal gaps.
%If $\!\Delta^{chg}_{i,g}$ is undetectable, then $\!\Delta^{chg}_{i,g-1}$ and $\!\Delta^{chg}_{i,g+1}$ are $\delta$-optimal gaps.
\end{assumption}
% i.e $0 \!<\!\!\Delta^{chg}_{i,g} \!\!<\!\! \Delta (t_g, \delta )$
%for $i\in A\setminus \{i^*_t \bigcup i^*_{t-1}\}$
%\begin{assumption}\textbf{(Minimum gap)}
%\label{assm:chg-gap}
%We assume that $\forall g\in\G$ the changepoint gap $\Delta^{chg}_{i,g},\forall i\in\A_g^{chg}$ are either $\delta$-optimal gaps or are undetectable.
%\end{assumption}
\begin{discussion}
\label{dis:gap-delay}
Thus, assumption \ref{assm:space-gap} makes sure that $t_g+d_{\pi^\ast}(t_{g} - t_{g-1})$ stays away from $t_{g+1}$. This ensures that when restarting the detection strategy from $t_g +d_{\pi^\ast}(t_{g} - t_{g-1})$, the detection of $t_{g+1}$ will not be too much endangered. This is a mild assumption as without this, changepoints could be \textcolor{blue}{too frequent} so that they cannot be detected on time before the next change happens. A $\delta$-optimal detectable changepoint gap requires a minimum sample of $n(t_g, \Delta, \delta)$ to ensure that $\Delta$ deviation from the mean can be detected with $(1-\delta)$ probability (Lemma  \ref{psbandit:Lemma:0}). An undetectable gap requires more than $n(t_g,\Delta,\delta)$ samples or possibly unbounded samples for ensuring $\Delta\geq \Delta(t_g,\delta)$ deviation. Assumption \ref{assm:chg-gap} ensures that there cannot be a series of undetectable gaps \textcolor{blue}{to avoid linear regret}. If at $t_g$, $\Delta^{chg}_{i,g}$ is undetectable then $\Delta^{chg}_{i,g-1}$ and $\Delta^{chg}_{i,g+1}$ are $\delta$-optimal gaps. Policy $\pi$ has its maximal delay $d_{\pi}(t_{g} - \tau_{g-1})\geq d_{\pi^\ast}(t_{g} \! - \! t_{g-1})$ as it only has observation to detect $t_g$ from $\tau_{g-1}$ and not from $t_{g-1}$ (unlike $\pi ^*$). $\pi$ tries to minimize $d_{\pi}(t_{g} \!- \!\tau_{g-1})$ in detecting $t_{g}$ such that $d_{\pi}(t_{g} \! -\! \tau_{g-1})\leq (1+\beta)d_{\pi^\ast}(t_{g}\! - \! t_{g-1})$ holds with high probability for some $\beta\in (0,1)$ and $\eta\in(0,1)$. This approach is different than the vast majority of current litertaure \cite{cao2018nearly}, \cite{liu2017change}, \cite{besson2019generalized} as they assume that all the previous changepoints can be detected with high probability or all changepoints have constant large separation between them which is unrealistic. In Lemma \ref{psbandit:Lemma:01} the assumption that $\Delta(t_g,\delta)$ scales atleast as $\Omega(\sqrt{\log t /t})$ is \textcolor{blue}{not a global minimum gap assumption} as there could be smaller undetectable isolated gaps by assumption \ref{assm:chg-gap}. In contrast \citet{cao2018nearly}, \citet{liu2017change} assumes that all gaps are large enough to be detectable. For a large $d_{\pi}(t_{g} - \tau_{g-1})$, sufficient number of observations may not be available to detect $t_g$ but this will not endanger detection of $t_{g+1}$ by assumption \ref{assm:space-gap} and \ref{assm:chg-gap}. The effect of $\eta$ on our proposed algorithms is shown in Figure \ref{psbandit:fig:eta}. The details of the environment are mentioned in section \ref{psbandit:expt}, experiment 1. The key takeaway is that lower $\eta$ values leads to lower subsequent segments $(t_{g+1} - t_g)$. So detection policy $\pi^*$ starting exactly at $t_{g-1}$ has to have lower delay $d_{\pi^\ast}(t_{g} - t_{g-1})$ to successfully detect subsequent changepoints. Conversely, $\pi^*$ is deemed good if it has less delay for low $\eta$ values. \UCBLCPD has higher percentage of successful detection and less variance for all $\eta$ values than \ImpCPD as it implements the Laplace bound as opposed to the loose union bound of \ImpCPD.
%(as CD-UCB, CUSUM)
%This is due to the time-uniform bound implemented by \UCBLCPD which improves upon the loose union bound of \ImpCPD (also M-UCB,  CUSUM). 
%In Theorem \ref{psbandit:Theorem:1} and \ref{psbandit:Theorem:2} from the respective $\eta$ values we will subsequently see that the delay of \UCBLCPD is less than \ImpCPD.
\end{discussion}

%In contrast, we assume that some changepoints may not be detectable.
%\begin{discussion}
%\label{dis:gap-delay}
%Thus, Assumption \ref{assm:space-gap} makes sure that $t_g+d\left(t_{g} - t_{g-1}\right)$ stays away from $t_{g+1}$. This ensures that when restarting the detection strategy from $t_g +d\left(t_{g} - t_{g-1}\right)$, the detection of $t_{g+1}$ will not be too much endangered. A reasonable detection strategy tries to minimize the maximal delay $d(t_{g} - t_{g-1})$ in detection of a changepoint at $t_{g}$ by achieving atmost a delay of $d\left(t_{g} - t_{g-1}\right) \leq O\left(\frac{K\log( t_{g} - t_{g-1} )}{(\Delta^{chg}_{i,g})^{2}}\right)$. Here, the gap $\Delta^{chg}_{i,g} \geq \Delta^{}_{\epsilon_0, g}$ is a $\epsilon_0$-optimal changepoint gap as in Assumption \ref{assm:chg-gap}. The additional factor $K$ in the maximal delay $d$ is there because the strategy may pull all arms $i = 1,2,\ldots K -1 $ at least $(n^{}_{\epsilon_0, g} - 1)$ times before detecting the changepoint for the last arm in $\A$. Note, that when the gaps are same such that for all $i\in\A$, $\forall g\in\G$, $\Delta_{i,g}^{chg}=\Delta_{\epsilon_0, g}^{}$, then a good strategy must ensure $d\left(t_{g} - t_{g-1}\right) \leq \left( \frac{CK\log( t_{g} - t_{g-1} )}{(\Delta^{}_{\epsilon_0, g})^{2}}\right)$, where $C$ is a constant. 
%\end{discussion}
%The gap $\Delta^{chg}_{i,g} \geq \Delta^{\epsilon_0}_{g}$ with probability  $1-\epsilon_0$, as proved in Lemma \ref{psbandit:Lemma:0}
\begin{definition}
\label{Def:opt-gap}
The optimality gap $\Delta^{opt}_{i,g}$ for an arm $i_{t'}\neq i^*_{t'},\forall t'\in[t_{g-1},t_{g}-1]$ is 
%\begin{align*}
$\Delta^{opt}_{i,g}= \mu_{i^*,g}-\mu_{i,g}.$
%\end{align*}
\end{definition}
%We denote $\Delta^{chg}_{\max,g} = \max_{i\in\A}  \Delta^{chg}_{i,g}$, $\Delta^{opt}_{\max,g} = \max_{i\in\A}  \Delta^{opt}_{i,g}$,  and $\Delta^{opt}_{\min}=\min_{i\in\A,g\in\G} \Delta^{opt}_{i,g}$.

%\textbf{Regret Definition:} 
\subsection{Regret Definition}
The objective of the learner is to minimize the cumulative regret till $T$, which is defined as, $R_{T}=\sum_{t'=1}^T \mu_{i^*_{t'}} - \sum_{t'=1}^T \mu_{ i_{t'}}  \indic{i_{t'}  \neq i^*_{t'} }$, 
%\vspace*{-0.6em}
%\begin{align*}
%R_{T}=\sum_{t'=1}^T \mu_{i^*_{t'}} - \sum_{t'=1}^T \mu_{ i_{t'}}  \indic{i_{t'}  \neq i^*_{t'} }
%\end{align*}
%\mu_{\mathbb{I}_{t'} = i \neq i^*_{t'}}
%\vspace*{-0.0 em}
where $T$ is the horizon, $\mu_{i^*_{t'}}$ is the expected mean of the optimal arm at the $t'$ timestep and $\mu_{ i_{t'}}  \indic{i_{t'}  \neq i^*_{t'} }$ is the expected mean of the arm chosen by the learner at the $t'$ timestep when it was not the optimal arm $i^*_{t'}$. The expected regret of an algorithm after $T$ timesteps can be written as,
%Let $N_{i,g}$ is the number of times the learner has chosen arm $i_{}$ between $t_{g-1}$ to $t_{g}$.
%when it was not the optimal arm $i^*_{j}$
%$n_{i_{t'}\neq i^*_{t'},\forall t'=1:T}$
%\vspace*{-4em}
\vspace*{-0.8em}
\begin{align*}
\E[R_{T}] = \E\!\left[\sum_{t'=1}^T \mu_{i^*_{t'}} - \sum_{t'=1}^T \mu_{ i_{t'}}  \indic{i_{t'}  \neq i^*_{t'} }\right]
&\overset{(a)}{=} \!\!\E\left[\sum_{g=1}^{G}\sum_{t'=t_{g-1}}^{t_{g}} \!\mu_{i^*_{t'}} \! - \!  \sum_{g=1}^{G}\sum_{t'=t_{g-1}}^{t_{g}}  \!\mu_{ i_{t'}}  \indic{i_{t'}  \neq i^*_{t'} } \!\right]\\
%\begin{align*}
%%%%%%%%%%%%%%%%%%%%%%%%%%%%%%%%%%%%%%%%
%%%%%%%%%%%%%%%%%%%%%%%%%%%%%%%%%%%%%%%%
&\overset{(b)}{=}\sum_{g=1}^{G}\sum_{i = 1}^K\Delta^{opt}_{i,g}\E[N_{i,t_{g-1}:t_g}]
\end{align*}
%\mu_{\mathbb{I}_{t'} = i \neq i^*_{t'}}
%\vspace*{-4em}
where $(a)$ is from Assumption \ref{assm:global}, and $(b)$ from Definition \ref{Def:opt-gap}. 
%\indic{ i = i_{t'}}

\subsection{Problem Complexity}
%\textbf{Problem Complexity:} 
We define the hardness of a changepoint $g\in\G$ using optimality and changepoint gaps by modifying the  definitions of \emph{problem complexity} as introduced in \citet{audibert2010best} for stochastic bandits. Let 
%\begin{align*}
$\! \textcolor{blue}{H^{}_{1,g}} \! = \!\max\!\left\lbrace\!\sum_{i=1}^{K}\!\frac{1}{(\Delta^{opt}_{i,g})^{2}}, \sum_{i\in\A^{chg}_g}^{}\!\frac{1}{(\Delta^{chg}_{i,g})^{2}}\! \right \rbrace\text{, and } \hspace{0mm} \textcolor{blue}{H_{2,g}}\!=\! \!\frac{\Delta^{opt}_{\max,g+1}}{\Delta(t_g,\delta)}$, 
%%%%%%%%%%%%%%%%%%%%%
%H^{}_{2,g} &=\max\left\lbrace\min_{i\in \mathcal{A}}\dfrac{i}{{(\Delta^{opt}_{(i),g})^{2}}}, \min_{i\in \mathcal{A}}\dfrac{i}{{(\Delta^{chg}_{(i),g})^{2}}} \right\rbrace
%\end{align*}
%where $(\Delta^{.}_{(i),g}: i\in\mathcal{A})$ is obtained by arranging $(\Delta^{.}_{i,g}:i\in\mathcal{A})$ in an increasing order. 
where $\Delta^{opt}_{\max,g+1}= \max_{i\in\A}\Delta^{opt}_{i,g+1}$. The hardness parameter $H_{2,g}$ captures the tradeoff between the minimum detectable gap $\Delta(t_g,\delta)$ and maximum optimality gap of the next changepoint $\Delta^{opt}_{\max,g+1}$ which serves as an upper bound to all such possible trade-offs at changepoint $g$. The relation between the above complexity terms can be derived as, 
%\begin{align*}
$H_{2,g}  \leq H^{}_{1,g} \leq \frac{K}{(\Delta(t_g,\delta))^2}(H_{2,g})$.
%H_{3,g} \leq H^{}_{2,g} \leq H^{}_{1,g}\leq \log(2K)H_{2,g}.
%\end{align*}
Note that, $\frac{\Delta^{opt}_{i,g}}{\Delta^{chg}_{i,g}} \leq H_{2,g}, \forall i\in\A^{chg}_g, g\in\G$. In the challenging case when all gaps are small and equal i.e. $\Delta^{opt}_{i,g} = \Delta^{chg}_{i,g} = \Delta(t_g,\delta), \forall i\in\A, g\in\G$, then,
%\begin{align*}
$H_{1,g} =  {K}{(\Delta(t_g,\delta))^{-2}},\text{ and } H_{2,g} = 1$.
%\end{align*}

%\textbf{Repeating worst case:} 
\subsection{Repeating worst case}
We think that both assumption \ref{assm:space-gap} and \ref{assm:chg-gap} are required to avoid a combination of worst cases in absence of forced exploration. There can be a scenario where for each $g\in\G$, $\pi$ keeps on pulling sub-optimal arm $i$ and then $t_g$ happens. At $t_g$, $\Delta^{chg}_{i,g}$ is undetectable so $\pi$ does not restart. So at $t_{g+1}$, assumption \ref{assm:chg-gap} will prevent this. Furthermore, assumption \ref{assm:space-gap} ensures that consecutive changepoints $t_{g-1}$, $t_g$ and $t_{g+1}$ are such that even if the detection of $t_g$ is a low probability event (undetectable changepoint), there are sufficient number of observations left to detect $t_{g+1}$. 

%Regardless, in future works we wish to remove one of these assumptions. 

%$\Delta^{chg}_{i^*,g}$ is $\delta$-optimal and all other gaps are undetectable. $\pi$ may select all $K-1$ arms uniform randomly and may not have sufficient number of observations left to detect $|\mu_{i^*, g} - \mu_{i^*, g+1}|$.So assumption \ref{assm:space-gap} guarantees that even if $t_g$ is not detected there are sufficient number of observations left to detect $t_{g+1}$. Regardless, in future works we wish to remove one of these assumptions. 

%Note that $\Delta^{chg}_{i^*,g}$ is $\delta$-optimal requires that $\Delta^{chg}_{i^*,g} \> \Omega(\sqrt{\log t /t})$ and thus 

%So, if $\eta$ scales atleast as in Theorem \ref{psbandit:Theorem:1} and \ref{psbandit:Theorem:2} then $t_{g+1}$ will be detectable if assumption \ref{assm:space-gap} holds. Regardless, in future works we wish to remove one of these assumptions.

\section{Algorithms}
\label{psbandit:algorithm}
%\subsection{Proposed Algorithms}
We first introduce the policy \UCBLCPD in Algorithm \ref{alg:UCBCPD} which is an adaptive algorithm based on the standard UCB1 \citep{auer2002finite} approach. \UCBLCPD pulls an arm at every timestep as like UCB1 but has the time-uniform concentration bound that holds simultaneously for all timestep $t$. It calls upon the Changepoint Detection (CPD)  subroutine in Algorithm \ref{alg:CPD} for detecting a changepoint. Note, that unlike CD-UCB, CUSUM, and M-UCB the \UCBLCPD \textcolor{blue}{does not conduct forced exploration} to detect changepoints. \UCBLCPD is an anytime algorithm which does not require the horizon as an input parameter or to tune its parameter $\delta$. This is in stark contrast with CD-UCB, CUSUM, M-UCB, DUCB or SWUCB, that require the knowledge of $G$ or $T$ for optimal performance. we define the confidence interval of \UCBLCPD as follows: 
\begin{eqnarray}
S_{i,t_s:t_p} &:= \sqrt{\left(1+\frac{1}{N_{i,t_s:t_p}}\right)\frac{\log(\sqrt{N_{i,t_s:t_p}+1}/\delta)}{2N_{i,t_s:t_p}}}
\vspace*{-3em}
 \label{eq:laplace}
\end{eqnarray}
%\begin{minipage}{0.56\linewidth}
\begin{algorithm}
\caption{UCB Laplace CPD  (\UCBLCPD)}
\label{alg:UCBCPD}
\begin{algorithmic}[1]
\State {\bf Input:} $\delta > 0$; 
\State {\bf Definition:} $S_{i,t_s:tp}$ from \eqref{eq:laplace} 
\State {\bf Initialization:} $t_s := 1$, $t_p := 1$.
%\State {\bf New Expert:} Start a new expert $f_{t_s}$ and add it to $\M$.
\State Pull each arm once
\For{$t=K+1,..,T$}
\State Pull arm $j\in\argmax_{i\in\A}\left\{ \hat{\mu}_{i,t_s:t_p} + S_{i,t_s:t_p}\right\}$, observe reward $X_{j,t}$.
\State Update $\hat{\mu}_{j,t_s:t_p}$, $N_{j,t_s:t_p}:=N_{j,t_s:t_p} + 1.$
\State $t_p := t_p + 1.$
\If{ (CPD($t_s$, $t_p$, $\delta$))}
\State {\bf Restart:} Set $\hat{\mu}_{i,t_s:t_p} := 0$, $N_{i,t_s:t_p}:=0$, $\forall i \in\A$, $t_s := t$, $t_p := t_s$. 
\State Pull each arm once.
\EndIf
\EndFor
\end{algorithmic} 
\end{algorithm}       
%\end{minipage}
%\quad
%\begin{minipage}{0.4\linewidth}
\begin{algorithm}
\caption{Changepoint Detection($t_s$, $t_p$, $\delta$) (CPD)}
\label{alg:CPD}
\begin{algorithmic}[1]
%\State \textbf{Definition: }\\ $S_{i,t_s:t_p} := \sqrt{\left(1+\dfrac{1}{N_{i,t_s:t_p}}\right)\dfrac{\log(\sqrt{N_{i,t_s:t_p}+1}/\delta)}{2N_{i,t_s:t_p}}}$.
\For{$i=1,..,K$}
\For{$t' = t_s ,..,t_p$}
\If{$\big(\hat{\mu}_{i,t_s:t'} + S_{i,t_s:t'} < \hat{\mu}_{i,t'+1:t_p} - S_{i,t'+1:t_p}\big)$  or $\big(\hat{\mu}_{i,t_s:t'} - S_{i,t_s:t'} > \hat{\mu}_{i,t'+1:t_p} + S_{i,t'+1:t_p}\big)$}
\State Return True
%\Else{$\hat{r}_{i,t_s , t'} - \sqrt{\dfrac{(n_{i,t_s:t'}+1)\log(\frac{(n_{i,t_s:t'}+1)}{\sqrt{\delta}})}{2n_{i,t_s:t'}^2}} > \hat{r}_{i,t'+1:t_p} + \sqrt{\dfrac{(n_{i,t'+1:t}+1)\log(\frac{(n_{i,t'+1:t}+1)}{\sqrt{\delta}})}{2n_{i,t'+1:t}^2}}$}
%\State Return True
\EndIf
\EndFor
\EndFor
\end{algorithmic}     
\end{algorithm}   
%\end{minipage}

%We introduce the phase-based \ImpCPD in Algorithm \ref{alg:ImpCPD} in Appendix \ref{app:Algo}. \ImpCPD calls upon the changepoint detector CPDI only at the end of phases and saves upon computation time without incurring additional regret. %\ImpCPD employs pseudo arm elimination like CCB \citep{liu2016modification} algorithm such that a sub-optimal arm $i$ is never actually eliminated but the active list $B_m$ is just modified to control the phase length. This helps \ImpCPD adapt quickly because this is a global changepoint scenario and for some sub-optimal arm the changepoint maybe detected very fast. Another important divergence from UCB-Improved is the exploration parameter $0<\gamma\leq 1$ that controls how often the changepoint detection sub-routine CPDI is called. After every phase, $\epsilon_m$ is reduced by a factor of $\left( 1+\gamma\right)$ instead of halving it (as like UCB-Improved) so that the number of pulls allocated for exploration to each arm is lesser than UCB-Improved. The CPDI sub-routine at the end of the $m-th$ phase scans statistics so that if there is a significant difference between the sample mean of any two slices then it raises an alarm. 

Next, we formally introduce the Improved Changepoint Detector (\ImpCPD) in Algorithm \ref{alg:ImpCPD}. This is a phase-based algorithm modeled on UCB-Improved \citep{auer2010ucb}, which calls upon the Changepoint Detection Improved (CPDI) in Algorithm \ref{alg:CPDI} for calculating scan statistics to actively detect changepoint. \ImpCPD employs pseudo arm elimination like CCB \citep{liu2016modification} algorithm such that a sub-optimal arm $i$ is never actually eliminated but the active list $B_m$ is just modified to control the phase length. This helps \ImpCPD adapt quickly because this is a global changepoint scenario and for some sub-optimal arm the changepoint maybe detected very fast. Another important divergence from UCB-Improved is the exploration parameter $0<\gamma\leq 1$ that controls how often the changepoint detection sub-routine CPDI is called. After every phase, $\epsilon_m$ is reduced by a factor of $\left( 1+\gamma\right)$ instead of halving it (as like UCB-Improved) so that the number of pulls allocated for exploration to each arm is lesser than UCB-Improved. The CPDI sub-routine at the end of the $m$-th phase scans statistics so that if there is a significant difference between the sample mean of any two slices then it raises an alarm. 

\begin{algorithm}
 \caption{Improved Changepoint Detector (\ImpCPD)}
\label{alg:ImpCPD}
\begin{algorithmic}[1]
\State {\bf Input:} Time horizon $T$, $0<\gamma\leq 1$.
\State {\bf Initialization:} $B_{0}:=\mathcal{A}$, $m:=0$, $\epsilon_{0}:=1$, $\psi :=\frac{T^2}{K^2\log K}$, $\alpha := \frac{3}{2}$.
\begin{align*}
 M := \left\lfloor \frac{1}{2}\log_{1+\gamma} \frac{T}{e}\right\rfloor, \hspace*{2mm}
\ell_{0} :=\left\lceil \frac{\log(\psi \epsilon_{0}^2)}{2\epsilon_{0}} \right\rceil,
 L_{0} :=K\ell_{0}\text{, }t_s :=1\text{, }t_p :=1.
\end{align*}
\State {\bf Definition:} $S_{i,t_s:t_p}:=\sqrt{\dfrac{\alpha\log(\psi \epsilon_m^2)}{2N_{i,t_s:t_p}}}$.
\For{$t=K+1,..,T$}
\State Pull arm $j\in\argmax_{i\in\A}\bigg\lbrace \hat{\mu}_{i,t_s:t_p} + S_{i,t_s:t_p}\bigg\rbrace$, observe reward $X_{j,t}$.
\State $t:= t+1, t_p := t_p + 1$. 
\If{$t\geq L_{m}$ and $m \leq M$}
\If{CPDI($m,L_m,\alpha,\psi$)}
\State \textbf{Restart:} $B_{m}:=\A$; $m:=0$. 
%\State $B_{m}=\A$; $m=0$; 
\State Set $N_{i,t_s:t_p} := 0, \hat{\mu}_{i,t_s:t_p}:=0, \forall i\in\A$.
%\State Set $\hat{\mu}_{i,t_s:t_p}:=0,\forall i\in\A$.
\State  $t_s := t$, $t_p := t_s$.
\State Pull each arm once.
\Else{
\For{$i\in\A$}
\If{$\hat{\mu}_{i,t_s:t_p} + S_{i,t_s:t_p}  < \max_{j\in\A}\lbrace\hat{\mu}_{j,t_s:t_p} - S_{j,t_s:t_p}\rbrace$}
\State $|B_m| := |B_m|-1$
\EndIf
\EndFor
\State \textbf{Reset Parameters: } 
\State $\epsilon_{m+1}:=\frac{\epsilon_{m}}{(1+\gamma)}$, $B_{m+1}:= B_{m}$.
\State $\ell_{m+1}:=\left\lceil \dfrac{\log(\psi \epsilon_{m}^2)}{2\epsilon_{m}} \right\rceil$.
\State $L_{m+1}:= t + |B_{m+1}|\ell_{m+1}$.
\State $m := m+1.$}
\EndIf
\EndIf
\EndFor
\end{algorithmic}
\end{algorithm}

\begin{algorithm}
\caption{Changepoint Detection Improved($m,L_m,\alpha,\psi$) (CPDI)}
\label{alg:CPDI}
\begin{algorithmic}[1]
\State \textbf{Definition:}$S_{i,t_s:L_{m}}=\sqrt{\dfrac{\alpha\log(\psi \epsilon_m^2)}{2N_{i,t_s:L_{m}}}}$.
\For{$i=1,..,K$}
\For{$m' = 1 ,\ldots,m$}
\If{$\big(\hat{\mu}_{i,t_s:L_{m'}}  + S_{i,t_s:L_{m'}} 
 < \hat{\mu}_{i,L_{m'}+1:L_{m}} - S_{i,L_{m'+1}:L_{m}}\big)$ or $\big(\hat{\mu}_{i,t_s:L_{m'}} - S_{i,t_s:L_{m'}} > \hat{\mu}_{i,L_{m'}+1:L_{m}} + S_{i,L_{m'+1}:L_{m}}\big)$}
\State Return True
\EndIf
\EndFor
\EndFor
\end{algorithmic}
\end{algorithm}

\textbf{Running time of algorithms:} \UCBLCPD (like CD-UCB, CUSUM)  calls the changepoint detection at every timestep, and \ImpCPD calls upon the sub-routine only at end of phases. Hence, for a horizon $T$, $K$ arms, \UCBLCPD calls the changepoint detection subroutine $O(KT)$ times (memory requirements may scale as $O(KT^2/G)$) while \ImpCPD calls the changepoint detection $O(K\!\log T )$ times, thereby substantially reducing the costly operation of calculating the changepoint detection statistics. By designing \ImpCPD appropriately and modifying the confidence interval, this reduction comes at no additional cost in the order of regret (see Discussion \ref{dis:Corollary:1} and Expt-5).

\section{Main Results}
\label{psbandit:results}
We outline the main ideas to prove Theorem \ref{psbandit:Theorem:1} and \ref{psbandit:Theorem:2}. In Lemma \ref{psbandit:Lemma:1} we consider a single changepoint $g\in\G$ and a single arm $i\in\A^{chg}_g$ scenario where policy $\pi$ has observations from $t_s$ to $t$ for $t_{g-1} \leq t_s < t_g <t$ and $\Delta_{i,g}^{chg}$ is $\delta$-optimal. With the term $S_{i,t_s:t'}$ as defined in equation \eqref{eq:laplace}, we define the bad event $\textcolor{blue}{\xi^{chg}_{i,t}} = \!\big\lbrace \!\exists t'\!\!\in\! [t_s , t]\!: \big(\hat{\mu}_{i,t_s:t'} \!-\! S_{i,t_s:t'} \!>\! \hat{\mu}_{i,t'\!+\!1:t} \!+\! S_{i,t'\!+\!1:t}\big) \bigcup
%%%%%%%%%%%%%%%%%%%%%%%%%%%%
  \big(\hat{\mu}_{i,t_s:t'} \!+\!  S_{i,t_s:t'} \!<\! \hat{\mu}_{i,t'\!+\!1:t} \!-\! S_{i,t'\!+\!1:t} \big)\!\big\rbrace $ which will lead to failure of detection of changepoint.  Let $X_{i,t_s}, X_{i,t_s+1}, \ldots, X_{i,t'}$ have their expectation as $\mu_{i,g-1}$ and $X_{i,t'+1}, X_{i,t'+2}, \ldots , X_{i,t}$ have their expectation as $\mu_{i,g}$. Then to tackle $|\mu_{i,g} - \mu_{i,g-1}|$, we study the concentration of $\hat{\mu}_{i,t_s:t'}$ and $\hat{\mu}_{i,t'+1:t}$ around $\mu_{i,g-1}$ and $\mu_{i,g}$ separately using Laplace method and combine them by using a simple union bound.  These two quantities can also be jointly studied and yielding a robust concentration inequality as in \citep{maillard2019sequential}. We define the optimality bad event $\xi^{opt}_{i,g}$ as when the arm $i^\ast_{t'}$ for $t' \in [t_{g-1}, t_g - 1]$ have not been detected.
\begin{customlemma}{3} \textbf{(Control of $\xi^{chg}_{i,t}$ by Laplace method)}
\label{psbandit:Lemma:1}
Let, $N_{i,t_s:t}$ be the number of times an arm $i$ is pulled from $t_s$ till the $t$-th timestep such that $t_{g-1} \leq t_s < t_{g} < t$, then at the $t$-th timestep for all $\delta\in (0,\frac{1}{4}]$  it holds that,
%\begin{align*}
$\Pb(\xi^{chg}_{i,t}) \leq 4\delta $, 
%\end{align*}
where the event $\xi^{chg}_{i,t}$ is defined above.
 %= \sqrt{\left(1 + \frac{1}{N_{i,t_s:t'}}\right)\frac{\log(\sqrt{N_{i,t_s:t'} + 1}/\delta)}{2N_{i,t_s:t'}}}$.
 %and $S_{i,t_s:t'}$ is from \eqref{eq:laplace}
\end{customlemma}
\begin{customproof}{2} \textbf{(Outline)} We use the sub-Gaussian property of the bounded random variables to define a non-negative super-martingale $M_{t'}^{\lambda}$. We show that it is well defined and introduce a new stopped version  $M_{\tau}^{\lambda}$. By Fatou's Lemma we show that it is bounded as well. Finally, we introduce an auxiliary variable $\Lambda$ independent of all other variables and use it to control $M_{\tau}^{\Lambda}$. We follow a similar procedure  and define another non-negative super-martingale $M_{t-t'}^{\lambda}$ and combine the result to bound the probability of the event $\xi^{chg}_{i,t}$. The proof is in Appendix \ref{sec:proof:Lemma:1}.
\end{customproof}
%We use Markov's inequality to bound the probability of the event $\xi^{chg}_{i,t}$ using $M_{\tau}^{\Lambda}$. 
%/*write how different from others*/
%refers to the minimum number of pulls required before either \UCBLCPD discards the sub-optimal arm $i$ or a changepoint is detected between $\tau_{i,g-1}:t_{g}$ which is given by $\max\{N_{i,\min_D}, N_{\min_E}$. \textbf{part B} refers to the number of pulls due to bad event that \UCBLCPD continues to pull the sub-optimal arm $i$ or the changepoint is not detected between $\tau_{i,g-1}:t_{g}$ (the event $\xi^{opt}_{i,g}$, $\xi^{chg}_{i,g}$) which is controlled by Lemma \ref{psbandit:Lemma:1}. \textbf{part C} refers to the pulls accrued due to the worst case events when the changepoint has occurred and has not been detected by \UCBLCPD from $t_{g}:\tau_{i,g}$. In this case, we cannot do anything but suffer the worst possible regret till the changepoint is detected due to another arm (by assumption \ref{assm:chg-gap}) or a new changepoint occurs. \textbf{part D} refers to number of pulls for the bad event that the changepoint is not detected from $t_{g}:\tau_{i,g}$, and finally \textbf{part E} refers to the case when the changepoint $g$ is not detected from time $\tau_{g-1} > t_{g-1}$ till time $t$.
\begin{discussion}
\label{Disc:Laplace}
Time-uniform bound in \eqref{eq:laplace} depends only on the number of pulls $N_{i,t_s:t_p}$ and implicitly on $t$ based on the parameter $\delta$. The concentration bounds based on peeling and union bound method depends explicitly on $t$ and has larger coefficients attached to them. In \textcolor{blue}{Table \ref{table:compCPD}} we give a comparison over the three concentration bound method involving union bound, peeling and \textcolor{blue}{Laplace method} and compare them empirically in experiment $5$. We provide the proof of the construction of concentration bound for our changepoint detection strategy by union bound in Lemma \ref{psbandit:Lemma:11} in Appendix \ref{sec:proof:Lemma:11} for completeness. The $\log\log (t)$ scaling of the peeling method is not better than the one derived by the Laplace method, unless for huge timestep $t$ ($t > 10^6$, for $\delta = 0.05$ and any $\alpha > 1$). Laplace method uses the sub-Gaussian nature of the variables to give such sharp concentration bounds as opposed to other methods. 

\begin{table}
\begin{center}
\caption{Comparison of Union, Peeling and Laplace method}
\label{table:compCPD}
\begin{tabular}{|p{2.7em}|p{14.5em}|p{2.7em}|}
\toprule
Method & Confidence interval &Uniform over $t$\\
\hline
Union & $\sqrt{\frac{\log(4t^2/\delta)}{2N_{i,t_s:t'}}}$  & No\\
Peeling & $\sqrt{\frac{\alpha}{N_{i,t_s:t'}}\log\left(\lceil\frac{\log(t)}{\alpha}\rceil\frac{1}{\delta}\right))}$, $\alpha > 1$  & No\\
Laplace & $\sqrt{\left(1 + \frac{1}{N_{i,t_s:t'}}\right)\frac{\log(\sqrt{N_{i,t_s:t'} + 1}/\delta)}{2N_{i,t_s:t'}}}$ & Yes\\
\midrule
\end{tabular}
\vspace*{-2.0em}
\end{center}
\end{table}

\end{discussion}
%identical to Eq (\ref{event:out:1})
\begin{customtheorem}{1}\textbf{(Gap-dependent bound of \UCBLCPD)}
\label{psbandit:Theorem:1}
For $\eta \geq \frac{6}{2\log t + 1}$, $\delta=\frac{1}{t}$, the expected cumulative regret of \UCBLCPD using the CPD is given by,
%(Algorithm \ref{alg:UCBCPD}) (Algorithm \ref{alg:CPD})
\begin{align*}
 \E_{\pi}[R_t]  \!\!&\leq \!\sum_{g=1}^{G} \!\bigg\lbrace\!\! \underbrace{\sum_{i=1}^{K}\dfrac{6\log t}{\Delta^{opt}_{i,g}}}_{(a)} \! +\!\!\sum_{i\in \A_g^{chg}}\!\!\bigg( \!\!\underbrace{\dfrac{16H_{2,g}\log t}{\Delta^{chg}_{i,g}}}_{(b)} 
 \! +\! \underbrace{\dfrac{30K H_{2,g}\log t}{\Delta(t_g, \delta)}}_{(c)} \!\!\bigg)\!\!\bigg\rbrace \! + \!\underbrace{\! \max_{i\in\A: \!\sqrt{\frac{e}{T}}\leq \! \Delta_i <\! \Delta(t_g,\delta)}\Delta_i t\!}_{(d)}.
\end{align*}
%%\E[R_t]  &\leq \sum_{i=1}^{K}\sum_{g=1}^{G} \bigg\lbrace 3 + \dfrac{8\log(t)}{\Delta^{opt}_{i,g}} + \dfrac{\pi^2}{3} + \dfrac{6\Delta^{opt}_{i,g}\log(t)}{(\Delta^{chg}_{i,g})^2} \\
%%%%%%%%%%%%%%%%%%%%%%%%%%%%
%& + \dfrac{6\Delta^{opt}_{\max,g+1}\log(t)}{(\Delta^{}_{\epsilon_0, g})^{2}} + \dfrac{12\Delta^{opt}_{i,g+1}\log(t)}{(\Delta^{chg}_{i,g})^{2}}\bigg\rbrace.
\end{customtheorem}
\begin{customproof}{3} \textbf{(Outline)}
Let $\tau_{i,g}$ be the stopping time for $i$ and $\tau_{g} > t_g$ be the time $g$ is detected. We decompose the number of pulls $N_{i,1:t}$ into five parts (see eq \eqref{eq:pull}, Appendix \ref{sec:proof:Theorem:1}). In \textbf{part A} we show that there exists a deterministic number of steps $\max\{N_{i,\min_E}, N_{i,\min_D}\}$ before either \UCBLCPD discards sub-optimal arm $i$ or a false detection of changepoint happens between $\tau_{i,g-1}:t_{g}$. \textbf{Part B} handles the complementary event of part A by bounding the bad events $\xi^{opt}_{i,t}$ and $\xi^{chg}_{i,t}$ using Lemma \ref{psbandit:Lemma:1}  to control false detection of $t_g$. In \textbf{part C} we control the pulls accrued due to the worst case events that the changepoint has occurred and $i$ has not been sampled enough to detect changepoint ($N_{i,t_g:\tau_{i,g+1}}< N_{i,min_D}$) or is an undetectable gap ($\sqrt{\frac{e}{T}} \leq \Delta^{chg}_{i,g} < \Delta(t_g,\delta)$). In this case, $\pi$ has to suffer the worst possible regret till a $\delta$-optimal changepoint is detected due to another arm (assumption \ref{assm:chg-gap}) or a new non-isolated changepoint occurs (assumption \ref{assm:space-gap}). \textbf{Part D} controls the bad event that the changepoint is not detected from $t_{g}:\tau_{i,g}$ even when $N_{i,t_g:\tau_{i,g}} > N_{i,\min_D}$ using Lemma \ref{psbandit:Lemma:1}. Finally \textbf{part E} refers to the case when the changepoint $g$ is not detected from time $\tau_{g-1}$ till time $t$. We control it by showing that $d_{\pi}(t_g - \tau_{g-1}) = d_{\pi}(t_g - (t_{g-1} + d_{\pi}(t_{g-1} - \tau_{g-2})))$ is bounded with high probability as long as $\eta > 6/(2\log t + 1)$ and assumption \ref{assm:space-gap} holds for all $g\in\G$ (see \text{step 6}). We derive the value of $\eta$ from the base case. So, our approach is different than \citet{cao2018nearly}, \citet{liu2017change},  \citet{besson2019generalized} as part \textbf{C}, \textbf{D} and \textbf{E} requires careful handling of assumptions \ref{assm:space-gap} and \ref{assm:chg-gap} which the previous works do not take. The proof of Theorem \ref{psbandit:Theorem:1} is given in Appendix \ref{sec:proof:Theorem:1}. In Theorem \ref{psbandit:Theorem:1}, (a) is the regret suffered before finding the optimal arm between changepoints $g-1$ to $g$, (b) is the maximal regret for delayed detection of $g$, (c) is the regret suffered for total compounded delayed detection and $(d)$ for undetectable changepoints. 
\end{customproof}
\begin{customtheorem}{2}\textbf{(Gap-dependent bound of \ImpCPD)}
\label{psbandit:Theorem:2}
For $\eta \geq \frac{8}{2\log T + 1}$, $\delta=\frac{1}{T}$, the expected cumulative regret of \ImpCPD using CPDI is upper bounded by,
%(Algorithm \ref{alg:ImpCPD}) (Algorithm \ref{alg:CPDI})
\begin{align*}
& \E_{\pi}[R_T] \leq \sum_{g=1}^{G}\sum_{i\in\A'}\bigg[ \underbrace{ \frac{48 K C_1\left(\gamma \right)\Delta^{opt}_{i,g}\log({T}/{K} )}{(K\log K)^{-\frac{3}{2}}} }_{(a)}  + 
%%%%%%%%%%%%%%%%%%%%%%%%%%%
\!  \underbrace{ \frac{16\log({T (\Delta^{opt}_{i,g})^2}/{K})}{(\Delta^{opt}_{i,g})} }_{(b)} \! \bigg]  \\
%\end{align*}
% \vspace*{-0.8em}
%\begin{align*}
 &  \!+\! \sum_{g=1}^{G}\sum_{i\in\A^{chg}_g}\bigg[ \! \underbrace{\frac{16 H_{2,g}\log({T(\Delta^{chg}_{i,g})^2}/{K})}{(\Delta^{chg}_{i,g})} }_{(c)} +\! \underbrace{ \frac{16 K H_{2,g}\log({T(\Delta^{chg}_{i,g})^2}/{K})}{(\Delta(t_g,\delta))}}_{(d)}\bigg] \!\! +\underbrace{\! \max_{i\in\A: \!\sqrt{\frac{e}{T}}\leq \! \Delta_i <\! \Delta(t_g,\delta)}\Delta_i T\!}_{(e)}
 \vspace*{-1em}
\end{align*}
where $\gamma \in (0,1]$ is exploration parameter, $C_1\left( \gamma\right)=\left( \frac{1+\gamma}{\gamma}\right)^{4}$, and $\A'=\big\lbrace i\in\A: \Delta^{opt}_{i,g}\geq \sqrt{\frac{e}{T}}, \Delta^{chg}_{i,g}\geq \sqrt{\frac{e}{T}},\forall g\in\G \big\rbrace$.
\end{customtheorem}
\vspace*{-0.5em}
%and $\Delta^{opt}_{i,{G+1}} = 0,\forall i\in\A$.
\begin{customproof}{4} \textbf{(Outline)}
This proof closely follows the approach of Theorem \ref{psbandit:Theorem:1}.  We carefully construct each geometrically increasing phase length $\ell_m$ so that the probability not pulling the optimal arm between two changepoints $g-1$ to $g$ is bounded. Simultaneously, we use the phase length $\ell_m$, confidence interval $S_{i,t_s:t_p}$ and exploration factor $\gamma$ and $\psi$ to control the bad event of not detecting the changepoint $g$. We use \textcolor{blue}{Chernoff-Hoeffding inequality} to bound the probability of the bad events. We have to further balance $\ell_{m}$, and $S_{i,t_s:t_p}$ by carefully defining $\psi$ so that $\gamma$ is small enough and CPDI is called more often. We have to use additional union bounds to control the event that arms are getting pulled unequal number of times within each phase length. The proof is in Appendix \ref{sec:proof:Theorem:2}. In Theorem \ref{psbandit:Theorem:2}, (a) is the regret for calling the CPDI at end of phases, (b) is the regret for finding the optimal arm between changepoints $g-1$ and $g$, (c) is the regret for delayed detection of $g$, (d) is the regret suffered for total compounded delayed detection and $(e)$ for undetectable changepoints. 
%We divide the proof into two modules. In the first module, we bound the optimality regret of not pulling the optimal arm between two changepoints $g-1$ to $g$ using steps $3,4$ and $5$. In the second module we bound the changepoint regret incurred for not detecting the $g$-th changepoint using steps $2,6,7$ and $8$. We use Chernoff-Hoeffding inequality to bound the probability of the bad events. We control the number of pulls of each sub-optimal arm using the definition of $\ell_{m_i}$ and exploration parameter $\gamma$. The proof of Theorem \ref{psbandit:Theorem:2} is given in Appendix \ref{sec:proof:Theorem:2}.
\end{customproof}

%\begin{discussion}
%\label{dis:Theorem:1}
%In Theorem \ref{psbandit:Theorem:2}, (a) is the regret suffered for calling the CPDI only at end of phases, (b) is the regret for finding the optimal arm between changepoints $g-1$ and $g$, (c)  is the regret for delayed detection of $g$, and (d) is the regret suffered for total compounded delayed detection. 
%\end{discussion}

%\begin{discussion}
%\label{dis:Theorem:2}
%In Theorem \ref{psbandit:Theorem:2}, the largest contributing factor to the gap-dependent regret of  ImpCPD is of the order $O\big(\max\big\lbrace\sum\limits_{g=1}^G\dfrac{H_{3,g}\log( \frac{T(\Delta^{chg}_{\max,g})^2}{K\sqrt{\log K}})}{(\Delta^{\epsilon_0}_{g})}$ , $\sum\limits_{g=1}^G\dfrac{\log(\frac{T (\Delta^{opt}_{\max,g})^2}{K\sqrt{\log K}})}{(\Delta^{opt}_{i,g})}\big\rbrace\big)$, lower than that of CUSUM, DUCB, and SWUCB when $\forall i\in\A, \forall g\in\G$, $\Delta^{opt}_{i,g} = \Delta^{chg}_{i,g} = \Delta^{}_{\epsilon_0, g}$ (see Table \ref{tab:comp-bds}). 
%\end{discussion}

\begin{discussion}
\label{dis:delay:corollary}
%From Theorem \ref{psbandit:Theorem:1} and \ref{psbandit:Theorem:2} we see that
\UCBLCPD (Theorem \ref{psbandit:Theorem:1}) and \ImpCPD (Theorem \ref{psbandit:Theorem:2}) performance is comparable to the best detection strategy (see Lemma \ref{psbandit:Lemma:01}) as they have the coefficient in their compounded detection delay of order $O(K)$ that is less than order $O(\eta\log(t/\delta))$ of $C(t,\delta,\eta)$ of $\pi^\ast$ when $\eta$ is greater than the respective values in the theorems. This is reasonable as in the bandit setup each arm $i\in\A$ might be pulled a logarithmic number of times before detecting a changepoint. 
%Note, that the delay of \UCBLCPD is less than \ImpCPD.
\end{discussion}
%of order $O(1)$ associated with their delays that is less than $C_\eta$ of order $O(\log(T/G))$ (see Lemma \ref{psbandit:Lemma:01}).
\begin{customcorollary}{1}\textbf{(Gap-independent bounds)}
\label{psbandit:Corollary:1}
In the challenging scenario, when all the gaps are equal and small, i.e. $\!\Delta^{opt}_{i,g}\!=\!\Delta^{chg}_{i,g}\!=\!\Delta(t_g,\delta)=\!\!\sqrt{\frac{K\log (T/G)}{T/G}}\! > \!\Omega(\sqrt{\frac{\log t}{t}}), \forall i\!\!\in\!\!\A,\forall g\!\!\in\G$, $t\!>\!\!\sqrt{T}$, $\delta \! =\frac{1}{T}$, $\gamma \!= 0.05$, $\! G\geq \!\!{\log T}\!$ then worst case gap-independent regret bound of \UCBLCPD and \ImpCPD is,
%\begin{align*}
%\E[R_{T}] \leq 3KG + \dfrac{KG \pi^2}{3} + \dfrac{32\sqrt{KGT}\log T}{\sqrt{{\log K}}}.
%\end{align*}
\begin{align*}
\E_{\text{\UCBLCPD}}[R_{T}] \leq O(\sqrt{GT}\log T), \text{ \hspace*{2em} } \E_{\text{\ImpCPD}}[R_T]&\leq  C_1 G^{1.5}K^{4.5}(\log K)^{2} + O(\sqrt{GT}).
\end{align*}
\end{customcorollary}
\begin{customproof}{5}
The proof of Corollary \ref{psbandit:Corollary:1} follows from Theorem \ref{psbandit:Theorem:1} and \ref{psbandit:Theorem:2} and is given in Appendix \ref{sec:proof:Corollary:1}.
\end{customproof}

\begin{discussion}
\label{dis:Corollary:1}
In Corollary \ref{psbandit:Corollary:1}, the largest contributing factor to the gap-independent regret of \UCBLCPD is of the order $O(\sqrt{GT} \log T)$, same as that of DUCB but weaker than CUSUM and SWUCB. The additional $O(\log T)$ factor is the cost \UCBLCPD must pay for not knowing $T$ and $G$. The largest contributing factor to the gap-independent regret of \ImpCPD is of the order $O( \sqrt{GT})$. This is lower than the regret upper bound of DUCB, SWUCB, EXP3.R and CUSUM (\textcolor{blue}{Table \ref{tab:comp-bds}}). The smaller the value of the exploration parameter $\gamma$ the larger is the constant $C_1$ associated with the factor $GK^{4.5}(\log K)^{2}$. Now, $\gamma$ determines how frequently CPDI is called by \ImpCPD and by modifying the confidence interval and phase-length we have been able to control the probability of not detecting the changepoint at the cost of additional regret that only scales with $K$ and not with $T$.
%Also, the constant $C = 32$ scales at a lower rate than $C_\eta$ indicating that UCB-CPD detects each changepoint $g\in\G$ without endangering the detection of the next changepoint.
\end{discussion}

%\begin{customcorollary}{2}\textbf{(Gap-independent bound of \ImpCPD)}
%\label{psbandit:Corollary:2}
%In the specific scenario, when all the gaps are same, that is $\Delta^{opt}_{i,g}=\Delta^{chg}_{i,g}=\Delta(t_g,\delta)=\sqrt{\frac{K\log (T/G)}{T/G}}, \forall i\in\A,\forall g\in\G$ and setting $\delta = \frac{1}{T}, \gamma = 0.05$ then the worst case gap-independent regret bound of \ImpCPD is given by,
%%$\alpha=1.5$, $\psi = \frac{T}{K^2\log K}$ and
%\begin{align*}
%\E[R_T]&\leq  C_1 G^{1.5}K^{4.5}(\log K)^{2}
%%%%%%%%%%%%%%%%
% + O(\sqrt{GT})
%\end{align*}
%where $C_1$ is an integer constant.
%\end{customcorollary}
%
%\begin{customproof}{6}
%The proof of Corollary \ref{psbandit:Corollary:2} is given in Appendix \ref{sec:proof:Corollary:2}.
%\end{customproof}

%\begin{discussion}
%\label{dis:Corollary:2}
%In Corollary \ref{psbandit:Corollary:2}, the largest contributing factor to the gap-independent regret of \ImpCPD is of the order $O\left( \sqrt{GT}\right)$. This is lower than the regret upper bound of DUCB, SWUCB, EXP3.R and CUSUM (Table \ref{tab:comp-bds}). The smaller the value of the exploration parameter $\gamma$ the larger is the constant $C_1$ associated with the factor $GK^{4.5}(\log K)^{2}$. Now, $\gamma$ determines how frequently CPDI is called by \ImpCPD and by modifying the confidence interval and phase-length we have been able to control the probability of not detecting the changepoint at the cost of additional regret that only scales with $K$ and not with $T$.
%\end{discussion}
\begin{customtheorem}{3}\textbf{(Lower Bounds for $\pi^o$)}
\label{psbandit:Theorem:3}
The lower bound of an oracle policy $\pi^o$ for a horizon $T$, $K$ arms and $G$ changepoints is given by 
%\begin{align*}
$\!\E_{\pi^o}[R_T]\geq \min \lbrace\!\Omega (\sum\limits_{g=1}^G\sum\limits_{i=1}^{K}\!\frac{ \log{(T/(GH^{}_{1,g}))}}{(\Delta_{i,g}^{opt})}), \!\Omega(\sqrt{GT}\!)\rbrace$.
%\end{align*}
%where, $H^{}_{1,g} = \sum\limits_{i=1}^{K}{(\Delta^{opt}_{i,g})^{-2}}$ is the hardness of the problem.
\end{customtheorem}
\begin{customproof}{7}
An oracle policy $\pi^o$ has access to the exact changpoints. The worst case scenario can occur when  environment changes uniform randomly. A similar argument has also been made in the adaptive-bandit setting of \citep{DBLP:journals/jmlr/MaillardM11}. So, let the horizon $T$ be divided into $G$ slices, each of length $(T/G)$. For each of these slices an oracle algorithm using OCUCB \citep{lattimore2015optimally} should get the optimal SMAB regret without suffering any delay. The proof is in Appendix \ref{proof:Theorem:3}.
\end{customproof}
\begin{discussion}
\label{dis:Theorem:3}
This lower bound is weaker than the bound proposed in \citet{wei2016tracking} as they do not require the knowledge of $G$ or $T$. But we provide this for completion to discuss oracle-based bounds and also because the previous approaches do not touch upon this approach. For a non-oracle policy the additional trade-off between the changepoint gap and the next optimality gap is captured by $H_{2,g}$. As long as the delayed detection is bounded with high probability we should get a similar scaling for a good detection algorithm minimizing regret for each of these slices of length $(T/G)$. \ImpCPD which has a gap-independent regret upper bound of $O(\sqrt{GT})$ reaches the lower bound of the  policy $\pi^o$ in an \textcolor{blue}{order optimal} sense.  In the challenging case when all the gaps are same and small such that for all $i\in\A,g\in\G$, $\Delta^{opt}_{i,g} = \Delta(t_g,\delta) = \Delta^{chg}_{i,g}$, $H^{}_{1,g} = K(\Delta(t_g,\delta))^{-2}$ and $H_{2,g}=1$, then ImpCPD with a gap-dependent bound of $O\big(\sum\limits_{g=1}^G\sum\limits_{i=1}^{K}\frac{\log({T}/{H^{}_{1,g}})}{\Delta_{i,g}^{opt}}\big)$ matches the gap dependent lower bound of $\pi^o$ except the factor $G$ in the log term (see \textcolor{blue}{Table \ref{tab:comp-bds}}).
% except the factor $G$ in the log term (ignoring the $\log(\frac{1}{\sqrt{\log K}})$ term).
\end{discussion}

%\begin{customproposition}{1}\textbf{(Lower Bounds for OUCB1)}
%\label{psbandit:Prop:1}
%The regret upper bound of OUCB1 for a horizon $T$ and $G$ changepoints is given by,
%\begin{align*}
%\E[R_T] \leq C_1\sqrt{GKT \log{\frac{T}{G}}}
%\end{align*}
%where, $C_1$ is an integer constant.
%\end{customproposition}
%
%\begin{customproof}{8}
%The proof of Proposition \ref{psbandit:Prop:1} is given in Appendix \ref{proof:Proposition:1}.
%\end{customproof}
%
%\begin{discussion}
%\label{dis:Prop:1}
%Hence, CUSUM and OUCB1 has the same regret upper bound of $O(\sqrt{GT \log{\frac{T}{G}}})$.
%\end{discussion}

\section{Experiments}
\label{psbandit:expt}
%In this section, we present numerical simulations to test the performance of 
We compare \UCBLCPD and \ImpCPD against Oracle Thompson Sampling (OTS), EXP3.R, Discounted Thompson Sampling (DTS), Discounted UCB (DUCB), Sliding Window UCB (SWUCB), Monitored-UCB (M-UCB) and CUSUM-UCB (CUSUM) in four environments . The oracle algorithm have access to the exact changepoints and are restarted at those changepoints. 

\subsection{Parameter Selection for Algorithms}
\label{appendix:param}
For DTS we use the discount factor $\gamma=0.75$ as specified in \citep{raj2017taming} for slow varying environment. For DUCB we set $\gamma=1-\frac{1}{4}\sqrt{\frac{1}{T}}$ and for SWUCB we use the window size of $W=4\sqrt{T\log T}$ as suggested in \citet{garivier2011upper}.  Note, that for implementing the two passive algorithms we do not use the knowledge of the number of changepoints $G$ to make these algorithms more robust, even though \citet{garivier2011upper} suggests using $\gamma=1-\frac{1}{4}\sqrt{\frac{G}{T}}$ or $W=4\sqrt{\frac{T\log T}{G}}$ for optimal performance. For CUSUM, we need to tune four parameters, $M$ which is the minimum number of pulls required for each arm which must be first satisfied, parameter $\alpha$ which determines random uniform exploration between arms, and the CUSUM changepoint detection parameter $h$ and tolerance factor $\epsilon$. We choose $M=40$, $\alpha =\sqrt{\frac{G}{T}\log\left(\frac{T}{G}\right)}$, $\epsilon=0.1$ and $h=\log\left(\frac{T}{G}\right)$ (as suggested in \citet{liu2017change}) for both the experiments. Hence, CUSUM requires the  knowledge of the number of changepoints in tuning its parameters. Similarly, Exp3.R requires tuning of three parameters $\delta$, $\gamma$ and $H$ for changepoint detection. We choose $\delta =\sqrt{\frac{\log T}{KT}}$, $\gamma =\sqrt{\frac{K \log K \log T}{T}}$ and $H=\sqrt{T\log T}$ as suggested in \citet{allesiardo2017non}. Finally, for \UCBLCPD we choose $\delta=\frac{1}{t}$ (Theorem \ref{psbandit:Theorem:1}) and for and \ImpCPD we choose $\delta=\frac{1}{T}$, $\gamma = 0.05$ (Corollary \ref{psbandit:Corollary:1}).

\subsection{Numerical Simulations}
\begin{figure}
\centering
\vspace*{-0.8em}
\begin{tabular}{cc}
\setlength{\tabcolsep}{0.1pt}
\subfigure[0.25\textwidth][Expt-$1$: $3$ Bernoulli-distributed arms.]
    %with $r_{i_{{i}\neq {*}}}=0.07$ and $r^{*}=0.1$
    {
    \includegraphics[scale=0.17]{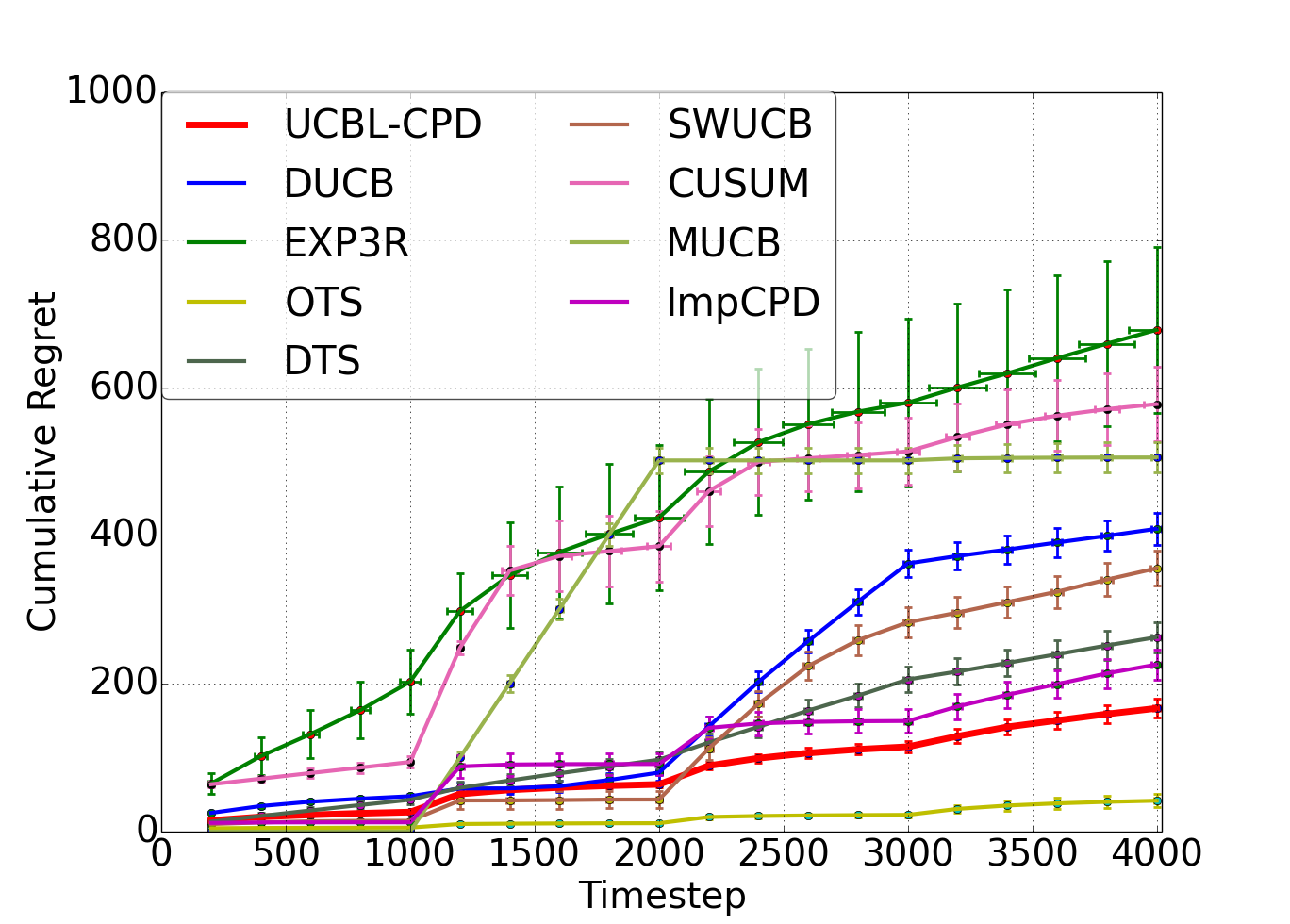}
    \label{psbandit:fig:1}
    	%\label{fig:5}
    }
\subfigure[0.25\textwidth][Expt-$2$: $20$ Users, $100$ items, Rank $3$ approximation of Jester Dataset]
    %with $r_{i_{{i}\neq {*}}}=0.07$ and $r^{*}=0.1$
    {
    \includegraphics[scale=0.17]{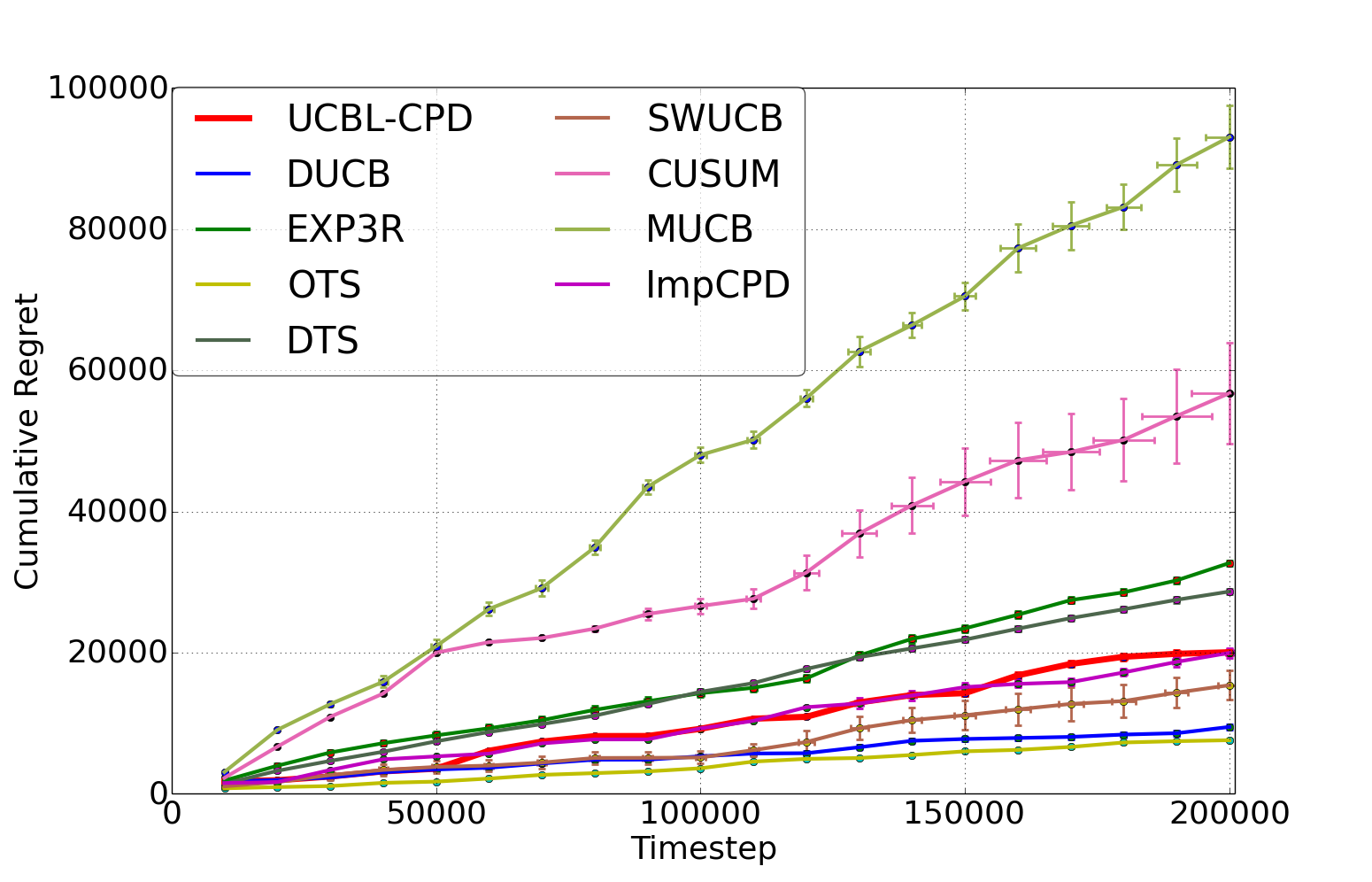}
  		\label{fig:6}
    }
 \end{tabular}
 \vspace*{-1em}
    \caption{Cumulative regret for various bandit algorithms on two  piecewise i.i.d environments}
    \label{fig:karmed11}
    \vspace*{-1em}
\end{figure}

\textbf{Experiment 1 (Bernoulli $3$ arms):} This experiment is conducted to test the performance of algorithms in Bernoulli distribution over a short horizon $T=4000$ and and small number of arms $K=3$. There are $3$ changepoints in this testbed and the expected mean of the arms changes as shown in equation \eqref{eq:expt:1} ($\eta < 0.5$). From the Figure \ref{psbandit:fig:1} we can clearly see that \UCBLCPD and \ImpCPD detect the changepoints at $t=1000$ and $t=2000$ with a small delay and restarts. However, because of the small changepoint gap at $t=3000$ it takes some time to adapt and restart. \UCBLCPD and \ImpCPD perform better than all the passively adaptive algorithms like DTS, DUCB, SWUCB, and actively adaptive algorithm like EXP3.R, CUSUM and is only outperformed by OUCB1 and OTS which have access to the oracle. The performance of \UCBLCPD is similar to \ImpCPD in this small testbed. Because of the short horizon and a small number of arms, the adaptive algorithms CUSUM and EXP3.R are outperformed by passive algorithms DUCB, SWUCB, and DTS.
\noindent
%\begin{figure}
    \begin{eqnarray}
 \hspace*{-2em }     && r_1 = 0.1, r_2 = 0.2 , r_3 = 0.9 \textbf{, if } t=[1,1000]; \nonumber\\
      && r_1 = 0.4, r_2 = 0.9 , r_3 = 0.1 \textbf{, if } t=[1001,2000];  \nonumber\\
\hspace*{-2em }     && r_1 = 0.5, r_2 = 0.1 , r_3 = 0.2 \textbf{, if } t=[2001,3000]; \nonumber\\
      && r_1 = 0.2, r_2 = 0.2 , r_3 = 0.3 \textbf{, if } t=[3001,4000]. \label{eq:expt:1}
\end{eqnarray}

\textbf{Experiment 2 (Jester dataset):} This experiment is conducted to test the performance of algorithms when our model assumptions are violated. We evaluate on the Jester dataset \citep{goldberg2001eigentaste} which consist of over 4.1 million continuous ratings of 100 jokes from 73,421 users collected over 5 years. We use Jester because there exist a high number of users who have rated all the jokes, and so we do not have to use any matrix completion algorithms to fill the rating matrix. The goal of the learner is to suggest the best joke when a new user comes to the system. We consider $20$ users who have rated all the $100$ jokes and use SVD to get a low rank approximation of this rating matrix. Most of the users belong to three classes who prefer either joke number $49$, $88$, or $90$. We uniform randomly sample $2$ users from each of the $3$ classes ($49$, $88$, $90$). Then we divide the horizon $T = 200000$ into $6$ changepoints starting from $t = 1$  and at an interval of $25000$ we introduce a new user from one of the three classes in round-robin fashion starting from users who prefer joke $49$. We change the user at changepoints to simulate the change of distributions of arms and hence a single learning algorithm has to adapt multiple times to learn the best joke for each user. A real-life motivation of doing this may stem from the fact that running an independent bandit algorithm for each user is a costly affair and when users are coming uniform randomly a single algorithm may learn quicker across users if all the users prefer a few common items. Note, that we violate assumption \ref{assm:space-gap} and \ref{assm:chg-gap} because the horizon is small, the number of arms is large and gaps are too small to be detectable with sufficient delay. In Figure \ref{fig:6} we see that \ImpCPD and  \UCBLCPD outperforms all other actively adaptive algorithms. \ImpCPD and \UCBLCPD is only able to detect $2$ of the $6$ changepoints and restart while CUSUM  failed to detect any of the changepoints. Note that \UCBLCPD and \ImpCPD performs slightly worse than SWUCB, DUCB in this testbed. This shows that when gaps are small, and changepoints are less separated, \textit{all} the change-point detection techniques will perform badly in those regimes. 

\begin{figure}
    \centering
    \begin{tabular}{cc}
    \hspace*{-3em}
    \subfigure[\textwidth][ Expt-$3$: $10$ Gaussian-distributed arms.]
    %with $r_{i_{{i}\neq {*}}}=0.07$ and $r^{*}=0.1$
    {
    \includegraphics[scale=0.18]{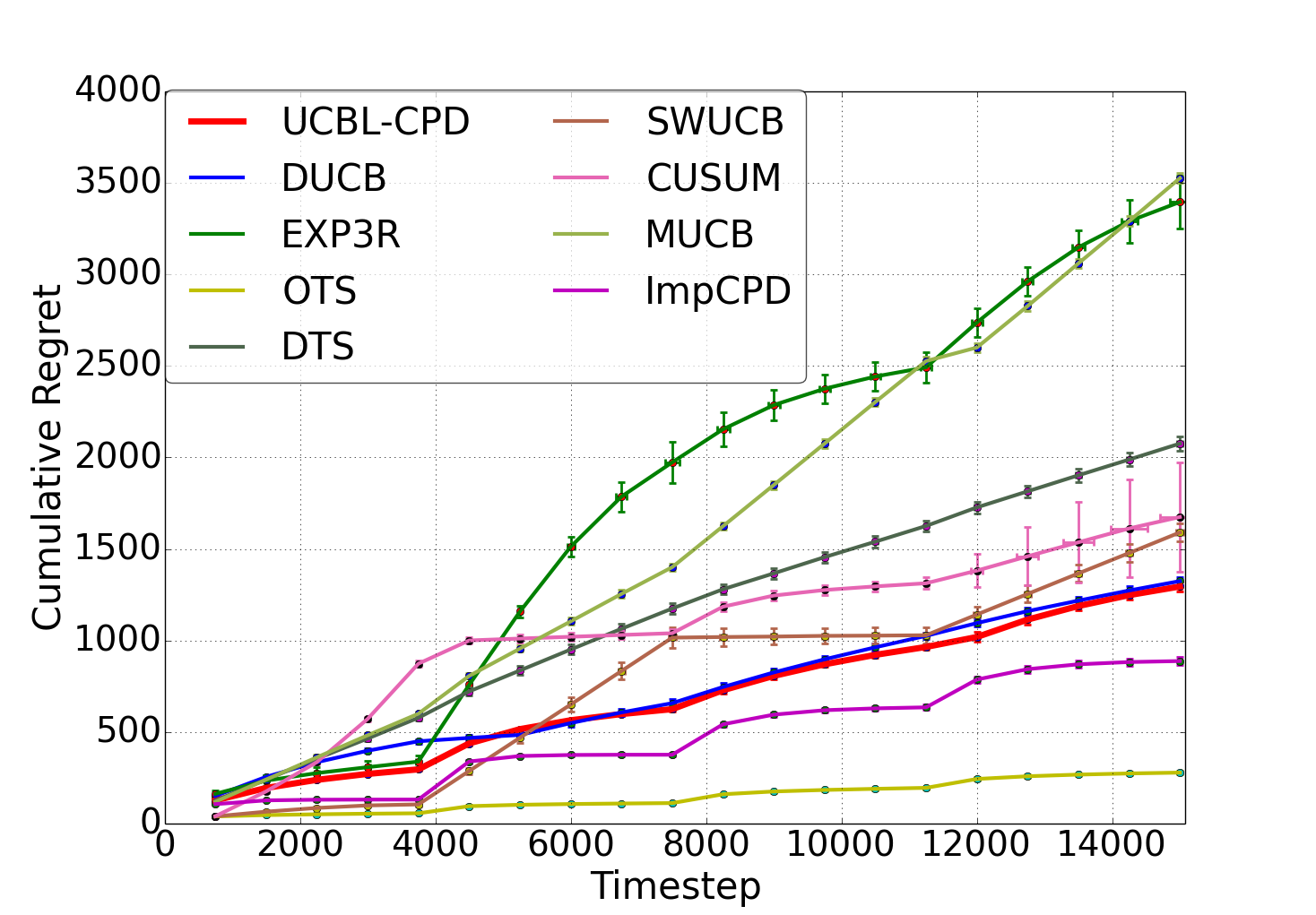}
          \label{psbandit:fig:2}
    }
    &
    \hspace*{-3em}
    \subfigure[\textwidth][ Expt-$4$: $3$ Bernoulli-distributed arms.]
    %with $r_{i_{{i}\neq {*}}}=0.07$ and $r^{*}=0.1$
    {
    \includegraphics[scale=0.17]{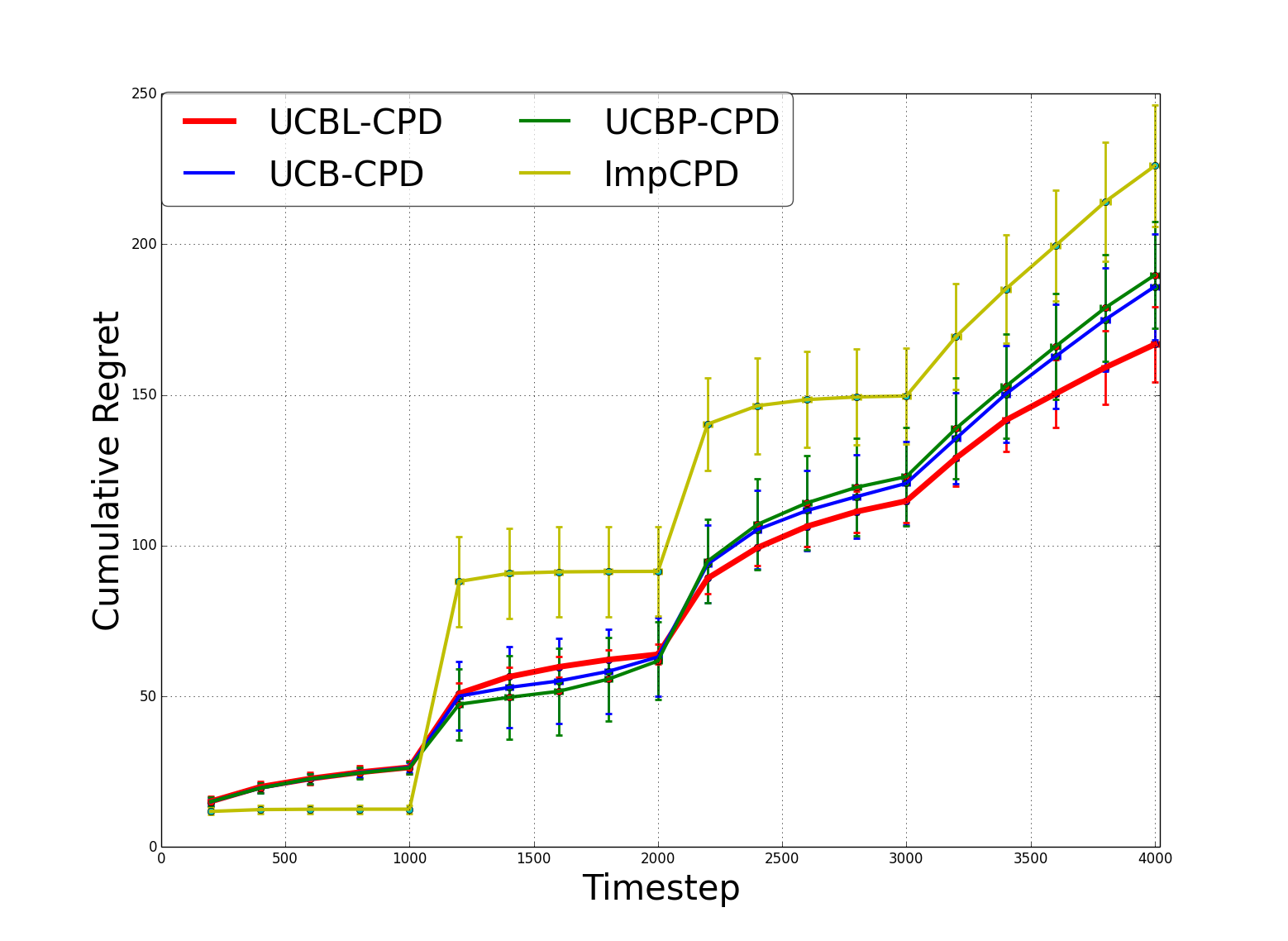}
          \label{psbandit:fig:21}
    }
\end{tabular}
    \caption{Additional Experiments 1.}
    \label{fig:karmed3}
\end{figure}

\textbf{Experiment 3 (Gaussian $10$ arms):} This experiment is conducted to test the performance of algorithms in Gaussian distribution where the distribution flips between two alternating environments. This experiment involves $10$ arms with Gaussian distribution. There are $3$ changepoints in this testbed where the horizon is $T=15000$ and the expected mean of the arms changes as shown in equation \ref{eq:expt:2}. The variance of all arms $i\in\A$ is set as $\sigma_{i}^2 = 0.25$ so that the sub-Gaussian distributions remain bounded in $[0,1]$ with high probability. The experiment is shown in Figure \ref{psbandit:fig:2} where we can clearly see that UCB-CPD and \ImpCPD detect all the changepoints at $t=1875$, $t=5001$ and $t=9001$ with a small delay and restarts. \ImpCPD clearly performs better than all the passively adaptive algorithms like  DUCB and DTS, actively adaptive algorithm like EXP3.R and is only outperformed by OTS which have access to the oracle. 
\begin{figure}[!th]
\begin{eqnarray}
      &r_{1:4} = 0.4 - 0.1^{1:4}, r_5 = 0.45 , r_6 = 0.55, 
        r_{7:10} = 0.6 + 0.1^{5 - 1:4}   \textbf{ if } t=[1,1875]; \nonumber\\
      & r_{4:1} = 0.6 + 0.1^{5 - 4:1}, r_5 = 0.55 , r_6 = 0.45,
         r_{7:10} = 0.4 - 0.1^{4:1}   \textbf{ if } t=[1876,5000]; \nonumber\\
      &r_{1:4} = 0.4 - 0.1^{1:4}, r_5 = 0.45 , r_6 = 0.55, 
       r_{7:10} = 0.6 + 0.1^{5 - 1:4}   \textbf{ if } t=[5001,9000]; \nonumber\\
      & r_{4:1} = 0.6 + 0.1^{5 - 4:1}, r_5 = 0.55 , r_6 = 0.45, 
        r_{7:10} = 0.4 - 0.1^{4:1}   \textbf{ if } t=[9001,15000]. \label{eq:expt:2} 
\end{eqnarray}
\end{figure}

\textbf{Experiment 4 (Comparison of $4$ approaches):} In figure \ref{psbandit:fig:21} we compare UCB-CPD, \ImpCPD, \UCBLCPD and UCBP-CPD for experiment 1 with $3$ Bernoulli Distributed arms. Note, that UCB-CPD and \ImpCPD uses Chernoff-Hoeffding inequality to derive their confidence interval, with \ImpCPD additionally having the knowledge of the time horizon $T$. UCBP-CPD ($\alpha = 1.5$) uses peeling argument which uses the confidence  interval mentioned in Table \ref{table:compCPD} and \UCBLCPD uses the confodence interval derived by Laplace method. In figure \ref{psbandit:fig:21} we see that the changepoint detection method using the Laplace method performs very well and outperforms the algorithms using confidence interval obtained union and peeling method. 

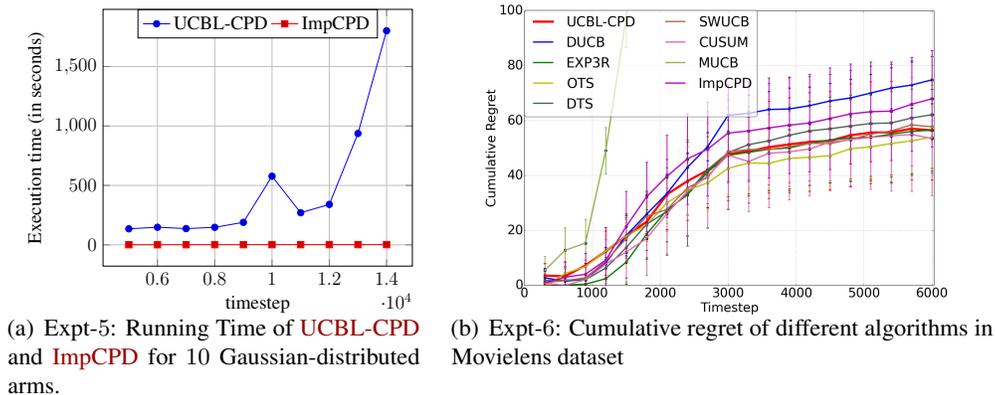
\begin{figure}
\centering
\begin{tabular}{cc}
\setlength{\tabcolsep}{0.1pt}
    \subfigure[\textwidth][ Expt-$5$: Running Time of \UCBLCPD and \ImpCPD for $10$ Gaussian-distributed arms.]
    %with $r_{i_{{i}\neq {*}}}=0.07$ and $r^{*}=0.1$
    {
            \pgfplotsset{
        tick label style={font=\large},
        label style={font=\large},
        legend style={font=\large},
        ylabel style={yshift=7pt},
        }
        \begin{tikzpicture}[scale=0.6]
          \begin{axis}[
        xlabel={timestep},
        ylabel={Execution time (in seconds)},
        grid=major,
        %clip mode=individual,grid,grid style={gray!30},
        clip=true,
        %clip mode=individual,grid,grid style={gray!30},
          legend style={at={(0.5,1.0)},anchor=north, legend columns=3} ]
          % UCB
        
        \addplot table{results/NewExpt1/Expt4/comp_subsampled_UCBCPDL03.txt};
        \addplot table{results/NewExpt1/Expt4/comp_subsampled_ImpCPD03.txt};
%        \addplot table{results/NewExpt1/Expt2_1/comp_subsampled_UCBCPD01.txt};
%          \addplot table{results/NewExpt1/Expt2_1/comp_subsampled_UCBCPDP01.txt};
          \legend{UCBL-CPD,ImpCPD}
          \end{axis}
          \end{tikzpicture}
          \label{psbandit:fig:22}
    }
    &
\subfigure[0.25\textwidth][Expt-$6$: Cumulative regret of different algorithms in Movielens dataset]
    %with $r_{i_{{i}\neq {*}}}=0.07$ and $r^{*}=0.1$
    {
    \includegraphics[scale=0.17]{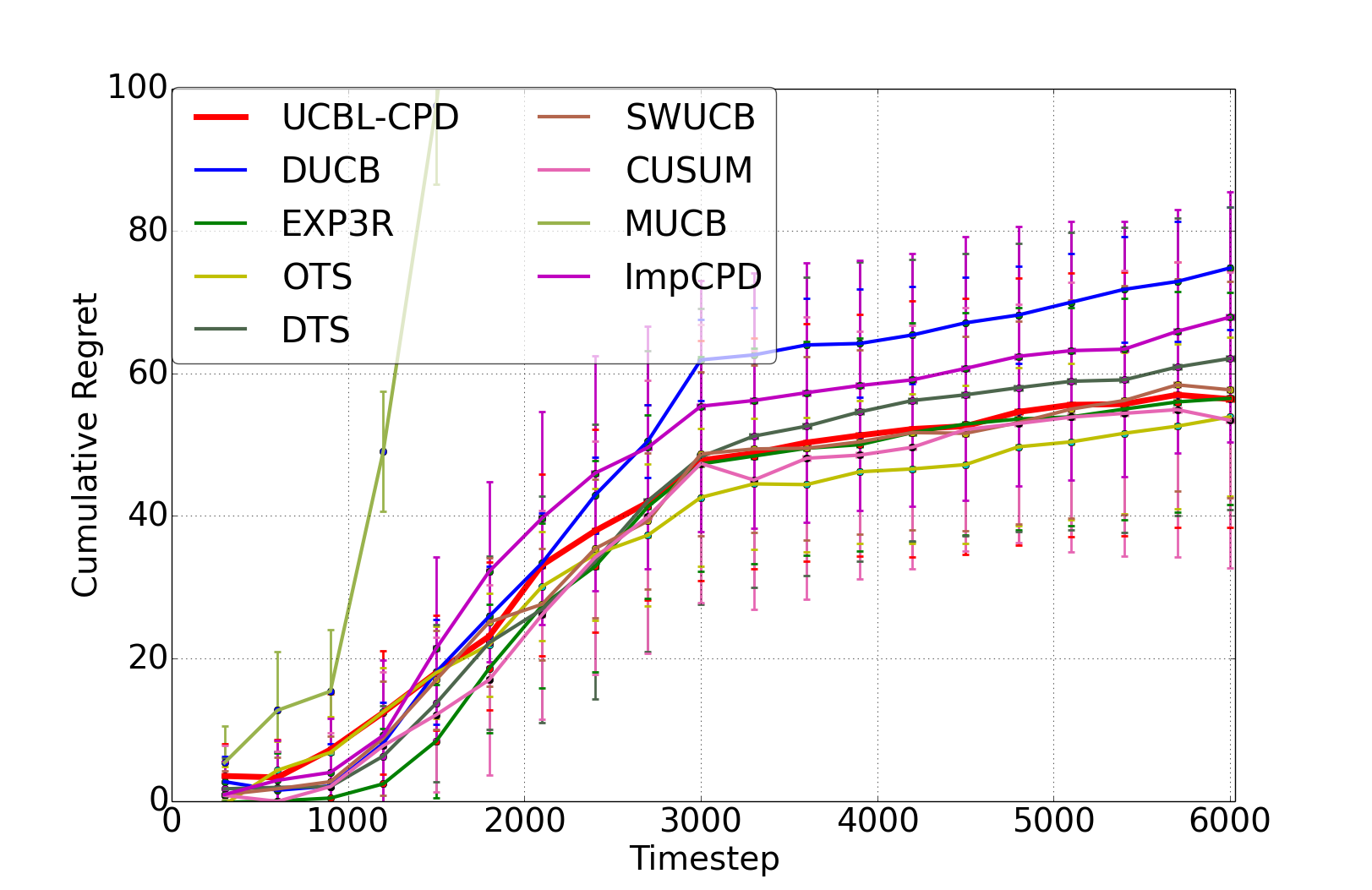}
    }
 \end{tabular}
    \caption{Additional Experiments 2.}
    \label{fig:karmed4}
    \vspace*{-1em}
\end{figure}

\textbf{Experiment 5 (Comparison of running time):} In figure \ref{psbandit:fig:22} we compare the running time of  \ImpCPD and \UCBLCPD for the experiment 3 with $10$ Gaussian Distributed arms (see equation \ref{eq:expt:2}). We test on different horizon length starting from $T=5000$ till $T=14000$ and average over $50$ times the performance of each algorithm on each of those horizon lengths. We run this experiment on a single core of 2.8 GHz Intel Core i7 processor. From the figure \ref{psbandit:fig:22} we can clearly see that \ImpCPD which has a runtime of $O(K\log T)$ clearly outperforms \UCBLCPD which has a runtime of $O(KT)$ as discussed in Section \ref{psbandit:algorithm}. 

\textbf{Experiment 6 (Movielens Dataset):} In this experiment we use another real life dataset to test the performance of our algorithm over unequal time intervals between changepoints. We experiment with the Movielens dataset from February 2003 \citep{harper2016movielens}, where there are 6k users who give 1M ratings to 4k movies. Again we obtain a rank-2 approximation of the dataset over 20 users and 100 movies. The users either prefer movie $91$ (group 1) or $99$ (group 2). Now we divide the horizon $T=6000$ into two changepoints at $t=1000$ and $t=3000$. For the time interval $t\in [1,1000]$ we choose one user from group 1, then for $t\in [1001,3000]$ we choose another user from group 2 and finally for the last piece $t\in[3001,6000]$ we choose a different user from group 1 again. Since we wanted to test only over a short horizon with unequal time intervals between changepoints, we choose only the top $10$ preferred movies (arms) over all users. We use the rating of each user to the movies (arms) as the mean of a Bernoulli Distribution. From Figure \ref{fig:karmed4} (or Figure \ref{fig:61}) we see that in this $10$ arm short horizon Bernoulli testbed \UCBLCPD and CUSUM perform really well and their performance matches the performance of the oracle algorithms OTS.

\section{Related Works}
\label{psbandit:related1}
The piecewise i.i.d setting is more general than the stochastic MAB (SMAB) setting and the adversarial setting. In the SMAB setting, the distribution associated with each arm is fixed throughout the time horizon whereas in the adversarial setting the distribution for each arm can change at any time step $t$. So for SMAB, $G=1$ whereas in the adversarial setting $G=T$. In the piecewise setting $G$ usually scales as $o(\sqrt{T})$ \citep{besson2019generalized}. The upper confidence bound (UCB) algorithms, which are a type of index-based frequentist strategy, were first proposed in \citet{agrawal1995sample} and the first finite-time analysis for the stochastic setting for this class of algorithms was proved in \citet{auer2002finite}. Strong minimax results for the SMAB setting was obtained in \citet{audibert2009minimax}, \citet{auer2010ucb}, \citet{lattimore2015optimally}. For a detailed survey of SMAB refer to \citet{bubeck2012regret}, \citet{lattimore2018bandit}. Following assumption \ref{assm:global}, we can give minimax (see 2.4.3 in \citet{bubeck2012regret}) regret bounds that incorporates the \textit{Hardness factor} $H^{}_{1,g}$ (introduced in \citet{audibert2010best}, best-arm identification) which characterizes how difficult is the environment and depends on both $\Delta^{chg}_{i,g}$ and $\Delta^{opt}_{i,g}$. Also, for this setting we need an additional hardness parameter $H_{2,g}$ which captures the trade-off between optimality and changepoint gaps. We conjecture that an order optimal regret upper bound in the piecewise i.i.d setting of the order 
\begin{align*}
\min\left\lbrace O\left(\sum_{g=1}^{G}\sum_{i=1}^{K} \frac{H_{2,g}\log(T/(GH_{1,g}))}{\Delta^{opt}_{i,g}}\right), O\left( \sqrt{GT}\right)\right\rbrace
\end{align*}
is attainable. Obtaining such optimal minimax bound for SMAB was discussed in \citet{audibert2009minimax}, \citet{auer2010ucb}, \citet{bubeck2012regret} and solved in \cite{lattimore2016regret}. We further extend the results to piecewise i.i.d algorithms (for a specific setting) which is non-trivial given the changepoint and optimality gaps have to be tackled independently.

Next, we survey some of the works for the piecewise i.i.d setting. Previous algorithms for this setting can be broadly divided into passive and actively adaptive algorithms. Passive algorithms like Discounted UCB (DUCB) \citep{kocsis2006discounted}, Sliding Window UCB (SWUCB) \citep{garivier2011upper} and Discounted Thompson Sampling (DTS) \citep{raj2017taming} do not actively try to detect changepoints and thus perform badly when changepoints are of large magnitude and are well-separated. The actively adaptive algorithm EXP3.R \citep{allesiardo2017non} is an adaptive alternative to EXP3.S \citep{auer2002nonstochastic} which was proposed for arbitrary changing environments. But EXP3.R is primarily intended for adversarial environments and thus is conservative when applied to a piecewise i.i.d. environment. The recently introduced actively adaptive algorithms CD-UCB \citep{liu2017change}, CUSUM \citep{liu2017change} and M-UCB \citep{cao2018nearly} rely on additional forced exploration for changepoint detection.  With $\alpha$ probability they employ some changepoint detection mechanism or pull the arm with highest UCB with $(1-\alpha)$ probability (exploitation). This $\alpha$ depends on either knowledge of minimum gap, $G$ or $T$. CUSUM \citep{liu2017change} performs a two-sided CUSUM test to detect changepoints and it empirically outperforms CD-UCB. CUSUM and M-UCB requires the knowledge of $G$ and $T$ for tuning $\alpha$ and CUSUM  theoretical guarantees only hold for Bernoulli rewards for widely separated changepoints. 

In \citet{garivier2011upper} the authors showed that the regret upper bound of DUCB and SWUCB are respectively $O( \sqrt{GT}\log T)$ and $O(\sqrt{GT\log T})$, where $G$ is the total number of changepoints and $T$ is the time horizon which are known apriori. Furthermore, \citet{garivier2011upper} showed that the cumulative regret in this setting is lower bounded in the order of $\Omega( \sqrt{T})$.The Restarting EXP3 (REXP3) \citep{DBLP:journals/corr/BesbesGZ14} behave pessimistically as like EXP3.S but restart after pre-determined phases. Hence, REXP3 can also be termed as a passively adaptive algorithm. The actively adaptive strategies like Adapt-EVE \citep{hartland2007change}, Windowed-Mean Shift \citep{yu2009piecewise}, EXP3.R  \citep{allesiardo2017non}, CUSUM \citep{liu2017change} try to detect the changepoints and restart. The regret bound of Adapt-EVE is still an open problem, whereas the regret upper bound of EXP3.R is $O( G\sqrt{T\log T})$ and that of CUSUM is $O( \sqrt{GT\log (T /G)})$. Note, that the regret bound of CUSUM is only valid for Bernoulli distributions. CUSUM wrongly applies Hoeffding inequality to a random number of pulls (see eq (31), (32) in \citet{liu2017change}) which raises serious concerns about the validity of the rest of their analysis. Also, there are Bayesian strategies like the Memory Bandits by \citep{alami2016memory} and Global Change-Point Thompson Sampling (GCTS) \citep{mellor2013thompson} which uses Bayesian changepoint detection to locate the changepoints. However, the regret bound of Memory Bandits and GCTS is an open problem and has not been proved yet. Both of these algorithms require very high memory usage and so have been excluded from the experiments.

Another approach involves the Generalized Likelihood Ratio Test which was  studied by \citet{maillard2019sequential} and \citet{besson2019generalized}. This is a different approach than others and looks at the ratio of the likelihood of the sequence of rewards coming from two different distributions and calculates the sufficient statistics to detect changepoints. 

%Finally, M-UCB also requires the knowledge of $G$ and $T$ for theoretical guarantees.
%The recently introduced actively adaptive algorithm CD-UCB \citep{liu2017change} is an $\alpha$-greedy policy that (at every timestep) samples uniform-randomly any arm to conduct \textit{additional forced exploration} for changepoint detection with $\alpha$ probability or plays the arm with highest UCB with $(1-\alpha)$ probability. This $\alpha$ is hard to tune in experiments and come with limited theoretical guarantees. CD-UCB requires that the exploration parameter is set to $\alpha=0$ for proving theoretical guarantees. CUSUM \citep{liu2017change} is also an $\alpha$-greedy policy that performs a two-sided CUSUM test to detect changepoints and it empirically outperforms CD-UCB. CUSUM requires the knowledge of $G$ and $T$ for tuning $\alpha$ and its theoretical guarantees only hold for Bernoulli rewards for widely separated changepoints. Also, CUSUM wrongly applies Hoeffding inequality to a random number of pulls (see eq (31), (32) in \citet{liu2017change}) which raises serious concerns about the validity of the rest of their analysis. An extended discussion can be found in Appendix \ref{sec:appendix:related}.

\section{Conclusions and Future Works}
\label{psbandit:conclusion}
We studied the piecewise i.i.d environment under assumption \ref{assm:global}, \ref{assm:space-gap} and \ref{assm:chg-gap} such that actively adaptive algorithms do not need to conduct \textcolor{blue}{forced exploration} to detect changepoints even when there are \textcolor{blue}{undetectable changepoint gaps}. We studied two UCB algorithms, \UCBLCPD and \ImpCPD which are adaptive and restarts once the changepoints are detected. We derived the \textcolor{blue}{first gap-dependent logarithmic bounds} for the piecewise i.i.d. setting incorporating the hardness factor. The anytime \UCBLCPD uses the Laplace method to derive sharp concentration bound, and \ImpCPD achieves the order optimal regret bound which is an improvement over all the existing algorithms (in a specific challenging setting). Empirically, they perform very well in various environments and is only outperformed by oracle algorithms. Future works include incorporating the knowledge of localization in these adaptive algorithms and carefully investigate the dependence on $K$ in the bounds.

\subsection*{Acknowledgments}
This work has been supported by CPER Nord-Pas de Calais/FEDER DATA Advanced data science
and technologies 2015-2020, the French Ministry of Higher Education and Research, Inria Lille
– Nord Europe, CRIStAL, and the French Agence Nationale de la Recherche (ANR), under grant
ANR-16-CE40-0002 (project BADASS).

\bibliographystyle{apalike}
\bibliography{biblio1}

\newpage

\onecolumn
\appendix
%\section{Appendix}
\label{psbandit:appendix}
%\section{Extended Related work}
%\label{sec:appendix:related}
%\input{appendix_related}

\section{Proof of Minimum Samples (Lemma \ref{psbandit:Lemma:0})}
\label{sec:proof:Lemma:0}

\begin{customproof}{1}
We consider a single arm $i\in\A$, single changepoint scenario such that $g=1$ (ignoring the expositions). Let, $t > t_g > t_{g-1} $. For the arm $i\in\A$, let $X_{i,t_{g-1}}, X_{i,t_{g-1} + 1}, X_{i,t_{g-1} + 2}, \ldots, X_{i,t_g - 1}$ be i.i.d real valued random variables having mean $\mu_{i,g-1}$ and $X_{i,t_g}, X_{i,t_g + 1}, X_{i,t_g + 2}, \ldots, X_{i,t}$ be i.i.d real valued random variables having mean $\mu_{i,g}$. Let all random variables be bounded in $[0,1]$. We want to show that for the arm $i$, after $n(t_g, \Delta, \delta) =\left\lceil \dfrac{\log(\frac{2 (t - t_{g-1})^2}{\delta})}{2\Delta^2} \right\rceil$ samples have been collected for it before (or after) $t_g$, the probability of large deviation of empirical mean $\hat{\mu}_{i,g}$ (or $\hat{\mu}_{i,g-1}$) from $\mu_{i,g}$ (or ${\mu}_{i,g-1}$) is bounded.
%n(t_{g-1},\Delta,\delta) =
%Let for an arm $i\in\A$, the changepoint $(t_{g})_{g=1}$ is detected at $\tau_g$, such that $t_{g-1} < t_g < \tau_g < t$ and $t_{g-1} = t_0 = 1$
% and hence if a changepoint occurs, it will be detected quickly.

\textbf{Step 1. (Deviation Event):} Let $\xi_{i,t}$ be the event that $\hat{\mu}_{i,g-1}$ or $\hat{\mu}_{i,g}$ is deviating from $\mu_{i,g-1}$ or $\mu_{i,g}$ by more than $\Delta$ at timestep $t >t_g$. Let $\Delta$ be the minimum gap the learner needs to detect a change for the $i$-th arm after the $g$-th changepoint has occurred. Note, that $t_{g-1}, t_g, n(t_g,\Delta,\delta)$ and $t$ are not random variables. 
%Hence for $n\in \mathbb{N}$ and $n\in [t_0,\tau_g]$, 
\begin{align*}
\xi_{i,t} &= \bigg\lbrace \exists t_{g-1} < t_g - 1, \Delta > 0 : \left | \dfrac{1}{t_g - n(t_{g},\Delta,\delta) - 1}\sum_{s=n(t_g,\Delta,\delta)}^{t_g - 1} X_{i,s}   -  \mu_{i,g-1}\right | >  \Delta \bigg\rbrace \\
& \bigcup\bigg\lbrace \exists t > t_g, \Delta > 0 : \left | \dfrac{1}{t - n(t_g,\Delta,\delta)}\sum_{s=n(t_g,\Delta,\delta)}^{t} X_{i,s}   -  \mu_{i,g}\right | >  \Delta \bigg\rbrace.
\end{align*}
%\\bigg\lbrace \left\mid \dfrac{1}{t - \tau_g}\sum_{s=1}^{t - tau_g} X_{i,s}   -  \mu_{i,g}\right\mid \geq  \Delta (t-\tau_g, \epsilon_0, g)\bigg\rbrace.
\textbf{Step 2. (Bounding the probability of deviation event):} Now we bound the probability of the detection event by using Chernoff-Hoeffding bound.
\begin{align*}
\Pb ( \xi_{i,t}) &=  \underbrace{\Pb\bigg( \exists t_{g-1} < t_g - 1, \Delta > 0 : \left | \dfrac{1}{t_g - n(t_g,\Delta,\delta) - 1}\sum_{s=n(t_g,\Delta,\delta)}^{t_g - 1} X_{i,s}   -  \mu_{i,g-1}\right | >  \Delta \bigg)}_{\textbf{term A}} \\
%%%%%%%%%%%%%%%%%
& + \underbrace{ \Pb \bigg( \exists  t_g < t, \Delta > 0 : \left | \dfrac{1}{t - n(t_g,\Delta,\delta)}\sum_{s=n(t_g,\Delta,\delta)}^{t} X_{i,s}   -  \mu_{i,g}\right | >  \Delta \bigg) }_{\textbf{term B}}.
%%%%%%%%%%%%%%%%%
\end{align*}
%Let, $t_g \geq n_\delta$, where $n_\delta =\lceil \dfrac{1}{2}\log(\frac{2 (t - t_{g-1})^2}{\delta})\rceil$.
First, we bound the probability of the term (A), 
\begin{align*}
\Pb ( A )   &=  \Pb \bigg( \left | \dfrac{1}{t_g - n(t_g,\Delta,\delta) - 1}\sum_{s=n({t_g},\Delta,\delta)}^{t_g - 1} X_{i,s}   -  \mu_{i,g-1}\right | > \Delta  \bigg)
%%%%%%%%%%%%%%%%%%%%%%%%%%%%%%%%%%%
  \leq \sum_{n = n(t_g,\Delta,\delta)}^{t_g - 1}\sum _{s = n(t_g,\Delta,\delta)}^{n} 2\exp\left( -2 \left(\Delta \right)^2 s\right)\\
%%%%%%%%%%%%%%%%%%%%%%%%%%%%%%%%%%%
& \overset{(a)}{\leq} \sum _{ n = t_{g-1} }^{t_g - 1}\sum_{s = t_{g-1}}^{n} 2\exp\left( -2 \left(\Delta \right)^2 \dfrac{\frac{1}{2}\log(\frac{2 (t - t_{g-1})^2}{\delta})}{\left(\Delta \right)^2}\right) \\
%%%%%%%%%%%%%%%%%%%%%%%%%%%%%%%%%%%
& \leq \sum _{ n = t_{g-1} }^{t_g - 1}\sum_{s = t_{g-1}}^{n} \dfrac{\delta}{2(t -t_{g-1})^2}
%%%%%%%%%%%%%%%%%
\leq \frac{\delta}{2}
\end{align*}
where, in $(a)$ we substitute $s \geq n(t_g, \Delta, \delta)$. Similarly, we can show that for $s \geq n(t_g,\Delta,\delta) =\left\lceil \dfrac{\log(\frac{2 (t - t_{g-1})^2}{\delta})}{2\Delta^2} \right\rceil$ for term (B),
\begin{align*}
\Pb ( B )  = \Pb \bigg( \exists  t_g < t, \Delta > 0 : \left | \dfrac{1}{t - n(t_g,\Delta,\delta)}\sum_{s=n(t_g,\Delta,\delta)}^{t} X_{i,s}   -  \mu_{i,g}\right | >  \Delta \bigg) \leq \frac{\delta}{2}.
\end{align*}
Summing over all of the above cases and taking a union bound, we can show that, $\Pb ( \xi_{i,t} ) \leq \delta$.
\end{customproof}

\section{Proof of Detection Delay of $\pi^*$ (Lemma \ref{psbandit:Lemma:01})}
\label{sec:proof:Lemma:01}

\begin{customproof}{2}
From Lemma \ref{psbandit:Lemma:0} we know that a minimum sample of $n(t_g, \Delta, \delta)$ is sufficient for an arm $i\in\A$ to control $\Delta$ deviation from $\mu_{i,g-1}$ or $\mu_{i,g}$ on both sides of $t_g$ with $(1-\delta)$ probability. Let $\Delta \geq \Delta(t_g, \delta)$. The changepoint detection policy may pull each arm $i=1,\ldots, K$, at most $n(t_g,\Delta,\delta) -1$ times before finally detecting the changepoint for the $K$-th arm. Let, $t_{g} <\tau_g = t < t_{g+1}$ be the time the best policy $\pi^\ast$ detects the changepoint $g$ starting exactly from $t_{g-1}$. 
%Note that $\tau_g$ is a random variable here.

\textbf{Step 1 (Deviation event):} Following Lemma \ref{psbandit:Lemma:0} we define the deviation event $\xi_{i,t}$ for an arm $i\in\A$ and $t>t_g$ as
\begin{align*}
\xi_{i,t} &= \bigg\lbrace \exists t_{g-1} < t_g - 1, \Delta \geq \Delta(t_g,\delta) : \left | \dfrac{1}{t_g - n(t_{g},\Delta,\delta) - 1}\sum_{s=n(t_g,\Delta,\delta)}^{t_g - 1} X_{i,s}   -  \mu_{i,g-1}\right | >  \Delta \bigg\rbrace \\
& \bigcup\bigg\lbrace \exists t > t_g, \Delta \geq \Delta(t_g,\delta) : \left | \dfrac{1}{t - n(t_g,\Delta,\delta)}\sum_{s=n(t_g,\Delta,\delta)}^{t} X_{i,s}   -  \mu_{i,g}\right | >  \Delta \bigg\rbrace.
\end{align*}
such that after sampling the arm $i$ for $n(t_g, \Delta, \delta)$ times the deviation of $\hat{\mu}_{i,g-1}$ (or $\hat{\mu}_{i,g}$) from $\mu_{i,g-1}$ (or $\mu_{i,g}$) is bounded with probability $(1-\delta)$.

\textbf{Step 2 (Total Delay):} The delay $d_{\pi^\ast}(t_g - t_{g-1})$ for detection policy $\pi^*$ can be upper bounded as,
\begin{align*}
d_{\pi^\ast}(t_g - t_{g-1}) &= \sum_{i=1}^{K}\bigg[ N_{i,t_{g-1}:t_g} + \Pb(\xi_{i,g}) \bigg]\\
& \leq (K - 1)(n(t_g,\Delta, \delta) - 1) + n(t_g, \Delta, \delta) + K\delta \\
& \leq  C(t_g, \delta, \eta) K \dfrac{\log(\frac{2 (t - t_{g-1})^2}{\delta})}{2\Delta(t_g, \delta)^2}  + K\delta
\end{align*}	
where, $C(t, \delta, \eta)$ depends on $t, \delta$ and $\eta$. Now, given that $t_{g-1}, t_g$ and $t_{g+1}$ follow Assumption \ref{assm:space-gap} such that $t_g + d_{\pi^\ast}(t_g - t_{g-1}) \leq t_g + \eta(t_{g+1} - t_g)$ holds. Then, 
\begin{align*}
	& C(t, \delta, \eta) K\dfrac{\log(\frac{2 (t - t_{g-1})^2}{\delta})}{2\Delta(t_g, \delta)^2} + K\delta \leq  \eta (t_{g+1} - t_{g})\\
%	& \Leftrightarrow C(t_g, \eta) K\dfrac{\log(2t (t - t_{g-1})^2)}{2\Delta(t_g, \delta)^2} + \dfrac{K}{t}  \leq \eta (t_{g+1} - t_g)\\
%	& \Leftrightarrow C(t_g, \eta) K\dfrac{\log(2t^3)}{2\Delta(t_g,\delta)^2} + \dfrac{K}{t}  \leq \eta (t_{g+1} - t_g)\\
	& \Leftrightarrow  C(t, \delta, \eta) \leq \eta \dfrac{(t_{g+1} - t_g)\Delta(t_g,\delta)^2}{2K\log (t/\delta)} - K\delta\Delta(t_g, \delta)^2 . 
	\end{align*}
Now, with a standard assumption that at $t_g$, $\Delta \geq \Delta(t_g, \delta)$ scales atleast as $\Omega(\sqrt{\frac{\log t}{t}})$ for all $i\in\A$,  $t_{g+1} - t_g \leq t$ and $\delta\in (0,1)$ we get,
\begin{align*}
C(t, \delta, \eta) \leq  \dfrac{\eta (t_{g+1} - t_g)\log (t)/t}{2K\log (t/\delta)} < \eta \log (t/\delta).
\end{align*}
\end{customproof}

\section{Proof of Control of bad-event by Laplace method (Lemma \ref{psbandit:Lemma:1})}
\label{sec:proof:Lemma:1}

%\begin{theorem}
%\label{Theorem:3}
%Let, $\mu_{i,t_0:t}$ be the expected mean of an arm $i$ for the segment $\rho_{t_0:t}$, $n_{i,t_0:t}$ be the number of times an arm $i$ is pulled from $t_0$ till the $t$-th timestep and $\hat{\mu}_{i,t_0:t}=\dfrac{1}{n_{i,t_0:t}}\sum_{q=t_0}^{t}X_{i,q}$, then at the $t$-th timestep for all $\delta\in (0,\frac{1}{4}]$  it holds that,
%\begin{align*}
%\Pb\bigg\lbrace\forall t'\in [t_0 , t]: & \big(\hat{\mu}_{i,t_0:t'} - \sqrt{\dfrac{(n_{i,t_0:t'}+1)\log(\frac{n_{i,t_0:t'}+1}{\sqrt{\delta}} )}{2n_{i,t_0:t'}^2}} > \hat{\mu}_{i,t'+1:t} + \sqrt{\dfrac{(n_{i,t'+1:t}+1)\log(\frac{n_{i,t'+1:t}+1}{\sqrt{\delta}} )}{2n_{i,t'+1:t}^2}}\big) \bigcup\\
%%%%%%%%%%%%%%%%%%%%%%%%%%%%%%%%%%%%
% &\big(\hat{\mu}_{i,t_0:t'} +  \sqrt{\dfrac{(n_{i,t_0:t'}+1)\log(\frac{n_{i,t_0:t'}+1}{\sqrt{\delta}} )}{2n_{i,t_0:t'}^2}} < \hat{\mu}_{i,t'+1:t} - \sqrt{\dfrac{(n_{i,t'+1:t}+1)\log(\frac{n_{i,t'+1:t}+1}{\sqrt{\delta}} )}{2n_{i,t'+1:t}^2}}\big)\bigg\rbrace \leq 4\delta.
%\end{align*}
%\end{theorem}

\begin{customproof}{3}
\label{proof:psbandit:Lemma:3}
%We proceed as like the proof in Theorem \ref{Theorem:2} but we rely on Laplace method to prove this. A similar approach has been followed in \citet{abbasi2011improved} for the stochastic bandit setup.
We start by noting that for any arm $i\in\A_g^{chg}$, $\pi$ detects a change-point if,
\begin{eqnarray}
&\hat{\mu}_{i,t_s:t'} - S_{i,t_s:t'} > \hat{\mu}_{i,t'+1:t} + S_{i,t'+1:t}, \text{ or } 
%%%%%%%%%%%%%%%%%%%%%%%%%%%%%%%%%%%%% \nonumber\\
&\hat{\mu}_{i,t_s:t'} +  S_{i,t_s:t'} < \hat{\mu}_{i,t'+1:t} - S_{i,t'+1:t}. \label{eq:CPD1}
\end{eqnarray}
where for any $t'\in[t_s,t]$ we define confidence interval term $S_{i,t_s:t'}$ for an arm $i\in\A$ at the $t$-th timestep as $S_{i,t_s:t'} = \sqrt{\dfrac{(N_{i,t_s:t'}+1)\log(\frac{\sqrt{(N_{i,t_s:t'}+1)}}{\delta} )}{2N_{i,t_s:t'}^2}}$.

\textbf{Step 1.(Define Bad Event:) } Let, $(X_{i,t})_{t\geq t_s}$ be a sequence of non-negative independent random variables defined on a probability space $(\omega,\mathcal{F},\Pb)$, is bounded in $[0,1]$. Let $X_{i,t_s}, X_{i,t_s + 1}, \ldots, X_{i,t'}$ have their expectation as $\mu_{i,g-1}$ and $X_{i,t'+1}, X_{i,t'+2}, \ldots, X_{i,t}$ have their expectation as $\mu_{i,g}$.  Let $\mathcal{F}_{i,t}$ be an increasing sequence of $\sigma$-fields of $\mathcal{F}_{i}$ such that for each $t$, $\sigma(X_{i,t_s},\ldots,X_{i,t})\subset \mathcal{F}_{i,t}$ and for $q>t$, $X_{i,q}$ is independent of $\mathcal{F}_{i,t}$. We define a bad event $\xi^{chg}_{i,t}$ for an arm $i\in\A^{chg}_g$ as the complementary of events in \eqref{eq:CPD1} such that,
\begin{eqnarray}
\xi^{chg}_{i,t} &=& \bigg\lbrace\exists t'\in [t_s , t]: \big(\hat{\mu}_{i,t_s:t'} - S_{i,t_s:t'} \leq \hat{\mu}_{i,t'+1:t} + S_{i,t'+1:t}\big) \nonumber\\
&& \bigcup \big(\hat{\mu}_{i,t_s:t'} +  S_{i,t_s:t'} \geq \hat{\mu}_{i,t'+1:t} - S_{i,t'+1:t}\big)\bigg\rbrace . \label{event:3}
\end{eqnarray}

\textbf{Step 2.(Define stopping times):} We define a stopping time $\tau_{i,g}$ and $\tau'_{i,g}$ for a $t'\in [t_s, t]$ such that,
\begin{align*}
\tau_{i,g} = \min\left\lbrace \exists t': \big(\hat{\mu}_{i,t_s:t'} - S_{i,t_s:t'} \leq \hat{\mu}_{i,t'+1:t} + S_{i,t'+1:t}\big) \right\rbrace ,\\
%%%%%%%%%%%%%%%%%%%%%%%%%%%%
\tau'_{i,g} = \min\left\lbrace \exists t': \big(\hat{\mu}_{i,t_s:t'} +  S_{i,t_s:t'} \geq \hat{\mu}_{i,t'+1:t} - S_{i,t'+1:t}\big) \right\rbrace .
\end{align*}

\textbf{Step 3.(Define a super-martingale): } We define the quantity $M_{t'}^{\lambda}$, for any $t'\in\mathbb{N}$ and for any $\lambda\in\mathbb{R}$ such that,
\begin{align*}
M_{t'}^{\lambda} = \exp\left( \sum_{s=t_s}^{t'}\left(\lambda (X_{i,s} - \mu_{i,t_s:t'}) - \dfrac{\lambda^2}{8}\right)\right).
\end{align*}
Since $X_{i,t_s},X_{i,t_s+1},\ldots, X_{i,t'}$ are i.i.d random variables bounded in $[0,1]$ we can consider them as $\frac{1}{2}$-sub-Gaussian; we can show that the quantity $M_{t'}^{\lambda},\forall t'\in\mathbb{N}$ is a non-negative super-martingale because, 1) $M_{t'}^{\lambda}$ is $\mathcal{F}_{i,t'-1}$ measurable, 2) $\E[|M_{t'}^{\lambda}|] < \infty, \forall t\in\mathbb{N}$, and 3) $\E[M_{t'+1}^{\lambda}|\mathcal{F}_{i,t'}]\leq M_{t'}^{\lambda}, \forall t'\in\mathbb{N}$.
%
%\begin{enumerate}
%\item $M_{t}^{\lambda}$ is $\mathcal{F}_{i,t-1}$ measurable
%\item $\E[|M_{t}^{\lambda}|] < \infty, \forall t\in\mathbb{N}$
%\item $\E[M_{t+1}^{\lambda}|\mathcal{F}_{i,t}]\leq M_{t}^{\lambda}, \forall t\in\mathbb{N}$.
%\end{enumerate} 

So, $M_{t'}^{\lambda}$ is well defined. Also, we can show that $\E[|M_{t'}^{\lambda}|]\leq 1 ,\forall t'\in \mathbb{N}$ and so $\log(\E[M_{t'}^{\lambda}])\leq 0$ is also satisfied. By convergence theorem we can show that $M_{\infty}^{\lambda}=\lim_{t'\rightarrow\infty}M_{t'}^{\lambda}$ is also well defined. Now, we introduce a new stopped version of $M_{t'}^{\lambda}$ such that $Q_{t'}^{\lambda} = M^{\lambda}_{\min\lbrace \tau_{i,g},t \rbrace}$. By Fatou's Lemma we can show that,
\begin{align*}
\E[M_{\tau_{i,g}}^{\lambda}] = \E[\lim\inf_{t'\rightarrow\infty} Q_{t'}^{\lambda}] \leq \lim\inf_{t'\rightarrow\infty}\E[ Q_{t'}^{\lambda}]\leq 1.
\end{align*}
Hence,$M_{\tau_{i,g}}^{\lambda}$ is well defined and $\E[M_{\tau_{i,g}}^{\lambda}]\leq 1$.

\textbf{Step 4.(Introduce auxiliary variable):} In this step we introduce an auxiliary variable $\Lambda = \mathcal{N}(0,4)$ which is independent of all other variables. The standard deviation of $\Lambda$ is $2$ since we are considering only $\frac{1}{2}$-sub-gaussian random variables. Hence, $\E[M_{\tau_{i,g}}^{\lambda}] = \E[\E[M_{\tau_{i,g}}^{\Lambda}|\Lambda]]\leq 1$. Let, $S_{t'} = t'\left(\frac{\sum_{s=t_s}^{t'} X_{i,s}}{t'} - \mu_{i,g-1}\right)$, then we can show that,
\begin{align*}
M_{t'} = \dfrac{1}{\sqrt{8\pi}}\int_{\mathbb{R}}\exp\left( \lambda S_{t'} - \dfrac{\lambda^2 t'}{8} - \dfrac{\lambda^2}{8} \right)d\lambda
%%%%%%%%%%%%%%%%%%%%%%%%%%%%%%%
&= \dfrac{1}{\sqrt{8\pi}}\int_{\mathbb{R}}\exp\left( - \left( \lambda\sqrt{\dfrac{t'+1}{8}} - \dfrac{\sqrt{2}S_{t'}}{\sqrt{t'+1}} \right)^2 + \dfrac{2S_{t'}^2}{t'+1} \right)d\lambda\\
%%%%%%%%%%%%%%%%%%%%%%%%%%%%%%%
&\leq \exp\left( \dfrac{2S_{t'}^2}{t'+1} \right)\dfrac{1}{\sqrt{8\pi}}\int_{\mathbb{R}}\exp\left( -\lambda^2\dfrac{t'+1}{8}\right)d\lambda\\
%%%%%%%%%%%%%%%%%%%%%%%%%%%%%%%
&= \exp\left( \dfrac{2S_{t'}^2}{t'+1}\right) \dfrac{\sqrt{8\pi}}{\sqrt{8\pi (q+1)}}\\
%\end{align*}
%\begin{align*}
%%%%%%%%%%%%%%%%%%%%%%%%%%5%%%%
\log (\sqrt{t'+1}M_{t'}) &= \dfrac{2S_{t'}^2}{t'+1}\\
%%%%%%%%%%%%%%%%%%%%%%%%%%%%%%%
S_{t'} &= \sqrt{\dfrac{t'+1}{2}\log(\sqrt{t' + 1}M_{t'})}
\end{align*}

From this we determine the confidence interval as $S_{i,t_s:t'} = \sqrt{\dfrac{(N_{i,t_s:t'}+1)\log(\frac{\sqrt{N_{i,t_s:t'}+1}}{\delta} )}{2N_{i,t_s:t'}^2}}$.

\textbf{Step 5.(Minimum pulls before $\tau_{i,g}$ and $\tau'_{i,g}$): } We define $N_{i,\min_{D}}$ as the minimum number of pulls  before $\tau_{i,g}$ such that the event $|\mu_{i, g} - \mu_{i, g-1}| > 2S_{i,t_s:\tau_{i,g}}$ is true with high probability. Then, for $N_{i,t_s:\tau_{i,g}}\geq N_{i,\min_{D}} = \dfrac{4\log(t\sqrt{t})}{(\Delta^{chg}_{i,g})^2}$ we can show that,
\begin{align*}
S_{i,t_s:\tau_{i,g}} &= \sqrt{\dfrac{(N_{i,t_s:\tau_{i,g}}+1)\log(\frac{\sqrt{N_{i,t_s:\tau_{i,g}}+1}}{\delta} )}{2N_{i,t_s:\tau_{i,g}}^2}} \leq \sqrt{\left(1+\dfrac{1}{t}\right)\dfrac{\log(\frac{\sqrt{(t+1)}}{\delta} )}{2N_{i,t_s:\tau_{i,g}}}} \\
%%%%%%%%%%%%%%%%%%%%%
&\leq \sqrt{\left(1+\dfrac{1}{t}\right)\dfrac{(\Delta^{chg}_{i,g})^2\log(\frac{\sqrt{(t+1)}}{\delta} )}{8\log(t\sqrt{t})}}
%%%%%%%%%%%%%%%%%%%%%
 \leq \dfrac{\Delta^{chg}_{i,g}}{2}.
\end{align*}
%\leq \sqrt{\left(1+\dfrac{1}{t}\right)\dfrac{(\Delta^{chg}_{i,g})^2\log(\frac{\sqrt{(t+1)}}{\delta} )}{4\left(\frac{t + 1}{t}\right)\log(\frac{\sqrt{(t+1)}}{\delta})}}\\
Similar result also holds for $\tau'_{i,g}$  such that a minimum number of pulls $N_{i,\min_D}$ is required for $|\mu_{i, g} - \mu_{i, g-1}| > 2S_{i,t_s:\tau'_{i,g}}$ to be true with high probability.

\textbf{Step 6.(Bound the probability of bad event):} To bound the probability of bad event $\xi^{chg}_{i,t}$ it suffices to show that,
\begin{align*}
\Pb&\left( N_{i,t_s:\tau_{i,g}}(\hat{\mu}_{i,t_s:\tau_{i,g}} - \mu_{i,g-1}) \leq \sqrt{\dfrac{N_{i,t_s:\tau_{i,g}} + 1}{2}\log\left(\dfrac{\sqrt{N_{i,t_s:\tau_{i,g}} + 1}}{\delta}\right)}\right) \\
%%%%%%%%%%%%%%%%%%%%%
&\leq \Pb\left( M_{\tau_{i,g}} \geq \dfrac{1}{\delta}\right) \overset{(a)}{\leq} \dfrac{\E[M_{\tau_{i,g}}]}{\frac{1}{\delta}} \overset{(b)}{\leq} \delta
\end{align*}
where, in $(a)$ we use Markov inequality and $(b)$ comes from step 3.
%Similarly, we can show that for any $\tau'_{i,c_j}\in[t_s,t]$,
%\begin{align*}
%\Pb\bigg\lbrace \hat{\mu}_{i,\tau'_{i,g}+1:t} \geq \mu_{i,\tau'_{i,g}+1:t} + s_{i,\tau'_{i,g}+1:t}\bigg\rbrace &\leq \delta.
%\end{align*}
Similarly, we can define another super-martingale $M_{t-t'}$ and follow the above procedure to show that 
\begin{align*}
\Pb\left( N_{i,t'+1:\tau_{i,g}}(\hat{\mu}_{i,t'+1:\tau_{i,g}} - \mu_{i,g}) \geq \sqrt{\dfrac{N_{i,t'+1:\tau_{i,g}} + 1}{2}\log\left(\dfrac{\sqrt{N_{i,t'+1:\tau_{i,g}} + 1}}{\delta}\right)}\right) \leq \delta
\end{align*}
Hence, the probability that the change-point for the arm $i$ is not detected by first condition of $\xi^{chg}_{i,t}$ for any $\tau_{i,g}\in[t_s,t]$ is bounded by $2\delta$. Again, for the second condition we can proceed in a similar way and show that for all $t'\in[t_s,t]$,
\begin{align*}
\Pb\bigg(\hat{\mu}_{i,g-1} +  S_{i,t_s:\tau_{i,g}} \geq \hat{\mu}_{i,\tau'_{i,g}+1:t} - S_{i,\tau'_{i,g}+1:t}\bigg) \leq 2\delta.
\end{align*}
Hence, we can bound the probability of the bad event $\xi^{chg}_{i,t}$ for an arm $i\in\A^{chg}_g$ for any $\tau_{i,g},\tau'_{i,g}\in[t_s,t]$ by taking the union of all the events above and get, $\Pb\lbrace \xi^{chg}_{i,t}\rbrace \leq 4\delta$.
%\begin{align*}
%\Pb\lbrace \xi^{chg}_{i,t}\rbrace \leq 2\delta.
%\end{align*}

\end{customproof}

%\begin{remark}
%\label{Rem:4}
%Choosing $\delta=\dfrac{1}{t}$, the above result of theorem \ref{Theorem:3} can be reduced to $\Pb\lbrace \xi^{chg}_{i,t}\rbrace \leq \dfrac{4}{t}$, where the event $\xi^{chg}_{i,t}$ is identical to Eq (\ref{event:3}).
%\end{remark}

\section{Proof of Control of bad-event by Union Bound  method}
\label{sec:proof:Lemma:11}

The UCB-CPD algorithm is exactly like Algorithm \ref{alg:UCBCPD} but uses the definition of $S_{i,t_s:t_p}$ as 
%\begin{align*}
$ \sqrt{\frac{\log(\frac{4t^2}{\delta})}{2N_{i,t_s:t_p}}}$.
%\end{align*}

\begin{customlemma}{4} \textbf{(Control of bad-event by Union Bound  method)}
\label{psbandit:Lemma:11}
Let, $\mu_{i,g}$ be the expected mean of an arm $i$ for the piece $\rho_{g}$, $N_{i,t_s:t}$ be the number of times an arm $i$ is pulled from $t_s$ till the $t$-th timestep such that $t>t_{g}$, then at the $t$-th timestep for all $\delta\in (0,1]$  it holds that,
\begin{align*}
\Pb\bigg(\!\exists & t'\!\!\in\! [t_s , t]\!: \big(\hat{\mu}_{i,t_s:t'} \!-\! S_{i,t_s:t'} \!>\! \hat{\mu}_{i,t'\!+\!1:t} \!+\! S_{i,t'\!+\!1:t}\big) \bigcup
%%%%%%%%%%%%%%%%%%%%%%%%%%%%
  \big(\hat{\mu}_{i,t_s:t'} \!+\!  S_{i,t_s:t'} \!<\! \hat{\mu}_{i,t'\!+\!1:t} \!-\! S_{i,t'\!+\!1:t} \big)\!\bigg) \leq \delta
\end{align*}
where $S_{i,t_s:t'} = \sqrt{\dfrac{\log(\frac{4t^2}{\delta})}{2N_{i,t_s:t'}}}$.
\end{customlemma}

\begin{customproof}{4}
\label{proof:Lemma:11}
We define confidence interval term $S_{i,t_s:t'}$ for an arm $i\in\A^{chg}_g$ at the $t$-th timestep as $S_{i,t_s:t'} = \sqrt{\dfrac{\log(\frac{4t^2}{\delta})}{2N_{i,t_s:t'}}}$ and $t_s$ is an arbitrary restarting timestep. 
 
\textbf{Step 1.(Define Bad Event):} We define a bad event $\xi^{chg}_{i,t}$ for the arm $i$ as the complementary of events in Eq (\ref{eq:CPD1}) such that,
\begin{eqnarray}
\xi^{chg}_{i,t} &=& \bigg\lbrace\exists t'\in [t_s , t]: \big(\hat{\mu}_{i,t_s:t'} - S_{i,t_s:t'} \leq \hat{\mu}_{i,t'+1:t} + S_{i,t'+1:t}\big) \nonumber\\
%%%%%%%%%%%%%%%%
&& \bigcup \big(\hat{\mu}_{i,t_s:t'} +  S_{i,t_s:t'} \geq \hat{\mu}_{i,t'+1:t} - S_{i,t'+1:t}\big)\bigg\rbrace. \label{event:1}
\end{eqnarray}

\textbf{Step 2.(Define stopping times):} We define a stopping time $\tau_{i,g}$ and $\tau'_{i,g}$ for a $t'\in [t_s, t]$ such that,
%\begin{align*}
%\tau_{i,g} = \min\left\lbrace t': S_{i,t_s:t'} < \dfrac{\Delta^{chg}_{i,g}}{2} \right\rbrace ,\hspace*{4em}
%%%%%%%%%%%%%%%%%%%%%%%%%%%%%
%\tau'_{i,g} = \min\left\lbrace t': S_{i,t'+1:t} < \dfrac{\Delta^{chg}_{i,g}}{2} \right\rbrace
%\end{align*}
\begin{align*}
\tau_{i,g} = \min\left\lbrace \exists t': \big(\hat{\mu}_{i,t_s:t'} - S_{i,t_s:t'} \leq \hat{\mu}_{i,t'+1:t} + S_{i,t'+1:t}\big) \right\rbrace ,\\
%%%%%%%%%%%%%%%%%%%%%%%%%%%%
\tau'_{i,g} = \min\left\lbrace \exists t': \big(\hat{\mu}_{i,t_s:t'} +  S_{i,t_s:t'} \geq \hat{\mu}_{i,t'+1:t} - S_{i,t'+1:t}\big) \right\rbrace .
\end{align*}

\textbf{Step 3.(Minimum pulls before $\tau_{i,g}$ and $\tau'_{i,g}$): } We define $N_{i,\min_{D}}$ as the minimum number of pulls  before $\tau_{i,g}$ such that the event $|\mu_{i, g} - \mu_{i, g-1}| > 2S_{i,t_s:\tau_{i,g}}$ is true with high probability. Then, for $N_{i,t_s:\tau_{i,g}}\geq N_{i,\min_{D}} = \dfrac{2\log(\frac{4t^2}{\delta})}{(\Delta^{chg}_{i,g})^2}$ we can show that,
\begin{align*}
S_{i,t_s:\tau_{i,g}} = \sqrt{\dfrac{\log(\frac{4t^2}{\delta})}{2N_{i,t_s:\tau_{i,g}}}} \leq \sqrt{\dfrac{\log(\frac{4t^2}{\delta})}{2N_{i,\min_D}}} \leq \sqrt{\dfrac{(\Delta^{chg}_{i,g})^2\log(\frac{4t^2}{\delta})}{4\log(\frac{4t^2}{\delta})}} \leq \dfrac{\Delta^{chg}_{i,g}}{2}.
\end{align*} 
Similar result also holds for $\tau'_{i,g}$  such that a minimum number of pulls $N_{i,\min_D}$ is required for $|\mu_{i, g} - \mu_{i, g-1}| > 2S_{i,t_s:\tau'_{i,g}}$ to be true with high probability.

%so that the event $|\mu_{i,\rho_{c_{j-1}:g}}-\mu_{i,\rho_{g:g+1}}| < 2S_{i,t_s:\tau_{i,g}}$ never occurs and $|\mu_{i,\rho_{c_{j-1}:g}}-\mu_{i,\rho_{g:g+1}}|$ are sufficiently far apart.

\textbf{Step 4.(Bounding the probability of bad event): } In this step we bound the probability of the bad event $\xi^{chg}_{i,t}$ in Eq (\ref{event:1}). 

%Now from Chernoff-Hoeffding inequality we know that for any constant $\epsilon > 0$,
%\begin{align*}
%\Pb\bigg(\left| \dfrac{\sum_{q=1}^{n} X_{i,q}}{n} - \mu_{i,g}\right| \geq \epsilon\bigg) \leq 2\exp\left(-2\epsilon^2 n\right)   
%\end{align*}
Let $X_{i,t_s}, X_{i,t_s+1}, \ldots, X_{i,t'}$ have their expectation as $\mu_{i,g-1}$ and $X_{i,t'+1}, X_{i,t'+2}, \ldots, X_{i,t}$ have their expectation as $\mu_{i,g}$. Then we define the event $\xi_1= \lbrace \exists q\in [N_{i,\min_D}, t'] :  \hat{\mu}_{i,q:t'} \leq \mu_{i,g-1} - S_{i,q:t}\rbrace$. Starting with the first condition of $\xi^{chg}_{i,t}$ and applying the Chernoff-Hoeffding inequality we can show that for a $\tau_{i,g}\in [t_s,t]$,
\begin{align*}
\Pb( \xi_1) \leq \sum_{t'=N_{i,\min_D}}^{t} \sum_{q=N_{i,\min_D}}^{t'}\Pb\bigg (  \dfrac{\sum_{s=1}^{q}X_{i,s}}{q} \leq \mu_{i,g-1} - \dfrac{\log(\frac{4t^2}{\delta})}{2q}\bigg) & \leq\!\!\!\! \sum_{t'=N_{i,\min_D}}^{t} \sum_{q=N_{i,\min_D}}^{t'}\exp\left(-2.\dfrac{\log(\frac{4t^2}{\delta})}{2q}q \right) \\
%%%%%%%%%%%%%%%%%%
& \leq \sum_{t'=1}^{t} \sum_{q=1}^{t'}\dfrac{\delta}{4t^2}\leq \frac{\delta}{4}.
\end{align*}
Again, defining the event $\xi_2 = \lbrace \exists q\in [N_{i,\min_D}, t'] :  \hat{\mu}_{i,q:t'} \geq \mu_{i,g} + S_{i,q:t'}\rbrace$  and proceeding similarly as above, we can show that for a $\tau'_{i,g}\in[t_s,t]$,
\begin{align*}
\Pb( \xi_2 ) \leq \sum_{t'=N_{i,\min_D}}^{t} \sum_{q=N_{i,\min_D}}^{t'}\Pb\bigg(  \dfrac{\sum_{s=1}^{q}X_{i,s}}{q} \geq \mu_{i,g} + \dfrac{\log(\frac{4t^2}{\delta})}{2q} \bigg) \leq \sum_{t'=1}^{t} \sum_{q=1}^{t'}\dfrac{\delta}{4t^2} \leq \frac{\delta}{4}.
\end{align*}
Summing the two up, the probability that the changepoint for the arm $i$ is not detected by first condition of $\xi^{chg}_{i,t}$ for a $t'\in[t_s,t]$ is bounded by $\frac{\delta}{2}$. Again, for the second condition we can proceed in a similar way and show that for all $t'\in [t_s,t]$, $\tau_{i,g},\tau'_{i,g}\in[t_s,t']$ and the two events $\xi_1 = \lbrace \exists q\in [N_{i,\min_D}, t] :  \hat{\mu}_{i,q:t} \leq \mu_{i,g} - S_{i,q:t}\rbrace $  and  $\xi_2 = \lbrace \exists q\in [N_{i,\min_D}, t] :  \hat{\mu}_{i,q:t} \geq \mu_{i,g} + S_{i,q:t}\rbrace $,
\vspace*{-1em}
\begin{align*}
 \Pb( \xi_1) + \Pb( \xi_2 )  & \leq  \sum_{t'=1}^{t} \sum_{q=1}^{t'}\Pb\bigg(  \dfrac{\sum_{s=1}^{q}X_{i,s}}{q} +  \dfrac{\log(\frac{4t^2}{\delta})}{2q} \geq \dfrac{\sum_{s=q+1}^{t}X_{i,s}}{t-(q+1)} - \dfrac{\log(\frac{4t^2}{\delta})}{2(t-(q+1))}  \bigg)\\
%%%%%%%%%%%%%%%%%%%%
&\leq  \sum_{t'=1}^{t} \sum_{q=1}^{t'}\dfrac{\delta}{4 t^2} +\sum_{t'=1}^{t} \sum_{q=1}^{t'} \dfrac{\delta}{4t^2} \leq \frac{\delta}{2} .
\end{align*} 
\vspace*{-0.3em}
Hence, we can bound the probability of the bad event $\xi^{chg}_{i,t}$ for an arm $i\in\A^{chg}_g$ by taking the union of all such events for all $t'\in[t_s,t]$  as $\Pb( \xi^{chg}_{i,t}) \leq \delta.$
%\begin{align*}
%\Pb( \xi^{chg}_{i,t}) \leq \delta.
%\end{align*}
\end{customproof}

\section{Proof of Regret Bound of \UCBLCPD}
\label{sec:proof:Theorem:1}

\begin{customproof}{5}

%\textbf{Step 0.(Vital Assumption for proving):} In this proof, we assume that all the changepoints are detected by the algorithm with some delay. Since, all the gaps are significant (Assumption \ref{assm:chg-gap}) and there is sufficient delay between two changepoints (Assumption \ref{assm:space-gap}), we can take this assumption for proving the result.

\textbf{Step 1.(Some notations and definitions):} Let, $N_{i,\tau_{g-1}:\tau_g}$ denote the number of times an arm $i\in\A_g^{chg}$ is pulled  between $\tau_{g-1}$ to $\tau_g$-th timestep. We recall the definition of stopping times $\tau_{i,g}$, $\tau'_{i,g}$ for an arm $i$  from Lemma \ref{psbandit:Lemma:1}, step 2 as,
%\begin{align*}
%\tau_{i,g} = \min\left\lbrace t': S_{i,\tau_{i,g-1}:t'} < \dfrac{\Delta^{chg}_{i,g}}{2} \right\rbrace ,\hspace*{4em}
%%%%%%%%%%%%%%%%%%%%%%%%%%%%%
%\tau'_{i,g} = \min\left\lbrace t': S_{i,t'+1:t} < \dfrac{\Delta^{chg}_{i,g}}{2} \right\rbrace
%\end{align*}
\begin{align*}
\tau_{i,g} = \min\left\lbrace \exists t': \big(\hat{\mu}_{i,\tau_{i,g-1}:t'} - S_{i,\tau_{i,g-1}:t'} \leq \hat{\mu}_{i,t'+1:t} + S_{i,t'+1:t}\big) \right\rbrace ,\\
%%%%%%%%%%%%%%%%%%%%%%%%%%%%
\tau'_{i,g} = \min\left\lbrace \exists t': \big(\hat{\mu}_{i,\tau_{i,g-1}:t'} +  S_{i,\tau_{i,g-1}:t'} \geq \hat{\mu}_{i,t'+1:t} - S_{i,t'+1:t}\big) \right\rbrace .
\end{align*}
where $t'\in [\tau_{i,g-1}, t]$ and $t> t_g$. We denote $N_{i,\min_{D}}= \dfrac{4\log(t\sqrt{t})}{(\Delta^{chg}_{i,g})^2}$ as the minimum number of pulls required for an arm $i$ such that the event $|\mu_{i, g} - \mu_{i, g-1}| > 2S_{i,t_s:\tau_{i,g}}$ is true with high probability. We also recall the definition of the bad event $\xi^{chg}_{i,t}$ when the changepoint is not detected as,
\begin{align*}
\xi^{chg}_{i,t} &= \bigg\lbrace\exists t'\in [\tau_{i,g-1}, t]: \big(\hat{\mu}_{i,\tau_{g-1}:t'} - S_{i,\tau_{i,g-1}:t'} \leq \hat{\mu}_{i,t'+1:t} + S_{i,t'+1:t}\big) \\
%%%%%%%%%%%%%%%%%%
& \bigcup \big(\hat{\mu}_{i,\tau_{i,g-1}:t'} +  S_{i,\tau_{i,g-1}:t'} \geq \hat{\mu}_{i,t'+1:t} - S_{i,t'+1:t}\big)\bigg\rbrace.
\end{align*}
Also, we introduce the term $\tau_g$ as the time the algorithm detects the changepoint $g\in\G$ and resets as opposed to the term $\tau_{i,g}$ when the algorithm detects the changepoint for the $i$-th arm.

\textbf{Step 2.(Define an optimality stopping time): } We define an optimality stopping time $\tau^{opt}_{i,g}$ for a $\tau^{opt}_{i,g}\in[\tau_{i,g-1},t]$such that,
\begin{align*}
\tau^{opt}_{i,g} = \min\bigg\lbrace t': \left( \hat{\mu}_{i,\tau_{i,g-1}:t'} < \mu_{i,g-1} + S'_{i,\tau_{i,g-1}:t'} \right)
\bigcup \left(\hat{\mu}_{i^*,\tau_{i,g-1}:t'} > \mu_{i^*,g-1} - S'_{i^*,\tau_{i,g-1}:t'} \right)\bigg\rbrace 
\end{align*}
where $S'_{i,\tau_{i,g-1}:t'} = \sqrt{\dfrac{(N_{i,t_s:t'}+1)\log(\frac{\sqrt{(N_{i,t_s:t'}+1)}}{\delta} )}{2N_{i,t_s:t'}^2}}$.  Note, that if the two events for $\tau^{opt}_{i,g}$ come true then the sub-optimal arm is no longer pulled.

%\sqrt{\dfrac{2\log(t)}{N_{i,\tau_{i,g-1}:t'}}}    

\textbf{Step 3.(Minimum pulls of a sub-optimal arm): } We denote $N_{i,\min_{E}}$ as the minimum number of pulls required for a sub-optimal arm $i$ such that $(\mu_{i^*,g-1} - \mu_{i,g-1}) > 2S'_{i,\tau_{i,g-1}:t'}$ is satisfied with high probability. Indeed we can show that for $N_{i,\tau_{i,g-1}:t}\geq N_{i,\min_{E}} = \frac{4\log (t\sqrt{t})}{(\Delta^{opt}_{i,g})^2}$,
\begin{align*}
 S'_{i,\tau_{i,g-1}:t'}= \sqrt{\dfrac{(N_{i,t_s:t'}+1)\log(\frac{\sqrt{(N_{i,t_s:t'}+1)}}{\delta} )}{2N_{i,t_s:t'}^2}} \leq \sqrt{\left( 1 + \dfrac{1}{N_{i,\min_E}} \right)\dfrac{\log(\frac{\sqrt{(t+1)}}{\delta} )}{2N_{i,\min_E}}}  \leq \dfrac{\Delta^{opt}_{i,g}}{2}
\end{align*}
 
%
%\begin{align*}
%S'_{i,\tau_{i,g-1}:t} = \sqrt{\dfrac{2\log(t)}{N_{i,\tau_{i,g-1}:t}}} \leq \sqrt{\dfrac{2\log(t)}{N_{i,\min_{E}}}}\leq \dfrac{\Delta^{opt}_{i,g}}{2}
%\end{align*}

\textbf{Step 4.(Define the optimality bad event):} We define the optimality bad event $\xi^{opt}_{i,t}$ as,
%\begin{align*}
%\xi^{opt}_{i,t} = \left\lbrace\forall t'\in [\tau_{i,c_{j-1}} , t]: S'_{i,\tau_{i,c_{j-1}}:t'} > \dfrac{\Delta_{i,g}^{opt}}{2} \right\rbrace
%\end{align*}
\begin{align*}
\xi^{opt}_{i,t} = \left\lbrace\exists t'\in [\tau_{i,g-1} , t]: \hat{\mu}_{i^*,\tau_{i,g-1}:t} + S'_{i^*,\tau_{i,g-1},t} \leq  \hat{\mu}_{i,\tau_{i,g-1}:t} + S'_{i,\tau_{i,g-1},t} \right\rbrace
\end{align*}
For the above event to be true the following three events have to be true,
\begin{align*}
\xi^{opt}_{i,t} \subset \bigg\lbrace\exists t'\in [\tau_{i,g-1} , t]: &\lbrace\hat{\mu}_{i,\tau_{i,g-1}:t'} \geq \mu_{i,g-1} + S'_{i,\tau_{i,g-1}:t'}\rbrace 
%%%%%%%%%%%%%%%%%%%%%%
\bigcup \lbrace\hat{\mu}_{i^*,\tau_{i,g-1}:t'} \leq \mu_{i^*,g-1} - S'_{i^*,\tau_{i,g-1}:t'}\rbrace \\
%%%%%%%%%%%%%%%%%%%%%%\\
&\bigcup \lbrace (\mu_{i^*,g-1} - \mu_{i,g-1}) \leq 2S'_{i,\tau_{i,g-1}:t'} \rbrace\bigg\rbrace
\end{align*}
which implies that the sub-optimal arm $i$ has been over-estimated, the optimal arm $i^*$ has been under-estimated or there is sufficient gap between $(\mu_{i^*,g-1} - \mu_{i,g-1})$. But, from step $3$, we know that for $N_{i,\tau_{i,g-1}:t'}\geq N_{\min_E}$ the third event of $\xi^{opt}_{i,t}$ is not possible with high probability.

%\xi^{opt}_{i,t} = \left\lbrace\forall t'\in [\tau_{i,c_{j-1}} , t]: \hat{\mu}_{i,\tau_{i,c_{j-1}}:t'} - \mu_{i,\tau_{i,c_{j-1}}:t'} > 2S'_{i,\tau_{i,c_{j-1}}:t'}\right\rbrace

\textbf{Step 5.(Bound the number of pulls):} Let $\tau_{i,t_0}=1$ denoting the first timestep. Now, the total number of pulls  $N_{i,1:t}$ for an arm $i\in\A$ till $t$-th timestep is given by,
\begin{align*}
N_{i,1:t} &= \left[ 1 + \sum_{t'=K+1}^{t}\mathbf{1}_{\lbrace i_{t'} \neq i^*_{t'}\rbrace} \right] \\
\end{align*}
We consider the contribution of each changepoint $g\in\G,\forall g=1,\ldots,G$ to the number of pulls $N_{i,1:t}$ and decompose the event $\lbrace i_{t'} \neq i^*_{t'}\rbrace$ into five parts,
\begin{eqnarray}
&&\!\! N_{i,1:t} \! =\!\! \sum_{j=1}^{G}\bigg[ \underbrace{1 \! +\!\!\!\!\!\! \sum_{t'=\tau_{i,g-1}+K+1}^{t_{g}}\hspace*{-2.4em} \mathbf{1}_{\lbrace i_{t'} \neq i^*_{t'},N_{i,\tau_{i,g-1}:t_{g}} < \max\lbrace N_{i,\min_E}, N_{i,\min_D}\rbrace\rbrace}}_{\textbf{part A}} \! +\! \!\underbrace{\sum_{t'=t_{g}}^{\tau_{i,g}}\mathbf{1}_{\lbrace i_{t'} \neq i^*_{t'},N_{i,t_{g}:\tau_{i,g}}\geq N_{i,\min_D}\rbrace}}_{\textbf{part D}}\nonumber\\
%%%%%%%%%%%%%%%%%%%%%%%%%%%%
&& + \underbrace{\sum_{t'=\tau_{i,g-1}+K+1}^{t_{g}}\mathbf{1}_{\lbrace i_{t'} \neq i^*_{t'},N_{i,\tau_{i,g-1}:t_{g}} \geq \max\lbrace N_{i,\min_E}, N_{i,\min_D}\rbrace\rbrace}}_{\textbf{part B}} + \underbrace{\sum_{t'=\tau_{g-1}}^{t}\mathbf{1}_{\{ i_{t'} \neq i^*_{t'},t > \tau_{g-1} > t_{g-1}\}  }}_{\textbf{part E}} \nonumber\\
%%%%%%%%%%%%%%%%%%%%%%%%%%%%
&& + \underbrace{ \sum_{t'=t_{g}}^{\tau_{i,g}}\mathbf{1}_{\left\lbrace { i_{t}' \neq i^*_{t'}, N_{i,t_g:\tau_{i,g}} < N_{i,min_D}, \tau_{i,g}=\tau_g}  \right\rbrace} + \sum_{t'=t_{g}}^{\tau_{g}}\mathbf{1}_{\lbrace i_{t'} \neq i^*_{t'}, N_{i,t_g:\tau_{i,g}} < N_{i,min_D}, \tau_{i,g} < \tau_g  \rbrace} }_{\textbf{part C}} \nonumber\\
&& + \underbrace{ \sum_{t'=t_{g}}^{\tau_{i,g+1}}\mathbf{1}_{\left\lbrace { i_{t}' \neq i^*_{t'}, N_{i,t_g:\tau_{i,g+1}} < 2N_{i,min_D}, \tau_{i,g}=\tau_{g+1}}  \right\rbrace} \! +\!\! \sum_{t'=t_{g}}^{\tau_{g+1}}\mathbf{1}_{\lbrace i_{t'} \neq i^*_{t'}, N_{i,t_g:\tau_{i,g+1}} < 2N_{i,min_D}, \tau_{i,g} < \tau_{g+1}  \rbrace} }_{\textbf{part C}}
 \bigg] \label{eq:pull}
\end{eqnarray}
where, \textbf{part A} refers to the minimum number of pulls required before either \UCBLCPD discards the sub-optimal arm $i$ or a changepoint is detected between $\tau_{i,g-1}:t_{g}$, \textbf{part B} refers to the number of pulls due to bad event that  \UCBLCPD continues to pull the sub-optimal arm $i$ or the changepoint is not detected between $\tau_{i,g-1}:t_{g}$, \textbf{part C} refers to the pulls accrued due to the worst case events when the changepoint has occurred and has not been detected by \UCBLCPD from $t_{g}:\tau_{i,g+1}$, \textbf{part D} refers to number of pulls for the bad event that the changepoint is not detected from $t_{g}:\tau_{i,g}$, and finally \textbf{part E} refers to the case when the changepoint $g$ is not detected from time $\tau_{g-1} > t_{g-1}$ till time $t$. Now, we will consider this parts individually and see their contribution to the number of pulls. For \textbf{part A},
\begin{align*}
\underbrace{1 + \sum_{t'=\tau_{i,g-1}+K+1}^{t_{g}}\mathbf{1}_{\lbrace i_{t'} \neq i^*_{t'},N_{i,\tau_{i,g-1}:t_{g}} < \max\lbrace N_{i,\min_E}, N_{i,\min_D}\rbrace\rbrace}}_{\textbf{part A}} & \leq 1 + N_{i,\min_E} + N_{i,\min_D} \\
%%%%%%%%%%%%%%%%%%%%%%%%%%%%%555
& \overset{(a)}{=} 1 + \dfrac{6\log t}{(\Delta^{opt}_{i,g-1})^2} + \dfrac{6\log t}{(\Delta^{chg}_{i,g})^2}
\end{align*}
% \dfrac{8\log t}{(\Delta^{opt}_{i,g-1})^2} + \dfrac{2\log(\frac{4t^2}{\delta})}{(\Delta^{chg}_{i,g})^2}
where $(a)$ is obtained from previous step $3,4$. \textbf{Part B} deals with the case of \textit{false detection} such that event of raising an alarm for $t_g$ between $\tau_{i,g-1}:t_{g} -1$ is bounded. By the uniform concentration bound property of $\xi^{opt}_{i,t_{g}}$ and $\xi^{chg}_{i,t_{g}}$ we can show that the event,
\begin{align*}
\lbrace i_{t'} \neq i^*_{t'},N_{i,\tau_{i,g-1}:t_{g}} \geq \max\lbrace N_{i,\min_E}, N_{i,\min_D}\rbrace\rbrace \subseteq \lbrace \xi^{opt}_{i,t_{g}} + \xi^{chg}_{i,t_{g}}  \rbrace
\end{align*}
which is obtained from step 4. Hence, we can upper bound \textbf{part B} as,
\begin{align*}
\underbrace{\sum_{t'=\tau_{i,g-1}+K+1}^{t_{g}}\mathbf{1}_{\lbrace i_{t'} \neq i^*_{t'},N_{i,\tau_{i,g-1}:t_{g}} \geq \max\lbrace N_{i,\min_E}, N_{i,\min_D}\rbrace\rbrace}}_{\textbf{part B}}  \leq \sum_{t'=\tau_{i,g-1}+K+1}^{t_{g}} \!\!\!\!\!\! \Pb\lbrace\xi^{opt}_{i,t'}\rbrace + \sum_{t'=\tau_{i,g-1}+K+1}^{t_{g}}\!\!\!\!\!\!  \Pb\lbrace\xi^{chg}_{i,t'}\rbrace.
\end{align*}

Now, \textbf{part C} consist of the worst case scenario where the changepoint has occurred and has not been detected. A good learner always try to minimize the number of pulls of the sub-optimal arms that might occur in this period from $t_{g}:\tau_{i,g}$. Then, we can show that the first two events in \textbf{part C} is the subset of the union of several worst case events. 
\begin{align*}
&\lbrace i_{t'}\neq i^*_{t'},N_{i,t_{g}:\tau_{i,g}} < N_{i,\min_D}, \tau_{i,g}=\tau_g\rbrace \bigcup \lbrace i_{t'}\neq i^*_{t'},N_{i,t_{g}:\tau_{i,g}} < N_{i,\min_D}, \tau_{i,g} < \tau_g\rbrace \\
& \subseteq \left\lbrace i_{t'}\in\argmax_{i\in\A}\big\lbrace\argmax_{k\in\A}\lbrace \Delta^{opt}_{k,g+1}\rbrace \bigcup \argmin_{k\in\A}\lbrace \Delta^{chg}_{k,g}\rbrace \bigcup \argmax_{k\in\A}\lbrace \frac{\Delta^{opt}_{k,g+1}}{\Delta^{chg}_{k,g}} \rbrace\big\rbrace \right\rbrace
\end{align*}
Now, to bound the number of pulls of the sub-optimal arm $i$ due to these several worst case events, we note that for the first event of \textbf{part C}, the changepoint is detected due to arm $i$. Hence,
\begin{align*}
\sum_{t'=t_{g}}^{\tau_{i,g}}\mathbf{1}_{\{ i_{t'} \neq i*_{t'}, N_{i,t_g:\tau_{i,g}} < N_{i,min_D}, \tau_{i,g}=\tau_g\}  } \leq N_{i,\min_D} 
\end{align*}
Next, for the second event of \textbf{part C}, the changepoint is detected not due to arm $i$, but for a different arm at a later timestep such that $\tau_g > \tau_{i,g}$. Hence,
\begin{align*}
\sum_{t'=t_{g}}^{\tau_{g}}\mathbf{1}_{\{ i_{t'} \neq i*_{t'}, N_{i,t_g:\tau_{i,g}} < N_{i,min_D}, \tau_{i,g} < \tau_g  \} } \leq 
\max_{N_{1,t_{g}:\tau_g},...,N_{K,t_{g}:\tau_g}} \left\lbrace \sum_{i=1}^{K} N_{i,t_{g}:\tau_g} : N_{i,t_{g}:\tau_g} < N_{i,min_D} \right\rbrace.
\end{align*}
Similarly following assumption \ref{assm:chg-gap} if $\Delta_{i,g}^{chg}$ is undetectable then $\Delta_{i,g+1}^{chg}$ is detectable, we can bound the third and fourth event in \textbf{part C} as,
\begin{align*}
\sum_{t'=t_{g}}^{\tau_{i,g+1}}\mathbf{1}_{\{ i_{t'} \neq i^*_{t'}, N_{i,t_g:\tau_{i,g+1}} < 2N_{i,min_D}, \tau_{i,g}=\tau_{g+1}\}  } \leq 2 N_{i,\min_D} 
\end{align*}
\begin{align*}
\sum_{t'=t_{g}}^{\tau_{g+1}}\mathbf{1}_{\{ i_{t'} \neq i^*_{t'}, N_{i,t_g:\tau_{i,g+1}} < N_{i,min_D}, \tau_{i,g} < \tau_{g+1}  \} } \leq 
\max_{N_{1,t_{g}:\tau_{g+1}},...,N_{K,t_{g}:\tau_{g+1}}} \left\lbrace \sum_{i=1}^{K} N_{i,t_{g}:\tau_g} : N_{i,t_{g}:\tau_g} < 2N_{i,min_D} \right\rbrace.
\end{align*}

%\leq \sum_{i=1}^{K}N_{i,\min_D} = \sum_{i=1}^{K} \dfrac{6\log(t)}{(\Delta^{chg}_{\epsilon_0,g})^{2}}
%we denote the total pulls of arm $i$ as  $N_{i_{w},t_{g}:\tau_{i,g}}$ to detect a minimum gap of $\Delta_{\epsilon_0,g}$ with a probability of $\epsilon_0$ and let $\xi_{\epsilon_0,t}$ denote the bad event of not detecting the gap $\Delta_{\epsilon_0,g}$ till the $t$-th timestep. Hence, 
%
%\begin{align*}
%\underbrace{\sum_{t'=t_{g}}^{\tau_{i,g}}\mathbf{1}_{\left\lbrace\mathbb{I}_{t'}=i\in\argmax_{i\in\A}\big\lbrace\argmax_{k\in\A}\lbrace \Delta^{opt}_{k,g+1}\rbrace \bigcup \argmin_{k\in\A}\lbrace \Delta^{chg}_{k,g}\rbrace \bigcup \argmax_{k\in\A}\lbrace \frac{\Delta^{opt}_{k,g+1}}{\Delta^{chg}_{k,g}} \rbrace\big\rbrace \right\rbrace}}_{\textbf{part C}} \leq N_{i_{w},t_{g}:\tau_{i,g}} + \sum_{t'=t_g}^{\tau_{i,g}}\Pb\lbrace \xi_{\epsilon_0,t'}\rbrace.
%\end{align*}

For \textbf{part D} we can show that for $t'\in[t_{g}, \tau_{i,g}]$ or $t'\in[t_{g}, \tau'_{i,g}]$,
\begin{align*}
\left\lbrace i_{t'}\neq i^*_{t'},N_{i,t_{g}:\tau_{i,g}}\geq  N_{\min_D}\right\rbrace \subseteq \lbrace \xi^{chg}_{i,\tau_{i,g}}  \rbrace \bigcup \lbrace (\mu_{i^*,{g}} - \mu_{i,t_{g}}) < 2S_{i,t_{g}:\tau_{i^*,g}} \rbrace.
\end{align*}
But from step 5 we know that for $N_{i,\tau_{g-1}:\tau_{g}}\geq N_{i,\min_D}$ then the event $ (\mu_{i^*,{g}} - \mu_{i,g}) < 2S_{i,t_{g}:\tau_{i,g}}$ is not possible. Hence, we can show that \textbf{part D} is upper bounded by,
\begin{align*}
\underbrace{\sum_{t'=t_{g}}^{\tau_{i,g}}\mathbf{1}_{\lbrace i_{t'} \neq i^*_{t'},N_{i,t_{g}:\tau_{i,g}}\geq N_{i,\min_D}\rbrace}}_{\textbf{part D}} \leq N_{i,\min_D} + \sum_{t'=t_{g}}^{\tau_{i,g}}\xi^{chg}_{i,t'} 
&\leq N_{i,\min_D} + \sum_{t'=\tau_{i,g-1}}^{\tau_{i,g}}\Pb\lbrace\xi^{chg}_{i,t'}\rbrace \\
%%%%%%%%%%%%%%%%%%%%%%%%%
&= \dfrac{2\log(\frac{4t^2}{\delta})}{(\Delta^{chg}_{i,g})^2} + \sum_{t'=\tau_{i,g-1}}^{\tau_{i,g}}\Pb\lbrace\xi^{chg}_{i,t'}\rbrace.
\end{align*}
Finally, for the \textbf{part E} we know from Assumption \ref{assm:space-gap} and Lemma \ref{psbandit:Lemma:01} that a good detection policy using only mean estimation suffers a maximal delay of $d_{\pi}(t_g - t_{g-1})$ with $(1-\delta)$ probability, where the probability for the detection policy of not detecting a changepoint of magnitude $\Delta(t_g,\delta)$ is bounded by $1 - \delta$ probability. Let $\tilde{\xi}^{chg}_{g,t}$ denote the probability that the changepoint $g$ is not detected till time $t$. Let, $d_{\pi}(t_g - \tau_{g-1})$ denote the maximal delay of \UCBLCPD for the piece $\rho_g$ for not detecting the changepoint $g$.
\begin{align*}
\underbrace{\sum_{t'=\tau_{g-1}}^{t}\mathbf{1}_{\{ i_{t}' \neq i*_{t'},t > \tau_{g-1} > t_{g-1}}  \}}_{\textbf{part E}} \leq d_{\pi}(t_g - \tau_{g-1}) + \sum_{t'=\tau_{g-1}}^{t}\Pb\lbrace\tilde{\xi}^{chg}_{g,t} \rbrace.
\end{align*}

\textbf{Step 6.(Maximal Delay $d_{\pi}(t_g - \tau_{g-1})$ of \UCBLCPD):}  The delay $d_{\pi}(t_g - \tau_{g-1})$ for the detection policy $\pi$ can be upper bounded as,
\begin{align*}
	d_{\pi}(t_g - \tau_{g-1}) &= d_{\pi}(t_g - (t_{g-1} + d_{\pi}(t_{g-1} - \tau_{g-2}))) =\ldots \\ &= d_{\pi}(t_g - (t_{g-1} + d_{\pi}(t_{g-1} - \ldots (t_1 + d_{\pi}(t_2 - \tau_1))))
\end{align*}
Note that $d_{\pi}(t_2 - \tau_1) = d_{\pi}(t_2 - (t_0 + d_{\pi}(t_1 - t_0)))$ is the delay in detecting the first changepoint $t_1$, $\tau_0 = t_0$ and $d_{\pi}(t_1 - t_0) = 0$. We proceed as like Lemma \ref{psbandit:Lemma:01} to bound $d_{\pi}(t_2 - \tau_1)$. We know from Lemma \ref{psbandit:Lemma:11} that a minimum sample of $N_{i,\min_D}$ is sufficient for an arm $i\in\A$ to detect a changepoint of gap $\Delta^{chg}_{i, g}$ with $(1-\delta)$ probability. A changepoint detection policy may pull each arm $i=1,\ldots, K$, at most $N_{i,\min_D} -1$ times before finally detecting the changepoint for the $K$-th arm. Hence, 
\begin{align*}
	d_{\pi}(t_2 - \tau_{1}) &\leq (K-1)\left(N_{i,\min_{D}}  +  \sum_{t'=N_{i,\min_{D}}}^{t_{2}}\Pb\lbrace \xi^{chg}_{i,t'}\rbrace\right) + N_{i,\min_{D}}  +  \sum_{t'=N_{i,\min_{D}}}^{t_{2}}\Pb\lbrace \xi^{chg}_{i,t'}\rbrace \\
	& < K N_{i,\min_D} + K\sum_{t'=N_{i,\min_{D}}}^{t_{2}}\Pb\lbrace \xi^{chg}_{i,t'}\rbrace = \dfrac{6K\log(t)}{(\Delta^{chg}_{i,1})^2} +  K\sum_{t'=N_{i,\min_{D}}}^{t_{2}}\Pb\lbrace \xi^{chg}_{i,t'}\rbrace.
\end{align*}		
Similarly we can show that for $d_{\pi}(t_3 - \tau_{2})$,
\begin{align*}
d_{\pi}(t_3 - \tau_{2}) &= d_{\pi}(t_3 - (t_2 + d_{\pi}(t_2 -\tau_1)) \\
%%%%%%%%%%%%%%%%%%
&\leq  \underbrace{\dfrac{6K\log(t)}{(\Delta^{chg}_{i,1})^2} +  K\sum_{t'=N_{i,\min_{D}}}^{t_{2}}\Pb\lbrace \xi^{chg}_{i,t'}\rbrace}_{\text{$t_1$ detection delay}} + \underbrace{\dfrac{6K\log(t)}{(\Delta^{chg}_{i,2})^2} +  K\sum_{t'=N_{i,\min_{D}}}^{t_{2}}\Pb\lbrace \xi^{chg}_{i,t'}\rbrace}_{\text{$t_2$ detection delay}}\\
%%%%%%%%%%%%%%%%%%
&\overset{(a)}{\leq} \dfrac{12K\log(\frac{4t^2}{\delta})}{(\Delta(t_g,\delta))^2} +  2K\sum_{t'=N_{i,\min_{D}}}^{t_{2}}\Pb\lbrace \xi^{chg}_{i,t'}\rbrace
\end{align*}
where, $(a)$ happens because $\forall i\in\A, g\in\G,\Delta^{chg}_{i,g} \geq \Delta(t_g,\delta)$. From Lemma \ref{psbandit:Lemma:01} we know that if a changepoint detection policy $\pi^*$ has observation from $t_{g-1}$ till $t_g < t <t_{g+1}$ then it needs a minimum sample of $n(t_g,\Delta,\delta)$ to detect a deviation of magnitude $\Delta$ with $(1-\delta)$ probability at $t_g$. This is the maximal delay of $\pi^\ast$ is denoted by $d_{\pi^\ast}(t_g - t_{g-1}) \leq  \left( \frac{C(t, \delta, \eta)K\log(\frac{t^2}{{\delta}})}{2(\Delta(t_g, \delta))^{2}}\right)$ where, $C(t, \delta, \eta) \leq \eta \log (t/\delta)$, and $\eta \in (0,1)$. Now, from Assumption \ref{assm:space-gap} we know that $t_g  +  d_{\pi^*}\left(t_{g}  -  t_{g-1}\right)  \leq  t_g +   \eta (t_{g+1}-t_g)$. So, \UCBLCPD cannot do any better than $d_{\pi^\ast}(t_g - t_{g-1})$ as it lacks observations from $t_{g-1}$ itself. But it suffices to show that atleast $d_{\pi}(t_g - \tau_{g-1})$ will not be much greater than $d_{\pi^\ast}(t_g - t_{g-1})$. Starting with the base case, the first detection delay of $d_{\pi}(t_2 - \tau_1)$ we can show that for $\delta = \frac{1}{t}$ and $\forall i\in\A_g^{chg}, g\in\G,\Delta^{chg}_{i,g} \geq \Delta(t_g, \delta) = \Omega\sqrt{\frac{\log(t)}{t}}$,
%\sqrt{\frac{K\log(T/G)}{T/G}}$,
\begin{align*}
& \left( \dfrac{12K\log(t)}{(\Delta(t_g,\delta))^2} +  2K\sum_{t'=N_{i,\min_{D}}}^{t_{2}}\Pb\lbrace \xi^{chg}_{i,t'}\rbrace \right) -  \left( \frac{C(t,\delta,\eta)K\log(\frac{t}{\delta})}{2(\Delta(t_g,\delta))^{2}}\right)  \leq \eta (t_3 - t_2)\\
%%%%%%%%%%%%%%%%
& \Leftrightarrow \left(\dfrac{12K\log(t)}{(\Delta(t_g,\delta))^2} +  2K\sum_{t'=N_{i,\min_{D}}}^{t_{2}}\delta\right) \leq \left( \frac{\eta\log (t/\delta)K\log(\frac{t}{\delta})}{2(\Delta(t_g,\delta))^{2}}\right) + \eta(t_3 - t_2)\\
& \Leftrightarrow \eta \geq  \dfrac{6 t}{2t\log t + t}.
\end{align*}
Now, we can show that for the $g$-th changepoint,
\begin{align*}
d_{\pi}(t_g - \tau_{g-1}) &= d_{\pi}(t_g - (t_{g-1} + d_{\pi}(t_{g-1} - \tau_{g-2}))) = \ldots \\
%%%%%%%%%%%%%%%%%%%%%
&= d_{\pi}(t_g - (t_{g-1} + d_{\pi}(t_{g-1} - \ldots (t_1 + d_{\pi}(t_2 - \tau_1))))\\
%%%%%%%%%%%%
& \overset{(a)}{\leq} d_{\pi}(t_g - (t_{g-1} + d_{\pi}(t_{g-1} - \tau_{g-2}))) \leq d_{\pi}(t_{g-1} - \tau_{g-2}) + d_{\pi}(t_g - \tau_{g-1})
\end{align*}
where $(a)$ holds for each changepoint from $1,\ldots,g-1$ by Assumption \ref{assm:space-gap} when $\eta \geq \frac{6}{\log t + 1}$. Proceeding similarly, we can show that,
%\begin{align*}
$d_{\pi}(t_{g-1} - \tau_{g-2}) + d_{\pi}(t_g - \tau_{g-1}) \leq \left(\dfrac{12K\log(t)}{(\Delta(t_g,\delta))^2} +  2K\sum_{t'=N_{i,\min_{D}}}^{t_{g}}\delta\right)$ 
%\end{align*}  
with high probability for $\eta \geq \frac{6}{2\log t + 1}$. 
%\begin{align*}
%& t_1 + d_{\pi}(t_2 - \tau_1) \leq t_1 + \eta (t_2 - t_1)\\
%& t_1 + d_{\pi}(t_2 - \tau_1) \leq t_1 + \eta (t_2 - t_1)\\
%&\Leftrightarrow d_{\pi}(t_2 - \tau_1) \leq \eta (t_2 - t_1)\\
%&\Leftrightarrow KN_{i,\min_D} + K\sum_{t'=N_{i,min_D}}^{t}\Pb\lbrace \xi^{chg}_{i,t'}\rbrace \leq \eta \frac{T}{G}\\
%&\Leftrightarrow \dfrac{2K\log(\frac{4t^2}{\delta})}{(\Delta^{chg}_{i,g})^2} + K\sum_{t'=N_{i,min_D}}^{t}\delta \leq \eta \frac{T}{G}\\
%&\Leftrightarrow  \dfrac{2K\log(\frac{4t^2}{\delta})}{K\log(T/G)}\frac{T}{G} + K\sum_{t'=N_{i,min_D}}^{t}\delta \leq \eta \frac{T}{G}
%\end{align*}
%
%
%Substituting the value of

%Furthermore, we can see that for $\eta \geq K$ the maximal delay of UCB-CPD is surely less than $d(t_g - \tau_{g-1})$ in Lemma \ref{psbandit:Lemma:01}.
%
%\begin{align*}
%\dfrac{2K\log(\frac{4t^2}{\delta})}{(\Delta^{chg}_{i,g})^2} < \dfrac{2\eta K\log(\frac{4t^2}{\delta})}{(\Delta^{chg}_{\epsilon_0,g})^2} < d(t_g - t_{g-1}) \leq  \left( \frac{C_{\eta}K\log(\frac{T^2}{G^2{\epsilon_0}})}{2(\Delta^{}_{\epsilon_0, g})^{2}}\right)
%\end{align*}
%where, $C_{\eta} \leq 4K\log (T/G)$.
%
%Similarly, we can argue that for each arm $i=1,2,\ldots,K$, the probability of not detecting the changepoint from time $\tau_{g-1}$ till $t$ is upper bounded by,
%
%\begin{align*}
%\sum_{t'=\tau_{g-1}}^{t}\Pb\lbrace\tilde{\xi}^{chg}_{g,t} \rbrace \leq \sum_{t'=t_{g}}^{t}K\delta
%\end{align*}

\textbf{Step 7.(Bound the expected regret):} Now, taking expectation over $N_{i,1:t}$, and considering the contributions form $\textbf{parts A, B, C, D, E}$ we can show that,
\begin{align*}
& \E[N_{i,1:t}] \leq \sum_{j=1}^{G}\bigg[ \underbrace{ 1 \! + \! N_{i,\min_{E}} \! + \!\!\!\!\! \sum_{t'=N_{i,\min_{E}}}^{t_{g}}\Pb\lbrace \xi^{opt}_{i,t'} \rbrace}_{ \textbf{ \scriptsize A1)  expected pulls from $\tau_{i,g-1}:t_{g}$} \atop \textbf{\scriptsize for optimality detection}} \!\!\!+\! \underbrace{N_{i,\min_{D}}  +  \sum_{t'=N_{i,\min_{D}}}^{t_{g}}\Pb\lbrace \xi^{chg}_{i,t'}\rbrace}_{\textbf{\scriptsize B1) expected pulls from $\tau_{i,g-1}:t_{g}$} \atop\textbf{\scriptsize for changepoint detection}}  \\
%%%%%%%%%%%%%%%%%%%%%%%%%
 &+ \underbrace{ 3N_{i,\min_{D}} + \max_{N_{1,t_{g}:\tau_{g+1}},...,N_{K,t_{g}:\tau_{g+1}}} \left\lbrace \sum_{i=1}^{K} N_{i,t_{g}:\tau_{g+1}} : N_{i,t_{g}:\tau_{g+1}} < 3N_{i,min_D} \right\rbrace}_{\textbf{C1) worst case expected pulls from $t_{g}:\tau_{i,g}$}} \! \\
 %%%%%%%%%%%%%%%%%%%
 &+ \underbrace{ N_{i,\min_{D}} + \sum_{t'=N_{i,\min_{D}}}^{\tau_{i,g}}\Pb\lbrace \xi^{chg}_{i,t'}\rbrace}_{\textbf{D1) expected pulls from $t_{g}:\tau_{i,g}$ for changepoint detection}} \bigg]
%%%%%%%%%%%%%%%%%%%%%%%%%
+ 2\underbrace{K N_{i,\min_D} + 2\sum_{t'=\tau_{g-1}}^{t}\Pb\lbrace\tilde{\xi}^{chg}_{g,t}\rbrace}_{\textbf{E1) pulls for not detecting g from $t_g$ to $t$}}
\end{align*}
Again, for \textbf{part C1} we can upper bound it as, we use the assumption \ref{assm:space-gap}, and assumption \ref{assm:chg-gap} to upper bound it as,
\begin{align*}
3N_{i,\min_D} &+ \max_{N_{1,t_{g}:\tau_g},...,N_{K,t_{g}:\tau_{g+1}}} \left\lbrace \sum_{i=1}^{K} N_{i,t_{g}:\tau_g} : N_{i,t_{g}:\tau_g} < 3N_{i,min_D} \right\rbrace \\
%%%%%%%%%%%%%%%%%%%%%%%%
 & {\leq} \dfrac{18\log(t)}{(\Delta^{chg}_{i,g})^{2}} + \max_{N_{1,t_{g}:\tau_{g+1}},...,N_{K,t_{g}:\tau_{g+1}}} \left\lbrace \sum_{i=1}^{K} N_{i,t_{g}:\tau_{g+1}} : N_{i,t_{g}:\tau_{g+1}} < 3N_{i,min_D} \right\rbrace\\
%%%%%%%%%%%%%%%%%%%%%%%%
& \overset{(a)}{\leq} \dfrac{18\log(t)}{(\Delta^{chg}_{i,g})^{2}} + \max_{N_{1,t_{g}:\tau_g},...,N_{K,t_{g}:\tau_g}} \left\lbrace \sum_{i=1}^{K} N_{i,t_{g}:\tau_g} : N_{i,t_{g}:\tau_g} < 3N_{i,min_D} \right\rbrace. \\
\end{align*}
%&\leq N_{i,\min_D} + \sum_{i=1}^{K}N_{i,\min_D}\\
%\begin{align*}
%\underbrace{N_{i_{w},t_{g}:\tau_{i,g}} + \sum_{t'=t_g}^{\tau_{i,g}}\Pb\lbrace \xi_{\epsilon_0,g}\rbrace }_{\textbf{C1) worst case expected pulls from $t_{g}:\tau_{i,g}$}} &\leq  N_{i,\min_D} + \sum_{t'=t_g}^{\tau_{i,g}}\Pb\lbrace \xi_{\epsilon_0,t'}\rbrace \\
%%%%%%%%%%%%%%%%%%%%%%
%& \leq  \dfrac{2\log(\frac{4t^2}{\delta})}{(\Delta^{chg}_{\epsilon_0,g})^{2}} + \sum_{t'=t_g}^{\tau_{i,g}}\Pb\lbrace \xi_{i,t'}^{chg}\rbrace  \\
%%%%%%%%%%%%%%%%%%%%%%
%& \overset{(a)}{\leq} \dfrac{6\log(t)}{(\Delta^{chg}_{\epsilon_0,g})^{2}} + \dfrac{\pi^2}{3}.
%\end{align*}
Here in $(a)$ we substitute the value of $\delta=\frac{1}{t}$.  Substituting these in $\E[N_{i,1:t}]$ we get,
\begin{align*}
&\E[N_{i,1:t}] \overset{(a)}{\leq} \sum_{j=1}^{G}\bigg[\underbrace{1+ \dfrac{6\log t}{(\Delta^{opt}_{i,g})^2} + \sum_{t'=N_{i,\min_{E}}}^{t_{g}}2\delta}_{\textbf{part A1}} + \underbrace{ \dfrac{6\log(t)}{(\Delta^{chg}_{i,g})^2} + \sum_{t'=\tau_{i,g-1}+K+1}^{t_{g}}2\delta }_{\textbf{part B1}} + \underbrace{\dfrac{6\log(t)}{(\Delta^{chg}_{i,g})^2} + \sum_{t'=\tau_{i,g}+K+1}^{\tau_{i,g}}2\delta}_{\textbf{part D1}} \\
%%%%%%%%%%%%%%%%%%%%%%%%
& + \underbrace{ \dfrac{18\log(t)}{(\Delta^{chg}_{i,g})^{2}} + \max_{N_{1,t_{g}:\tau_g},...,N_{K,t_{g}:\tau_g}} \left\lbrace \sum_{i=1}^{K} N_{i,t_{g}:\tau_g} : N_{i,t_{g}:\tau_g} < 3N_{i,min_D} \right\rbrace}_{\textbf{part C1}} + \underbrace{\dfrac{12K\log(t)}{(\Delta^{chg}_{i,g})^2} + \sum_{t'=\tau_{g-1}}^{t}2K\delta}_{\textbf{part E1}}\bigg]\\
%%%%%%%%%%%%%%%%%%%%%%%%%
&\overset{(b)}{\leq} \sum_{j=1}^{G}\bigg[1+ \dfrac{6\log t}{(\Delta^{opt}_{i,g})^2} + \sum_{t'=N_{i,\min_{E}}}^{t_{g}}\dfrac{8}{t'} + \dfrac{12\log(t)}{(\Delta^{chg}_{i,g})^2} + \sum_{t'=\tau_{i,g-1}+K+1}^{t_{g}}\dfrac{8}{t'} \\
%%%%%%%%%%%%%%%%%%%%%%%%%
&+ \dfrac{18\Delta^{opt}_{\max,g+1}\log(t)}{(\Delta(t_g, \delta))^2)^{2}}
%%%%%%%%%%%%%%%%%%%%%%%%%
+ \dfrac{18\log(t)}{(\Delta^{chg}_{i,g})^2} + \sum_{t'=\tau_{i,g-1}+K+1}^{\tau_{i,g}}\dfrac{8}{t'} + \dfrac{12K\Delta^{opt}_{i,g+1}\log(t)}{(\Delta(t_g, \delta))^2} + K\sum_{t'=\tau_{g-1}}^{t}\dfrac{8}{t'}\bigg]\\
\end{align*}
where, $(a)$ comes from the result of Lemma \ref{psbandit:Lemma:1} and in $(b)$ we substitute $\delta=\frac{4}{t}$. Hence, the regret upper bound is given by summing over all arms in $\A$ and considering $\Delta^{opt}_{i,g} \leq \Delta^{opt}_{\max,g+1}$ and considering that $\pi$ suffers the repeating worst case of undetectable gaps as,
\begin{align*}
\E[R_t] & \overset{}{\leq} \sum_{g=1}^{G}\sum_{i=1}^{K} \bigg\{ 25 + \dfrac{6\log(t)}{\Delta^{opt}_{i,g}}\bigg\} \! +\! \sum_{g=1}^{G}\sum_{i\in \A_g^{chg}}\!\bigg\{ \dfrac{12\Delta^{opt}_{i,g}\log(t)}{(\Delta^{chg}_{i,g})^2} + \dfrac{18\Delta^{opt}_{\max,g+1}\log(t)}{(\Delta(t_g,\delta))^{2}} \\
%%%%%%%%%%%%%%%%%%%%
&+ \dfrac{12K\Delta^{opt}_{\max,g+1}\log(t)}{(\Delta(t_g, \delta))^2} + 8K\!\bigg\rbrace  
+ \max_{i\in\A: \!\!\frac{e}{\sqrt{T}}\leq \! \Delta_i <\! \Delta(t_g,\delta)}\Delta_i T
\end{align*}
%Here, in $(a)$ we use the Assumption \ref{assm:space-gap}, Assumption \ref{assm:chg-gap} and Discussion \ref{dis:gap-delay} to substitute $\Delta_{\epsilon_0,g}$ as the minimal gap that can be detected when $t_g > \tau_{i,g}$. In $(b)$ we substitute $\delta=\dfrac{1}{t}$.
\end{customproof}

%and maximally bound the regret for not detecting $\Delta^{opt}_{\max}$ by invoking Lemma \ref{psbandit:Lemma:01} when $\Delta^{opt}_{\max}$ scales as $\sqrt{\dfrac{K\log (T/G)}{T/G}}$.

\section{Proof of Regret Bound of \ImpCPD}
\label{sec:proof:Theorem:2}

\begin{customproof}{6}
\label{proof:Theorem:2}

We proceed as like Theorem 3.1 in \citet{auer2010ucb} and we combine the changepoint detection sub-routine with this. 

%\textbf{Step 0.(Vital Assumption for proving):} In this proof, we assume that all the changepoints are detected by the algorithm with some delay. Since, all the gaps are significant (Assumption \ref{assm:chg-gap}) and there is sufficient delay between two changepoints (Assumption \ref{assm:space-gap}), we can take this assumption for proving the result.

\textbf{Step 1.(Define some notations):} In this proof, we use $N_{i,g}$ instead of $N_{i,t_s:t_p}$ denoting the number of times the arm $i$ is pulled between two restarts, ie, between $t_s$ to $t_p$ (as shown in algorithm) for the piece $\rho_{g}$. We define the confidence interval for the $i$-th arm as $S_{i,g}=\sqrt{\frac{\alpha\log(\psi \epsilon_m^2)}{N_{i,g}}}$ and $\psi =\frac{T^2}{K^2\log K}$. The phase numbers are denoted by $m=0,1,\ldots,M$ where $M=\frac{1}{2}\log_{1+\gamma}\frac{T}{e}$. For each sub-optimal arm $i$, we consider their contribution to cumulative regret individually till each of the changepoint is detected. We also define $\A'=\big\lbrace i\in\A: \Delta^{opt}_{i,g}\geq \sqrt{\frac{e}{T}}, \Delta^{chg}_{i,g}\geq \sqrt{\frac{e}{T}},\forall g\in\G \big\rbrace$.

\textbf{Step 2.(Define an optimality stopping phase $m_{i,g}$):} We define an optimality stopping phase $m_{i,g}$ for a sub-optimal arm $i\in\A'$ as the first phase after which the arm $i$ is no longer pulled till the $g$-th changepoint is detected such that for $0<\gamma\leq 1$,
\begin{align*}
m_{i,g} = \min\left\lbrace m: \sqrt{\alpha\epsilon_{m}} < \frac{\Delta^{opt}_{i,g}}{4(1+\gamma)^2} \right\rbrace
\end{align*} 

\textbf{Step 3.(Define a changepoint stopping phase $p_{i,g}$):} We define a changepoint stopping phase $p_{i,g}$ for an arm $i\in\A'$ such that it is the first phase when the changepoint $g$ is detected such that for $0<\gamma\leq 1$,
\begin{align*}
p_{i,g} = \min\left\lbrace m: \sqrt{\alpha\epsilon_{m}} < \frac{\Delta^{chg}_{i,g}}{4(1+\gamma)^2} \right\rbrace
\end{align*}

\textbf{Step 4.(Regret for sub-optimal arm $i$ being pulled after $m_{i,g}$-th phase):} Note, that on or after the $m_{i,g}$-th phase a sub-optimal arm $i\in\A'$ is not pulled anymore if these four conditions hold,
\begin{eqnarray}
\hat{\mu}_{i,g} < \mu_{i,g} + S_{i,g}, \hspace*{2em}  \hat{\mu}_{i*,g} > \mu_{i*,g} - S_{i*,g}, \hspace*{2em} S_{i,g} > S_{i^* ,g}, \hspace*{2em} N_{i,g} \geq \ell_{m_{i,g}} \label{eq:arm-pull-opt}
\end{eqnarray}
Also, in the $m_{i,g}$-th phase if $N_{i,g} \geq \ell_{m_{i,g}} = \dfrac{\log(\psi \epsilon_{m_{i,g}}^2)}{2\epsilon_{m_{i,g}}}$ then we can show that,
\begin{align*}
S_{i,g} = \sqrt{\dfrac{\alpha\log(\psi \epsilon_{m_{i,g}}^2)}{2N_{i,g}}} \leq \sqrt{\dfrac{\alpha\log(\psi \epsilon_{m_{i,g}}^2)}{2\ell_{m_{i,g}}}} \leq \sqrt{\alpha\epsilon_{m_{i,g}}\dfrac{\log(\psi \epsilon_{m_{i,g}}^2)}{\log(\psi \epsilon_{m_{i,g}}^2)}} \leq \dfrac{\Delta^{opt}_{i,g}}{4(1+\gamma)^2}.
\end{align*}
If indeed the four conditions in equation \eqref{eq:arm-pull-opt} hold then we can show that in the $m_{i,g}$-th phase,
\begin{align*}
\hat{\mu}_{i,g} + S_{i,g} \leq {\mu}_{i,g} + 4(1+\gamma)^2 S_{i,g} - 2(1+\gamma)^2S_{i,g} &\leq {\mu}_{i,g} + \Delta^{opt}_{i,g} - 2(1+\gamma)^2S_{i,g} \\
%%%%%%%%%%%%%%%
&\leq {\mu}_{i*,g} - 2(1+\gamma)^2S_{i^*,g} \\
%%%%%%%%%%%%%%%
&\leq \hat{\mu}_{i*,g} - S_{i^* , g}
\end{align*}
Hence, the sub-optimal arm $i$ is no longer pulled on or after the $m_{i,g}$-th phase. Therefore, to bound the number of pulls of the sub-optimal arm $i$, we need to bound the probability of the complementary of the four events in equation \eqref{eq:arm-pull-chg}.

For the first event in equation \eqref{eq:arm-pull-opt}, using Chernoff-Hoeffding bound we can upper bound the probability of the complementary of that event by,
\begin{align*}
&\sum_{m=0}^{m_{i,g}}\sum_{n=1}^{\ell_{m_{}}}\Pb\bigg( \dfrac{\sum_{s=1}^{n} X_{i,s}}{n} \geq  \mu_{i,g} + \sqrt{\dfrac{\alpha\log(\psi \epsilon_{m_{}}^2)}{2n}} \bigg)  \leq \sum_{m=0}^{m_{i,g}}\sum_{n=1}^{\ell_{m_{}}}\exp\left(-2 \left(\sqrt{\dfrac{\alpha\log(\psi \epsilon_{m_{}}^2)}{2n}}\right)^2 n \right)\\
%%%%%%%%%%%%%%%%%%%%%%%%%%%%%%%%%%
&\leq \sum_{m=0}^{m_{i,g}}\sum_{n=1}^{\ell_{m_{}}}\exp\left(-2\dfrac{\alpha\log(\psi \epsilon_{m_{}}^2)}{2n}n \right)
%%%%%%%%%%%%%%%%%%%%%%%%%%%%%%%%%%
%&\leq \sum_{m=0}^{m_{i,g}}\sum_{n=1}^{\ell_{m_{}}}\dfrac{1}{\psi \epsilon_{m_{}}^2} \\
%%%%%%%%%%%%%%%%%%%%%%%%%%%%%%%%%%
\leq \sum_{m=0}^{m_{i,g}}\dfrac{\log(\psi \epsilon_{m_{}}^2)}{2\epsilon_{m_{}}}\dfrac{1}{(\psi \epsilon_{m_{}}^2)^{\alpha}} \\
%%%%%%%%%%%%%%%%%%%%%%%%%%%%%%%%%%
& \leq  \dfrac{\log(\psi  \sum_{m=0}^{m_{i,g}}\epsilon_{m}^2)}{2(\psi )^{\alpha}}\sum_{m=0}^{m_{i,g}}\dfrac{1}{\epsilon_{m_{}}^{2\alpha +1}} 
%%%%%%%%%%%%%%%%%%%%%%%%%%%%%%%%%%
 \leq  \dfrac{\log(\psi )}{2(\psi )^{\alpha}}\sum_{m=0}^{m_{i,g}}\dfrac{1}{\epsilon_{m_{}}^{2\alpha +1}}.
\end{align*}
Similarly, for the second event in equation \eqref{eq:arm-pull-opt}, we can bound the probability of its complementary event by,
\begin{align*}
\sum_{m=0}^{m_{i,g}}\sum_{n =1}^{\ell_{m_{}}}\Pb\bigg\lbrace \dfrac{\sum_{s=1}^{n} X_{i^*,s}}{n} \leq  \mu_{i*,g} - \sqrt{\dfrac{\alpha\log(\psi \epsilon_{m_{}}^2)}{2n}} \bigg\rbrace &\leq \sum_{m=0}^{m_{i,g}}\sum_{n=1}^{\ell_{m_{}}}\exp\left(-2\left(\sqrt{\dfrac{\alpha\log(\psi \epsilon_{m_{}}^2)}{2n}}\right)^2n_{} \right)\\
%%%%%%%%%%%%%%%%%%%%%%%%%%%%%%%%%%%%%%%%%
&\leq \dfrac{\log(\psi )}{2(\psi )^{\alpha}}\sum_{m=0}^{m_{i,g}}\dfrac{1}{\epsilon_{m_{}}^{2\alpha +1}}.
\end{align*}
Also, for the third event in equation \eqref{eq:arm-pull-opt}, we can bound the probability of its complementary event by,
\begin{align*}
\sum_{m=0}^{m_{i,g}}\Pb\lbrace S_{i,g} < S_{i*,g} \rbrace &\leq \sum_{m=0}^{m_{i,g}}\Pb \lbrace \hat{\mu}_{i,g} + S_{i,g} > \hat{\mu}_{i*,g} + S_{i*,g}\rbrace \\
%%%%%%%%%%%%%%%%%%%%%%%%%%%%%%%%%%5
\end{align*}
%\begin{align*}
%\sum_{m=0}^{m_{i,g}}\Pb\lbrace S_{i,g} < S_{i*,g} \rbrace & \leq \sum_{m=0}^{m_{i,g}}\sum_{n=1}^{\ell_{m_{}}}\sum_{q=1}^{\ell_{m_{}}}\Pb\bigg\lbrace  \dfrac{\sum_{s=1}^{n} X_{i,s}}{n} + \sqrt{\dfrac{\alpha\log(\psi T\epsilon_{m_{}}^2)}{2n}} >  \dfrac{\sum_{s=1}^{q} X_{i^*,s}}{q} + \sqrt{\dfrac{\alpha\log(\psi T\epsilon_{m_{}}^2)}{2q}} \bigg\rbrace \\
%%%%%%%%%%%%%%%%%%%%%%%%%%%%%%%%%%%%%%%%%%
%& \leq \sum_{m=0}^{m_{i,g}}\sum_{n =1}^{\ell_{m_{}}}\Pb\bigg\lbrace  \dfrac{\sum_{s=1}^{n} X_{i,s}}{n} \geq  \mu_{i,g} + \sqrt{\dfrac{\alpha\log(\psi T\epsilon_{m_{}}^2)}{2n}} \bigg\rbrace +  \sum_{m=0}^{m_{i,g}}\sum_{q =1}^{\ell_{m_{}}}\Pb\bigg\lbrace  \dfrac{\sum_{s=1}^{q} X_{i^*,s}}{q} \leq  \mu_{i^*,g} - \sqrt{\dfrac{\alpha\log(\psi T\epsilon_{m_{}}^2)}{2q}} \bigg\rbrace\\
%%%%%%%%%%%%%%%%%%%%%%%%%%%%%%%%%%%%%%%%%%
%&\leq \dfrac{\log(\psi T )}{2(\psi T)^{\alpha}}\sum_{m=0}^{m_{i,g}}\dfrac{1}{\epsilon_{m_{}}^{2\alpha +1}}
%\end{align*}
%
%%
But the event $\hat{\mu}_{i,g} + S_{i,g} > \hat{\mu}_{i*,g} + S_{i*,g}$ is possible only when the following three events occur, $\lbrace \hat{\mu}_{i*,g} \leq \mu_{i*,g} - S_{i*,g} \rbrace \cup \lbrace\hat{\mu}_{i,g} \geq \mu_{i,g} + S_{i,g}\rbrace \cup \lbrace \mu_{i*,g}-\mu_{i,g} < 2(1+\gamma)^2 S_{i,g} \rbrace$. But the third event will not happen with high probability for $N_{i,g}\geq \ell_{m_{i,g}}$. Proceeding as before, we can show that the probability of the remaining two events is bounded by,
\begin{align*}
&\sum_{m=0}^{m_{i,g}}\Pb\lbrace S_{i,g}  < S_{i*,g} \rbrace  \leq \sum_{m=0}^{m_{i,g}}\sum_{n=1}^{\ell_{m_{}}}\sum_{q=1}^{\ell_{m_{}}}\Pb\bigg\lbrace  \dfrac{\sum_{s=1}^{n} X_{i,s}}{n} + \sqrt{\dfrac{\alpha\log(\psi \epsilon_{m_{}}^2)}{2n}} >  \dfrac{\sum_{s=1}^{q} X_{i^*,s}}{q} + \sqrt{\dfrac{\alpha\log(\psi \epsilon_{m_{}}^2)}{2q}} \bigg\rbrace \\
%%%%%%%%%%%%%%%%%%%%%%%%%%%%%%%%%%%%%%%%%
& \leq \sum_{m=0}^{m_{i,g}}\sum_{n =1}^{\ell_{m_{}}}\Pb\bigg\lbrace  \dfrac{\sum_{s=1}^{n} X_{i,s}}{n} \geq  \mu_{i,g} + \sqrt{\dfrac{\alpha\log(\psi \epsilon_{m_{}}^2)}{2n}} \bigg\rbrace 
%%%%%%%%%%%%%%%%%%%%
\!\! + \!\! \sum_{m=0}^{m_{i,g}}\sum_{q =1}^{\ell_{m_{}}}\Pb\bigg\lbrace  \dfrac{\sum_{s=1}^{q} X_{i^*,s}}{q} \leq  \mu_{i^*,g} - \sqrt{\dfrac{\alpha\log(\psi \epsilon_{m_{}}^2)}{2q}} \bigg\rbrace\\
%%%%%%%%%%%%%%%%%%%%%%%%%%%%%%%%%%%%%%%%%
&\leq \sum_{m=0}^{m_{i,g}}\sum_{n=1}^{\ell_{m_{}}}\exp\left(-2\dfrac{\alpha\log(\psi \epsilon_{m_{}}^2)}{2n}n \right) + \sum_{m=0}^{m_{i,g}}\sum_{q=1}^{\ell_{m_{}}}\exp\left(-2\dfrac{\alpha\log(\psi \epsilon_{m_{}}^2)}{2q}q \right)
%%%%%%%%%%%%%%%%%%%%%%%%%%%%%%%%%%%%%%%%%
\leq \dfrac{\log(\psi  )}{2(\psi )^{\alpha}}\sum_{m=0}^{m_{i,g}}\dfrac{1}{\epsilon_{m_{}}^{2\alpha +1}}.
\end{align*}
Finally for the fourth event in equation \eqref{eq:arm-pull-opt}, we can bound the probability of its complementary event by following the same steps as above,
\begin{align*}
\sum_{m=0}^{m_{i,g}}\Pb\lbrace N_{i,g} &< \ell_{m_{i,g}}\rbrace \leq \sum_{m=0}^{m_{i,g}}\Pb\lbrace \hat{\mu}_{i,g} + S_{i,g} < \hat{\mu}_{i*,g} + S_{i*,g} \rbrace\\
%%%%%%%%%%%%%%%%%%%%%%%%%%
&\leq \sum_{m=0}^{m_{i,g}}\sum_{n=1}^{\ell_{m_{}}}\sum_{q=1}^{\ell_{m_{}}}\Pb\bigg\lbrace  \dfrac{\sum_{s=1}^{n} X_{i,s}}{n} + \sqrt{\dfrac{\alpha\log(\psi \epsilon_{m_{}}^2)}{2n}} <  \dfrac{\sum_{s=1}^{q} X_{i^*,s}}{q} + \sqrt{\dfrac{\alpha\log(\psi \epsilon_{m_{}}^2)}{2q}} \bigg\rbrace \\
%%%%%%%%%%%%%%%%%%%%%%%%%%
&\leq \dfrac{\log(\psi )}{(\psi )^{\alpha}}\sum_{m=0}^{m_{i,g}}\dfrac{1}{\epsilon_{m}^{2\alpha +1}}.
\end{align*}
Combining the above four cases we can bound the probability that a sub-optimal arm $i$ will no longer be pulled on or after the $m_{i,g}$-th phase by,
\begin{align*}
\dfrac{4\log(\psi )}{(\psi )^{\alpha}}\sum_{m=0}^{m_{i,g}}\dfrac{1}{\epsilon_{m_{i,g}}^{2\alpha +1}} &\leq \dfrac{4\log(\psi )}{(\psi )^{\alpha}}\sum_{m=0}^{M}\left(\dfrac{1}{\epsilon_{m}}\right)^{2\alpha +1}
%%%%%%%%%%%%%%%%%%%%%%%%%%
 \overset{(a)}{\leq} \dfrac{4\log(\psi  )}{(\psi )^{\alpha}}\left(\dfrac{(1+\gamma)((1+\gamma)^M - 1)}{(1+\gamma) - 1}\right)^{2\alpha +1} \\
%%%%%%%%%%%%%%%%%%%%%%%%%%%
&\overset{(b)}{\leq} \dfrac{4\log(\psi )}{(\psi )^{\alpha}}\left( \left( \dfrac{1+\gamma}{\gamma}\right) \sqrt{T}\right)^{2\alpha +1} \overset{(c)}{=} \dfrac{4\sqrt{T}C\left(\gamma, \alpha \right)\log(\psi )}{(\psi)^{\alpha}}.
\end{align*}
Here, in $(a)$ we use the standard geometric progression formula, in $(b)$ we substitute the value of $M=\dfrac{1}{2}\log_{1+\gamma}\frac{T}{e}$ and in $(c)$ we substitute $C\left(\gamma, \alpha \right)= \left( \frac{1+\gamma}{\gamma}\right)^{2\alpha + 1}$. Bounding this trivially by $T\Delta^{opt}_{i,g}$ for each arm $i\in\A'$ we get the regret suffered for all arm $i\in\A'$ after the $m_{i,g}$-th phase  as,
\begin{align*}
\sum_{i\in\A'}\left( \dfrac{4T\Delta^{opt}_{i,g}\sqrt{T}C_{\gamma}\log(\psi  )}{(\psi)^{\alpha}} \right) =  \sum_{i\in\A'}\left( \dfrac{4T^{\frac{3}{2}}C\left(\gamma, \alpha \right)\Delta^{opt}_{i,g}\log(\psi)}{(\psi)^{\alpha}} \right) = \sum_{i\in\A'}\left( \dfrac{4C\left(\gamma, \alpha \right)\Delta^{opt}_{i,g}\log(\psi )}{(\psi T^{-\frac{1}{2\alpha}})^{\alpha}} \right).
\end{align*}

\textbf{Step 5.(Regret for pulling the sub-optimal arm $i$ on or before $m_{i,g}$-th phase):} Either a sub-optimal arm gets pulled $\ell_{m_{i,g}}$ number of times till the $m_{i,g}$-th phase or after that the probability of it getting pulled is exponentially low (as shown in \textbf{step 4}). Hence, the number of times a sub-optimal arm $i$ is pulled till the $m_{i,g}$-th phase is given by,
\begin{align*}
N_{i,g} < \ell_{m_{i,g}} = \left\lceil \dfrac{\log(\psi \epsilon_{m_{i,g}}^2)}{2\epsilon_{m_{i,g}}} \right\rceil
\end{align*}
Hence, considering each arm $i\in\A'$ the total regret is bounded by,
\begin{align*}
\sum_{i\in\A'}\Delta^{opt}_{i,g}\left\lceil \dfrac{\log(\psi \epsilon_{m_{i,g}}^2)}{2\epsilon_{m_{i,g}}} \right\rceil < \sum_{i\in\A'}\Delta^{opt}_{i,g}\left[ 1 + \dfrac{\log(\psi \epsilon_{m_{i,g}}^2)}{2\epsilon_{m_{i,g}}} \right] \leq \sum_{i\in\A'}\Delta^{opt}_{i,g}\left[ 1 + \dfrac{8\log(\psi (\Delta^{opt}_{i,g})^4)}{(\Delta^{opt}_{i,g})^2}\right].
\end{align*} 

\textbf{Step 6.(Regret for not detecting changepoint $g$ for arm $i$ after the $p_{i,g}$-th phase):} First we recall a few additional notations, starting with $\hat{\mu}_{i,L_{p_{i,g-1}}:L_{m'}}$ denoting the sample mean of the $i$-th arm from $L_{p_{i,g-1}}$ to $L_{m'}$ timestep while $\hat{\mu}_{i,L_{m'}+1:L_{p_{i,g}}}$ denotes the sample mean of the $i$-th arm from $L_{m'}+1$ to $L_{p_{i,g}}$ timesteps where $m'=1,\ldots,p_{i,g}$. Proceeding similarly as in \textbf{step 4} we can show that in the $p_{i,g}$-th phase for an arm $i\in\A'$ if $N_{i,L_{p_{i,g-1}}:L_{m}}\geq\ell_{p_{i,g}}$ then,

\begin{align*}
S_i = \sqrt{\dfrac{\alpha\log(\psi \epsilon_{p_{i,g}}^2)}{2N_{i,L_{p_{i,g-1}}:L_{m}}}} \leq \sqrt{\dfrac{\alpha\log(\psi \epsilon_{p_{i,g}}^2)}{2\ell_{p_{i,g}}}} \leq \sqrt{\alpha\epsilon_{p_{i,g}}\dfrac{\log(\psi \epsilon_{p_{i,g}}^2)}{\log(\psi \epsilon_{p_{i,g}}^2)}} \leq \dfrac{\Delta^{chg}_{i,g}}{4(1+\gamma)^2}.
\end{align*}
Furthermore, we can show that for a phase $m\in[1,p_{i,g}]$ if the following  conditions hold for an arm $i\in\A'$ then the changepoint will definitely get detected, that is,
\begin{eqnarray}
\hat{\mu}_{i,L_{p_{i,g-1}}:L_{m}} + S_{i,N_{p_{i,g-1}}:L_{m}} &<& \hat{\mu}_{i,L_{m}+1:L_{p_{i,g}}} - S_{i,L_{m}+1:L_{p_{i,g}}},\hspace*{4mm}
\nonumber\\
%%%%%%%%%%%%%%%%%%%%%%%%%%%%%%%
\hat{\mu}_{i,L_{p_{i,g-1}}:L_{m}} - S_{i,L_{p_{i,g-1}}:L_{m}} &>&  \hat{\mu}_{i,L_{m}+1:L_{p_{i,g}}} + S_{i,L_{m}+1:L_{p_{i,g}}},\hspace*{4mm} 
\nonumber\\
%%%%%%%%%%%%%%%%%%%%%%%%%%%%%%%
 S_{i,L_{p_{i,g-1}}:L_{m}} &<& S_{i,L_{m}+1:L_{p_{i,g}}}, \nonumber\\
%%%%%%%%%%%%%%%%%%%%%%%%%%%%%%%
 N_{i,L_{p_{i,g-1}}:L_{m}} &\geq& \ell_{p_{i,g}} \label{eq:arm-pull-chg}.\\\nonumber
\end{eqnarray}

Indeed, we can show that if there is a changepoint at $g$ such that $\mu_{i,g-1} < \mu_{i,g}$ and if the first and third  conditions in equation \eqref{eq:arm-pull-chg} hold  in the $p_{i,g}$-th phase then,
\begin{align*}
\hat{\mu}_{i,L_{p_{i,g-1}}:L_{m}} + S_{i,L_{p_{i,g-1}}:L_{m}} &\leq {\mu}_{i,g-1} + 4(1+\gamma)^2 S_{i,L_{p_{i,g-1}}:L_{m}} - 2(1+\gamma)^2S_{i,L_{p_{i,g-1}}:L_{m}}\\
%%%%%%%%%%%%%%%%%%%
& \leq {\mu}_{i,g-1} + \Delta^{chg}_{i,g} - 2(1+\gamma)^2S_{i,L_{p_{i,g-1}}:L_{m}}\\
%%%%%%%%%%%%%%%%%%%
 &\leq {\mu}_{i,g} - 2(1+\gamma)^2S_{i,L_{m}+1:L_{p_{i,g}}}\\
%%%%%%%%%%%%%%%%%%%
 & \leq \hat{\mu}_{i,L_{m}+1:L_{p_{i,g}}} - S_{i,L_{m}+1:L_{p_{i,g}}}.
\end{align*}
Conversely, for the case $\mu_{i,g-1} > \mu_{i,g}$ the second and third conditions will hold for the $p_{i,g}$-th phase. Now, to bound the regret we need to bound the probability of the complementary of these four events. For the complementary of the first event in equation \eqref{eq:arm-pull-chg} by using Chernoff-Hoeffding bound we can show that for all $m'\in[1,m]$,
\begin{align*}
\sum_{m'=1}^{m}\Pb( \hat{\mu}_{i,L_{p_{i,g-1}}:L_{m'}} \geq  \mu_{i,g-1} + S_{i,L_{p_{i,g-1}}:L_{m'}} ) & \leq \sum_{m'=0}^{p_{i,g}}\sum_{n=1}^{\ell_{p_{i,g}}}\Pb\bigg( \dfrac{\sum_{s=1}^n X_{i,s}}{n} \geq \mu_{i,g-1} + \sqrt{\dfrac{\alpha\log(\psi \epsilon_{m'_{}}^2)}{2n}} \bigg)\\
%%%%%%%%%%%%%%%%%%%%%%%%%%%%%%%%%%
%&\leq \sum_{m'=0}^{p_{i,g}}\sum_{n=1}^{\ell_{m}}\exp\left(-2\dfrac{\alpha\log(\psi \epsilon_{m'}^2)}{2n_{}}n_{} \right)\\
%%%%%%%%%%%%%%%%%%%%%%%%%%%%%%%%%%%
%&\leq \sum_{m'=0}^{p_{i,g}}\sum_{n=1}^{\ell_{m}}\dfrac{1}{\psi \epsilon_{m'}^2}\\
%%%%%%%%%%%%%%%%%%%%%%%%%%%%%%%%%%%
%&\leq \sum_{m'=0}^{p_{i,g}}\dfrac{\log(\psi \epsilon_{m'}^2)}{2\epsilon_{m'}}\dfrac{1}{(\psi \epsilon_{m'}^2 )^{\alpha}} \\
%%%%%%%%%%%%%%%%%%%%%%%%%%%%%%%%%%
&\leq  \dfrac{\log(\psi )}{2(\psi )^{\alpha}}\sum_{m'=0}^{p_{i,g}}\dfrac{1}{\epsilon_{m'}^{2\alpha + 1}}.
\end{align*}
Also, again by using Chernoff-Hoeffding bound we can show that for all $m'\in[m,p_{i,g}]$,
\begin{align*}
\sum_{m'=m}^{p_{i,g}}\Pb( \hat{\mu}_{i,L_{m'}+1:L_{p_{i,g}}} \leq  \mu_{i,g} - S_{i,L_{m'}+1:L_{p_{i,g}}} ) &\leq \sum_{m'=0}^{p_{i,g}}\sum_{n=1}^{\ell_{p_{i,g}}}\Pb\bigg( \dfrac{\sum_{s=1}^n X_{i,s}}{n} \geq \mu_{i,g} - \sqrt{\dfrac{\alpha\log(\psi \epsilon_{m'_{}}^2)}{2n}} \bigg)\\
%%%%%%%%%%%%%%%%%%%%%%%%%%%%%%%%%%
%&\leq \sum_{m'=0}^{p_{i,g}}\sum_{n=1}^{\ell_{m}}\exp\left(-2\dfrac{\alpha\log(\psi \epsilon_{m'}^2)}{2n_{}}n_{} \right)\\
%%%%%%%%%%%%%%%%%%%%%%%%%%%%%%%%%%%
%&\leq \sum_{m'=0}^{p_{i,g}}\sum_{n=1}^{\ell_{m}}\dfrac{1}{\psi \epsilon_{m'}^2} \\
%%%%%%%%%%%%%%%%%%%%%%%%%%%%%%%%%%%
%&\leq \sum_{m'=0}^{p_{i,g}}\dfrac{\log(\psi T\epsilon_{m'}^2)}{2\epsilon_{m'}}\dfrac{1}{(\psi \epsilon_{m'}^2 )^{\alpha}} \\
%%%%%%%%%%%%%%%%%%%%%%%%%%%%%%%%%%%
&\leq  \dfrac{\log(\psi )}{2(\psi )^{\alpha}}\sum_{m'=0}^{p_{i,g}}\dfrac{1}{\epsilon_{m'}^{2\alpha + 1}}.
\end{align*}
Hence, the probability that the changepoint is not detected by the first event in equation \eqref{eq:arm-pull-chg} is bounded by,
%\begin{align*}
$\dfrac{\log(\psi  )}{(\psi  )^{\alpha}}\sum_{m=0}^{p_{i,g}}\dfrac{1}{\epsilon_{m}^{2\alpha + 1}}$.
%\end{align*}
Similarly, we can bound the probability of the complementary of the second event in equation \eqref{eq:arm-pull-chg} for a $m'\in[1,m]$ or an $m'\in[m,p_{i,g}]$ as,
\begin{align*}
&\sum_{m'=1}^{m}\Pb( \hat{\mu}_{i,t_s:L_{m'}} \leq  \mu_{i,g-1} - S_{i,t_s:L_{m'}} )\leq \dfrac{\log(\psi )}{2(\psi )^{\alpha}}\sum_{m'=0}^{p_{i,g}}\dfrac{1}{\epsilon_{m'}^{2\alpha + 1}} \textbf{,  and  }\\
%%%%%%%%%%%%%%%%%%%%%%%%%
&\sum_{m'=m}^{p_{i,g}}\Pb( \hat{\mu}_{i,L_{m'}+1:p_{i,g}} \geq  \mu_{i,g} + S_{i,L_{m'}+1,p_{i,g}} )\leq \dfrac{\log(\psi )}{2(\psi )^{\alpha}}\sum_{m'=0}^{p_{i,g}}\dfrac{1}{\epsilon_{m'}^{2\alpha + 1}}.
\end{align*}
Hence, the probability that the changepoint is not detected by the second event in equation \eqref{eq:arm-pull-chg} is again upper  bounded by,
%\begin{align*}
$\dfrac{\log(\psi  )}{(\psi  )^{\alpha}}\sum_{m=0}^{p_{i,g}}\dfrac{1}{\epsilon_{m}^{2\alpha + 1}}$.
%\end{align*}
Again, for the third event, following a similar procedure in \textbf{step 4}, we can bound its complementary event by,
\begin{align*}
\sum_{m=0}^{p_{i,g}} \Pb( S_{i,L_{p_{i,g-1}}:L_{m}} \geq S_{i,L_{m}+1:L_{p_{i,g}}}) \leq 2\left(\dfrac{\log(\psi )}{2(\psi )^{\alpha}}\right)\sum_{m=0}^{p_{i,g}}\dfrac{1}{\epsilon_{m_{}}^{2\alpha + 1}}.
\end{align*}
And finally, for the fourth event we can bound its complementary event by,
\begin{align*}
\sum_{m=0}^{p_{i,g}} \Pb( N_{i,L_{p_{i,g-1}}:L_{m}} < \ell_{p_{i,g}}) \leq 2\left(\dfrac{\log(\psi )}{2(\psi )^{\alpha}}\right)\sum_{m=0}^{p_{i,g}}\dfrac{1}{\epsilon_{m_{}}^{2\alpha + 1}}.
\end{align*}
Combining the contribution from these four events, we can show that the probability of not detecting the changepoint $g$ for the $i$-th arm after the $p_{i,g}$-th phase is upper bounded by,
\begin{align*}
\dfrac{2\log(\psi )}{(\psi \epsilon_{p_{i,g}}^2)^{\alpha}}\sum_{m=0}^{p_{i,g}}\dfrac{1}{\epsilon_{p_{i,g}}^{2\alpha + 1}} \overset{(a)}{\leq} \dfrac{4C\left(\gamma,\alpha\right)\sqrt{T}\log(\psi  )}{(\psi)^{\alpha}} .
\end{align*}
Here, in $(a)$ we substitute $C\left(\gamma,\alpha\right)=\left( \dfrac{1+\gamma}{\gamma}\right)^{2\alpha + 1}$ and we follow the same steps as in \textbf{step 4} to reduce the expression to the above form. Furthermore, bounding the regret trivially (after the changepoint $g$) by $\Delta^{opt}_{i,g+1}$ for each arm $i\in\A'$, we get 
\begin{align*}
\sum_{i\in\A'}\left( \dfrac{2T\Delta^{opt}_{i,g+1}\sqrt{T}C\left(\gamma,\alpha\right)\log(\psi  )}{(\psi)^{\alpha}} \right) &= \sum_{i\in\A'}\left( \dfrac{4T^{\frac{3}{2}}C\left(\gamma,\alpha\right)\Delta^{opt}_{i,g+1}\log(\psi )}{(\psi)^{\alpha}} \right) \\
%%%%%%%%%%%%%%%%%%%%%%%
&= \sum_{i\in\A'}\left( \dfrac{4 C\left(\gamma, \alpha \right)\Delta^{opt}_{i,g}\log(\psi  )}{(\psi T^{-\frac{1}{2\alpha}})^{\alpha}} \right).
\end{align*}

\textbf{Step 7.(Regret for not detecting a changepoint $g$ for arm $i\in\A$ on or before the $p_{i,g}$-th phase):} The regret for not detecting the changepoint $g$ on or before the $p_{i,g}$-th phase can be broken into two parts, \textbf{(a)} the worst case events from $t_{g}$ to $p_{i}$ and \textbf{(b)} the minimum number of pulls $\ell_{p_{i,g}}$ required to detect the changepoint. For the first part (a) we can use assumption \ref{assm:space-gap}, assumption \ref{assm:chg-gap}, discussion \ref{dis:gap-delay} and definition \ref{Def:e-chg-gap} to upper bound the regret as,

\begin{align*}
\sum_{i\in\A}\Delta^{opt}_{\max,g+1}\left\lceil \dfrac{8\log(\psi \epsilon_{p_{i,g}}^2)}{(\Delta^{}_{\epsilon_0, g})^2} \right\rceil < \sum_{i\in\A}\Delta^{opt}_{\max,g+1}\left[ 1 + \dfrac{8\log(\psi \epsilon_{p_{i,g}}^2)}{(\Delta^{\epsilon_0}_{g})^2} \right].
\end{align*}
Again, for the second part (b) the arm $i\in\A'$ can be pulled no more than $\ell_{p_i}$ number of times. Hence, for each arm $i\in\A$  the regret for this case is bounded by,
\begin{align*}
\sum_{i\in\A'}\Delta^{opt}_{i,g+1}\left\lceil \dfrac{\log(\psi \epsilon_{p_{i,g}}^2)}{2\epsilon_{p_{i,g}}} \right\rceil <  \sum_{i\in\A'}\Delta^{opt}_{i,g+1}\left[ 1 + \dfrac{\log(\psi \epsilon_{p_{i,g}}^2)}{2\epsilon_{p_i, g}} \right].
\end{align*}
Therefore, combining these two parts (a) and (b) we can show that the total regret for not detecting the changepoint till the $p_{i,g}$-th phase is given by,
\begin{align*}
&\sum_{i\in\A}\left\lbrace\Delta^{opt}_{\max,g+1} + \dfrac{8\Delta^{opt}_{\max,g+1}\log(\psi \epsilon_{p_{i,g}}^2)}{(\Delta^{}_{\epsilon_0, g})^2}\right\rbrace + \sum_{i\in\A'}\left\lbrace\Delta^{opt}_{i,g+1} + \dfrac{\Delta^{opt}_{i,g+1}\log(\psi \epsilon_{p_{i,g}}^2)}{2\epsilon_{p_{i,g}}}\right\rbrace \\
%%%%%%%%%%%%%%%%%%%%%%%%%
&\leq \sum_{i\in\A}\left\lbrace\Delta^{opt}_{\max,g+1} + \dfrac{8\Delta^{opt}_{\max,g+1}\log(\psi (\Delta^{chg}_{i,g})^4)}{(\Delta^{}_{\epsilon_0, g})^2} \right\rbrace + \sum_{i\in\A'}\left\lbrace \Delta^{opt}_{i,g+1} + \dfrac{8\Delta^{opt}_{i,g+1}\log(\psi (\Delta^{chg}_{i,g})^4)}{(\Delta^{chg}_{i,g})^2}\right\rbrace.
\end{align*}

\textbf{Step 8.(Maximal delay of \ImpCPD ($d_{\pi}(t_g - \tau_{g-1})$):} Following the same approach as in \textbf{Step 6} from Theorem \ref{psbandit:Theorem:1} we can argue that the changepoint detection policy \ImpCPD suffers a maximal delay when it may pull each arm $i=1,\ldots, K$, at most $N_{i,min_D} = \dfrac{8K\log(\psi (\Delta^{chg}_{i,g})^4)}{(\Delta^{chg}_{i,g})^2}$ times (from \textbf{Step 7}) before finally detecting the changepoint for the $K$-th arm. Similarly, we can argue that for each arm $i=1,2,\ldots,K$, the probability of not detecting the changepoint from time $t_g$ till $T$ is upper bounded by $\Pb\lbrace \xi^{chg}_{i,t'}\rbrace \leq \dfrac{4K C\left(\gamma, \alpha \right)\log(\psi  )}{(\psi T^{-\frac{1}{2\alpha}})^{\alpha}}$ from (from \textbf{Step 6}).

Now for $\Delta^{opt}_{i,g}=\Delta^{chg}_{i,g}=\Delta(t_g, \delta) = \sqrt{\frac{K\log (T/G)}{\frac{T}{G}}} > \Omega(\sqrt{\frac{\log t}{t}}) , \forall i\in\A,\forall g\in\G$ and $\gamma=0.05$ we can show that,
\begin{align*}
\dfrac{8K\log(\psi (\Delta^{chg}_{i,g})^4)}{(\Delta^{chg}_{i,g})^2} \leq   8KT
\text{ and   } \Pb( \xi^{chg}_{i,t'}) \leq \dfrac{4K C\left(\gamma, \alpha \right)\log(\psi  )}{(\psi T^{-\frac{1}{2\alpha}})^{\alpha}} \leq G\sqrt{\dfrac{K^3\log (T)}{T}}.
\end{align*}
Now, again using assumption \ref{assm:space-gap} and $\delta =\frac{1}{T}$ we can show that,
\begin{align*}
& \left(KN_{i,\min_D} +  2K\sum_{t'=N_{i,\min_{D}}}^{t_{2}}\Pb\lbrace \xi^{chg}_{i,t'}\rbrace\right) -  \left( \frac{C(t,\delta,\eta)K\log(\frac{T}{{\delta}})}{2(\Delta(t_g,\delta))^{2}}\right)  \leq \eta (t_3 - t_2)\\
%%%%%%%%%%%%%%%%
& \Leftrightarrow \eta \geq  \dfrac{8 T}{2T\log T + T}.
\end{align*}
Finally, $d_{\pi}(t_g - \tau_{g-1})$ for the $g$-th changepoint,
\begin{align*}
d_{\pi}(t_g - \tau_{g-1}) &= d_{\pi}(t_g - (t_{g-1} + d_{\pi}(t_{g-1} - \tau_{g-2}))) = \ldots \\
%%%%%%%%%%%%%%%%%%%%%%%
&= d_{\pi}(t_g - (t_{g-1} + d_{\pi}(t_{g-1} - \ldots (t_1 + d_{\pi}(t_2 - \tau_1))))\\
%%%%%%%%%%%%%%%%%%%%%%%
& \overset{(a)}{\leq} d_{\pi}(t_g - (t_{g-1} + d_{\pi}(t_{g-1} - \tau_{g-2}))) \leq d_{\pi}(t_{g-1} - \tau_{g-2}) + d_{\pi}(t_g - \tau_{g-1})
\end{align*}
where $(a)$ holds for each changepoint from $1,\ldots,g-1$ by assumption \ref{assm:space-gap}. Hence, the for $ \eta \geq  \frac{8 }{2\log T + 1}$ can be upper bounded as,
\begin{align*}
d_{\pi}(t_{g-1} - \tau_{g-2}) + d_{\pi}(t_g - \tau_{g-1}) < \dfrac{8K\log(\psi T(\Delta^{chg}_{i,g})^4)}{(\Delta^{chg}_{i,g})^2} +  \dfrac{4K C\left(\gamma, \alpha \right)\log(\psi T )}{(\psi T^{-\frac{3}{2\alpha}})^{\alpha}}.
\end{align*}
%\eta > \dfrac{32}{\log(T/G)(1 + 4\log(T/G))}$ delay $d_{\pi}(t_g - \tau_{g-1})
%Again note that for $\eta \geq K$ the maximal delay of ImpCPD is surely less than $d(t_g - \tau_{g-1})$ in Lemma \ref{psbandit:Lemma:01}.
%
%\begin{align*}
%\dfrac{8K\log(\psi T(\Delta^{chg}_{i,g})^4)}{(\Delta^{chg}_{i,g})^2} < \dfrac{8\eta\log(\psi T(\Delta^{chg}_{i,g})^4)}{(\Delta^{chg}_{\epsilon_0,g})^2} < d(t_g - \tau_{g-1}) \leq  \left( \frac{C_{\eta}K\log(\frac{T^2}{G^2{\epsilon_0}})}{2(\Delta^{}_{\epsilon_0, g})^{2}}\right)
%\end{align*}
%where, $C_{\eta} \leq 4K\log (T/G)$.

\textbf{Step 9.(Final Regret bound):} Combining all the steps before and considering the repeating worst case of undetectable gaps, the expected regret till the $T$-th timestep is bounded by,
\begin{align*}
& \E[R_{T}] \leq \sum_{j=1}^{G}\sum_{i=1}^{K}\bigg[ 1 
%%%%%%%%%%
+ \underbrace{ \dfrac{8C\left(\gamma, \alpha \right)\Delta^{opt}_{i,g}\log(\psi T )}{(\psi T^{-\frac{3}{2\alpha}})^{\alpha}}  }_{\textbf{from Step 4}}
%%%%%%%%%%
+ \underbrace{\Delta^{opt}_{i,g} + \dfrac{8\log(\psi T(\Delta^{opt}_{i,g})^4)}{(\Delta^{opt}_{i,g})}}_{\textbf{from Step 5}}
%%%%%%%%%%%%%%%%%%%%%%%%%%%
 + \underbrace{ \dfrac{8C\left(\gamma, \alpha \right)\Delta^{opt}_{i,g}\log(\psi T )}{(\psi T^{-\frac{3}{2\alpha}})^{\alpha}}  }_{\textbf{from Step 6}}\bigg]\\
%%%%%%%%%%%%%%%%%%%%%%%%%%%
&+ \underbrace{ \sum_{j=1}^{G}\sum_{i\in \A^{chg}_{g}}\bigg[\Delta^{opt}_{\max,g+1} + \dfrac{8\Delta^{opt}_{\max,g+1}\log(\psi T(\Delta^{chg}_{i,g})^4)}{(\Delta(t_g,\delta))^2}\bigg] + \sum_{i\in\A'}\sum_{j=1}^{G}\bigg[\Delta^{opt}_{i,g+1} + \dfrac{8\Delta^{opt}_{i,g+1}\log(\psi T(\Delta^{chg}_{i,g})^4)}{(\Delta^{chg}_{i,g})^2}}_{\textbf{from Step 7}}\bigg]\\
%%%%%%%%%%%%%%%%%%%%%%%%%%%
&+  \sum_{j=1}^{G}\sum_{i\in \A^{chg}_{g}}\bigg[\underbrace{\dfrac{8K\log(\psi T(\Delta^{chg}_{i,g})^4)}{(\Delta(t_g,\delta))^2} +  \dfrac{4K C\left(\gamma, \alpha \right)\log(\psi T )}{(\psi T^{-\frac{3}{2\alpha}})^{\alpha}}}_{\textbf{from Step 8}}\bigg] + \max_{i\in\A: \!\!\frac{e}{\sqrt{T}}\leq \! \Delta_i <\! \Delta(t_g,\delta)}\Delta_i T.
\end{align*}
Now substituting $\alpha  = \frac{3}{2}$ and $\psi = \frac{T^2}{K^2\log K}$ in the above result we derive the bound in Theorem \ref{psbandit:Theorem:2}.
\end{customproof}

\section{Proof of Corollary \ref{psbandit:Corollary:1}}
\label{sec:proof:Corollary:1}

\begin{customproof}{7}
\label{proof:Corollary:1}

%\subsection{Gap-independent bound of \UCBLCPD}
We first recall the result of Theorem \ref{psbandit:Theorem:1} below,
\begin{align*}
\E[R_t] & \overset{}{\leq} \underbrace{\sum_{g=1}^{G}\sum_{i=1}^{K} \bigg\{ 25 + \dfrac{6\log(t)}{\Delta^{opt}_{i,g}}\bigg\}}_{\textbf{part A}} \! +\! \sum_{g=1}^{G}\sum_{i\in \A_g^{chg}}\!\bigg\{ \underbrace{\dfrac{12\Delta^{opt}_{i,g}\log(t)}{(\Delta^{chg}_{i,g})^2}}_{\textbf{part B}}\!+\! \underbrace{\dfrac{18\Delta^{opt}_{\max,g+1}\log(t)}{(\Delta(t_g,\delta))^{2}}}_{\textbf{part C}}\\
%%%%%%%%%%%%%%%%%%%
& \!+\! \underbrace{\dfrac{12K\Delta^{opt}_{\max,g+1}\log(t)}{(\Delta(t_g, \delta))^2} \!+\! 8K}_{\textbf{part D}}\!\bigg\rbrace  
 + \underbrace{\max_{i\in\A: \!\!\frac{e}{\sqrt{T}}\leq \! \Delta_i <\! \Delta(t_g,\delta)}\Delta_i t}_{\textbf{part E}}.
\end{align*}
%\begin{align*}
%\E[R_t]  &\leq \sum_{i=1}^{K}\sum_{j=1}^{G} \bigg\lbrace \underbrace{7 + \dfrac{6\log(t)}{\Delta^{opt}_{i,g}}}_{\textbf{part A}} + \underbrace{\dfrac{6\Delta^{opt}_{i,g}\log(t)}{(\Delta^{chg}_{i,g})^2} }_{\textbf{part B}} + \underbrace{ \dfrac{6\Delta^{opt}_{\max,g+1}\log(t)}{(\Delta^{}_{\epsilon_0, g})^{2}} }_{\textbf{part C}} + \underbrace{\dfrac{12\Delta^{opt}_{i,g+1}\log(t)}{(\Delta^{chg}_{i,g})^{2}}}_{\textbf{part D}}+ \underbrace{\dfrac{12K\Delta^{opt}_{\max,g+1}\log(t)}{(\Delta^{}_{\epsilon_0, g})^{2}} + 4K}_{\textbf{part E}}\bigg\rbrace.
%\end{align*}
Then, substituting $\Delta^{opt}_{i,g}=\Delta^{chg}_{i,g}=\Delta(t_g,\delta)=\sqrt{\dfrac{K\log (T/G)}{\frac{T}{G}}}, \forall i\in\A,\forall g\in\G$ we can show that,
% for \textbf{part A},
\begin{align*}
\text{(part A)} &\leq 25KG + \dfrac{6\sqrt{KGT}\log T}{\sqrt{{\log (T/G)}}}  \leq 25KG  + 6\sqrt{KGT}\log T,\\
\text{(part B)} &\leq 12\sqrt{GKT}\log T,\\
\text{(part C)} &\leq  18\sqrt{GKT}\log T,\\
\text{(part D)} &\leq 12K\sqrt{KGT}\log T + 8K^2G,\\
\text{(part E)} &\leq \sqrt{T\log T}.
\end{align*}

%\subsection{Gap-independent of \ImpCPD}
Again, we first recall the result of Theorem \ref{psbandit:Theorem:2} below,
%\begin{align*}
%\E[R_{T}] &\leq \sum_{i\in\A'}\sum_{j=1}^{G}\bigg[ 1 
%%%%%%%%%%%
%+ \underbrace{ \dfrac{32 C_1\left(\gamma \right)\Delta^{opt}_{i,g}\log(\frac{T}{K\sqrt{\log K}} )}{(K^2\log K)^{-\frac{3}{2}}}  }_{\textbf{part A}}
%%%%%%%%%%%
%+ \underbrace{ \Delta^{opt}_{i,g} +  \dfrac{16\log(\frac{T (\Delta^{opt}_{i,g})^2}{K\sqrt{\log K}})}{(\Delta^{opt}_{i,g})}   }_{\textbf{part B}}\bigg]\\
%%%%%%%%%%%%%%%%%%%%%%%%%%%%
%% + \underbrace{ \dfrac{8C\left(\gamma, \alpha \right)\Delta^{opt}_{i,g}\log(\psi T )}{(\psi T^{-\frac{3}{2\alpha}})^{\alpha}}  }_{\textbf{part C}}\bigg]\\
%%%%%%%%%%%%%%%%%%%%%%%%%%%%
%&+ \underbrace{ \sum_{i\in\A}\sum_{j=1}^{G}\bigg[\Delta^{opt}_{\max,g+1} + \dfrac{16 \Delta^{opt}_{\max,g+1}\log( \frac{T(\Delta^{chg}_{i,g})^2}{K\sqrt{ \log K}})}{(\Delta^{}_{\epsilon_0, g})^2}\bigg] + \sum_{i\in\A'}\sum_{j=1}^{G}\bigg[\Delta^{opt}_{i,g+1} + \dfrac{16\Delta^{opt}_{i,g+1}\log(\frac{T(\Delta^{chg}_{i,g})^2}{K\sqrt{\log K}})}{(\Delta^{chg}_{i,g})^2}\bigg]  }_{\textbf{part C}}
%\bigg] \\
%%%%%%%%%%%%%%%%%%%%%%%%%%%%
%& + \sum_{i\in\A'}\sum_{j=1}^{G}\bigg[\underbrace{\dfrac{8K\log(\psi T(\Delta^{chg}_{i,g})^4)}{(\Delta(t_g,\delta))^2} +  \dfrac{4K C\left(\gamma, \alpha \right)\log(\psi T )}{(\psi T^{-\frac{3}{2\alpha}})^{\alpha}}}_{\textbf{part D}}\bigg] 
%\end{align*}
%
\begin{align*}
&\E[R_{T}]  \leq \sum_{j=1}^{G}\sum_{i=1}^{K}\bigg[ 1 
%%%%%%%%%%
+ \underbrace{ \dfrac{16C\left(\gamma, \alpha \right)\Delta^{opt}_{i,g}\log(\psi T )}{(\psi T^{-\frac{3}{2\alpha}})^{\alpha}}  }_{\textbf{part A}}
%%%%%%%%%%
+ \underbrace{\Delta^{opt}_{i,g} + \dfrac{8\log(\psi T(\Delta^{opt}_{i,g})^4)}{(\Delta^{opt}_{i,g})}}_{\textbf{part B}}\bigg]\\
%%%%%%%%%%%%%%%%%%%%%%%%%%%
%%%%%%%%%%%%%%%%%%%%%%%%%%%
&+ \underbrace{ \sum_{j=1}^{G}\sum_{i\in \A^{chg}_{g}}\bigg[\Delta^{opt}_{\max,g+1} + \dfrac{8\Delta^{opt}_{\max,g+1}\log(\psi T(\Delta^{chg}_{i,g})^4)}{(\Delta(t_g,\delta))^2}\bigg] + \sum_{i\in\A'}\sum_{j=1}^{G}\bigg[\Delta^{opt}_{i,g+1} + \dfrac{8\Delta^{opt}_{i,g+1}\log(\psi T(\Delta^{chg}_{i,g})^4)}{(\Delta^{chg}_{i,g})^2}}_{\textbf{part C}}\bigg]\\
%%%%%%%%%%%%%%%%%%%%%%%%%%%
&+  \sum_{j=1}^{G}\sum_{i\in \A^{chg}_{g}}\bigg[\underbrace{\dfrac{8K\log(\psi T(\Delta^{chg}_{i,g})^4)}{(\Delta(t_g,\delta))^2} +  \dfrac{4K C\left(\gamma, \alpha \right)\log(\psi T )}{(\psi T^{-\frac{3}{2\alpha}})^{\alpha}}}_{\textbf{part D}}\bigg] + \underbrace{\max_{i\in\A: \!\!\frac{e}{\sqrt{T}}\leq \! \Delta_i <\! \Delta(t_g,\delta)}\Delta_i T}_{\textbf{part E}}.
\end{align*}
Then, substituting $\Delta^{opt}_{i,g}=\Delta^{chg}_{i,g}=\Delta(t_g,\delta)=\sqrt{\dfrac{K\log (T/G)}{\frac{T}{G}}}, \forall i\in\A,\forall g\in\G$ and $\gamma=0.05$ such that $C_1\left( \gamma\right) = C_1 = 9261$ we can show that for \textbf{part A},
%$\alpha=1.5$, $\psi = \dfrac{T}{(K^2\log K)}$ and
\begin{align*}
\text{(part A)} &\leq 32C_1 G^{1.5}K^4(\log K)^{1.5} \sqrt{\log K}\dfrac{(\log T)^{1.5}}{\sqrt{T}} \leq C_1 G^{1.5}K^{4.5}(\log K)^{2}.
\end{align*}
Again, for \textbf{part B} we can show that,
\begin{align*}
\text{(part B)} &\leq G^{1.5}K\sqrt{\dfrac{K\log (T/G)}{T}} + \dfrac{16G\sqrt{KT}\log\left(\log TG\right) }{\sqrt{G\log T}} \overset{(a)}{\leq} G^{1.5}\sqrt{\dfrac{K^3\log (T/G)}{T}} + 16\sqrt{GKT}.
\end{align*}
where, $(a)$ comes from the identity that $\dfrac{\log(\log (T))}{\sqrt{\log T}}\leq 1$ and $\dfrac{\log(\log G)}{\sqrt{\log T}}\leq 1$. 

And finally, for \textbf{part C} and \textbf{part D}, similar to \textbf{part B}, we can show that,
\begin{align*}
\text{(part C)} &\leq  G^{1.5}\sqrt{\dfrac{K^3\log (T/G)}{T}} + 16\sqrt{GKT} + G\sqrt{\dfrac{K^3\log (T/G)}{T}} + 16\sqrt{GKT} \\
%%%%%%%%%%%%%%%%%%%%%%%%%%%%%
& = 2G^{1.5}\sqrt{\dfrac{K^3\log (T/G)}{T}} + 32\sqrt{GKT}.\\
\text{(part D)} &\leq 32K\sqrt{GKT} + C_1 G^{1.5}K^{5.5}(\log K)^{2}.\\
\text{(part E)} &\leq \sqrt{T\log T}.
\end{align*}
So, combining the four parts we get that that the gap-independent regret upper bound of \ImpCPD is,
\begin{align*}
\E[R_T]&\leq 3G^{1.5}\sqrt{\dfrac{K^3\log (T/G)}{T}} + C_1 G^{1.5}K^{4.5}(\log K)^{2} + 48\sqrt{GKT} + 32K\sqrt{GKT} \\
& + C_1 G^{1.5}K^{5.5}(\log K)^{2} + \sqrt{T\log T}.
\end{align*}
\end{customproof}

\section{Proof of Regret Lower Bound of Oracle Policy $\pi^o$}
%\begin{customtheorem}{3}\textbf{(Lower Bounds for oracle policy)}
%\label{psbandit:Theorem:3}
%The lower bound of an oracle policy $\pi^o$ for a horizon $T$, $K$ arms and $G$ changepoints is given by,
%%\begin{align*}
%$\!\E_{\pi^o}[R_T]\geq \!\!\min\!\left\lbrace\!\Omega\!\left(\!\sum\limits_{g=1}^G\sum\limits_{i=1}^{K}\!\frac{ \log{(T/(GH^{}_{1,g}))}}{(\Delta_{i,g}^{opt})}\right), \!\Omega\left(\sqrt{GT}\!\right)\!\right\rbrace$
%%\end{align*}
%where, $H^{}_{1,g} = \sum\limits_{i=1}^{K}{(\Delta^{opt}_{i,g})^{-2}}$ is the hardness of the problem.
%\end{customtheorem}
%
%\begin{customproof}{7}
%An oracle policy $\pi^o$ has access to the exact changpoints. The worst case scenario can occur when  environment changes uniform randomly. A similar argument has also been made in the adaptive-bandit setting of \citep{DBLP:journals/jmlr/MaillardM11}. So, let the horizon $T$ be divided into $G$ slices, each of length $(T/G)$. For each of these slices an oracle algorithm using OCUCB \citep{lattimore2015optimally} should get the optimal SMAB regret without suffering any delay. The proof is in Appendix \ref{proof:Theorem:3}.
%\end{customproof}
%
%
\label{proof:Theorem:3}
\begin{customproof}{9} 
We follow the same steps as in Theorem 2 of \citet{DBLP:journals/jmlr/MaillardM11} for proving the lower bound.

\textbf{Step 1.(Assumption):} The change of environment is not controlled by the learner and so the worst case scenario can be that the environment changes uniform randomly.  A similar argument has also been made in the adaptive-bandit setting of \cite{DBLP:journals/jmlr/MaillardM11}. So, let the horizon $T$ be divided into $G$ slices, each of length $\frac{T}{G}$. Hence, $T \ = G\frac{T}{G}$.

\subsection*{Gap-dependent bound}
\textbf{Step 2.(Gap-dependent result):} From the stochastic bandit literature \citep{audibert2009minimax}, \citep{bubeck2012regret}, \citep{lattimore2015optimally} we know that the gap-dependent regret bound for a horizon $T$ and $K$ arms in the stochastic bandit setting is lower bounded by,
\begin{align*}
\E[R_T]_{SMAB-gap-dependent} \geq C\sum_{i=1}^{K} \dfrac{\log(\frac{T}{(H_{1,g}})}{\Delta^{opt}_{i,g}}
\end{align*}
where, $C$ is a constant and $H_{1,g} = \sum\limits_{i=1}^{K}\frac{1}{\Delta^{opt}_{i,g}}$ is the optimality hardness for the $g$-th changepoint. Note, that the SMAB setting is a special case of the piecewise i.i.d setting where there is a single changepoint at $t_0 = 1$.

\textbf{Step 3.(Regret for $G$ changepoints):} The oracle policy $\pi^o$ has access to the changepoints and is restarted without suffering any delay. Hence,  combining Step 1 and Step 2 we can show that the gap-dependent regret lower bound scales as,
\begin{align*}
\E[R_T]_{gap-dependent} & \geq \sum\limits_{g=1}^{G}\left(C\sum_{i=1}^{K} \dfrac{\log(\frac{\frac{T}{G}}{(H_{1,g}})}{\Delta^{opt}_{i,g}}\right)
%%%%%%%%%%%%%%%%%%%%%%
 \geq C_1\sum\limits_{g=1}^G\sum\limits_{i=1}^K\dfrac{ \log{\frac{T}{GH_{1,g}}}}{\Delta_{i,g}^{opt}}
\end{align*}
where, $C_1$ is a constant and $H_{1,g}$ is the optimality hardness defined above.
 %for the changepoint $g\in\G$ such that $H_{1,g} =  \sum\limits_{i=1}^{K}\frac{1}{\Delta_{i,g}^{opt}}$.

\subsection*{Gap-independent bound}

\textbf{Step 4.(Reward of arms):} Let, for the interval $\rho_g$ the optimal arm has a Bernoulli reward distribution of $\mu_{i^*,g} = Ber(\frac{(1+\epsilon_g)}{2})$ and all the other arms $i\in\A \setminus \lbrace i^*_g\rbrace$ have a Bernoulli reward distribution of $\mu_{i,g} = Ber(\frac{(1 - \epsilon_g)}{2})$.

\textbf{Step 5.(Regret for $G$ changepoints):} Let ${I}_g(t)$ denote the number of times a sub-optimal arm $i$ is pulled between $t_g - 1$ timestep to $t_g$ timestep. Now from Lemma 6.6 in \citet{bubeck2010bandits} we know that for an $\epsilon_g$ of the order of $\sqrt{\frac{K}{s}}$ the following inequality holds,
\begin{align*}
\sup_{i^*}\sum_{t=1}^{s}(\mu_{i^*,g} - \mu_{i_t,g}) \geq s\epsilon_g\left(1 - \dfrac{1}{K} - \sqrt{\dfrac{s\epsilon_g}{2K}\log\left( \dfrac{1-\epsilon_g}{1+\epsilon_g}\right)}\right).
\end{align*}

In the adversarial setup, the adversary(or environment) chooses $t_g$ such that $T$ exactly divided into $\frac{T}{G}$ slices of $G$ times, then the regret is lower bounded as,
\begin{align*}
\sup_{(i^*_g)_g}\sum_{g=1}^{G}\sum_{t=t_{g-1}}^{t_g}(\mu_{i^*,g} - \mu_{i_t,g})&\geq \sum_{g=1}^{G}\sum_{t=t_{g-1}}^{t_g}\epsilon_g\left(1 - \dfrac{1}{K} - \sqrt{\dfrac{{I}_g(t)\epsilon_g}{2K}\log\left( \dfrac{1-\epsilon_g}{1+\epsilon_g}\right)}\right )\\
& \geq \sum_{g=1}^{G}\dfrac{T}{G}\epsilon_g\left(1 - \dfrac{1}{K} - \sqrt{\dfrac{T\epsilon_g}{2KG}\log\left( \dfrac{1-\epsilon_g}{1+\epsilon_g}\right)}\right).
\end{align*}
The right hand side of the above expression can be optimized to yield that for any $\epsilon_g \approx \sqrt{\frac{T}{{I}_g(t)}}$,
\begin{align*}
\sup_{(i^*_g)_g}\sum_{g=1}^{G}\sum_{t=t_{g-1}}^{t_g}(\mu_{i^*,g} - \mu_{i_t,g})
& \geq \dfrac{1}{20}\sum_{g=1}^{G}\sqrt{\dfrac{T}{KG}} =  \dfrac{1}{20}\sqrt{KGT}.
\end{align*}
Hence, the gap-independent regret bound for the oracle policy $\pi^o$ is given by 
\begin{align*}
\E[R_T]_{gap-independent} \geq \dfrac{1}{20}\sqrt{KGT}.
\end{align*}

\end{customproof}

\end{document}